\documentclass[journal]{IEEEtran}

\usepackage{graphics} %
\usepackage{graphicx}
\usepackage{algpseudocode,algorithm}

\usepackage{amsmath,amsfonts,amssymb}
\usepackage{color}
\usepackage{wrapfig}
\usepackage{subfigure}
\usepackage{multirow}
\usepackage{array}
\usepackage[table]{xcolor}
\usepackage{soul}
\usepackage{mathtools}
\usepackage{fixmath}
\usepackage{nccmath}
\usepackage{cuted}
\usepackage{float}
\usepackage{tikz}
\usepackage{caption}
\usepackage{siunitx}
\usepackage{url}
\usepackage{graphicx}
\usepackage{booktabs}
\usepackage{makecell}
\newcolumntype{P}[1]{>{\centering\arraybackslash}p{#1}}
\newcolumntype{M}[1]{>{\centering\arraybackslash}m{#1}}

\DeclareMathOperator*{\argmin}{arg\,min}

\newcommand{\ie}{\textit{i}.\textit{e}.}
\newcommand{\eg}{\textit{e}.\textit{g}.}

\newcolumntype{L}[1]{>{\raggedright\let\newline\\\arraybackslash\hspace{0pt}}m{#1}}
\newcolumntype{C}[1]{>{\centering\let\newline\\\arraybackslash\hspace{0pt}}m{#1}}
\newcolumntype{R}[1]{>{\raggedleft\let\newline\\\arraybackslash\hspace{0pt}}m{#1}}

\usepackage{threeparttable}
\usepackage{balance}

\global\long\def\bx{\mathbf{x}}

\global\long\def\br{\mathbf{r}}
\global\long\def\bt{\mathbf{t}}
\global\long\def\bc{\mathbf{c}}

\global\long\def\Rot{\mathtt{R}}
\global\long\def\tl{\text{left}}
\global\long\def\tr{\text{right}}
\global\long\def\taul{\tau_{\tl}}
\global\long\def\taur{\tau_{\tr}}

\global\long\def\rhos{\rho^{\star}}
\global\long\def\cT{\mathcal{T}}
\global\long\def\btau{\boldsymbol{\tau}}

\global\long\def\mJ{\mathbf{J}} %
\global\long\def\btheta{\boldsymbol{\theta}}
\global\long\def\bzero{\mathbf{0}}

\global\long\def\ThreeDtoTwoD{3\text{D}-2\text{D}}

\global\long\def\cS{\mathcal{S}} %
\global\long\def\cF{\mathcal{F}} %
\global\long\def\cFref{\cF_{\text{ref}}} %
\global\long\def\cD{\mathcal{D}} %
\global\long\def\bT{\mathbf{T}} %

\global\long\def\GTS{GTS}
\global\long\def\rpgreader{\emph{rpg\_reader}}
\global\long\def\rpgbin{\emph{rpg\_bin}}
\global\long\def\rpgbox{\emph{rpg\_box}}
\global\long\def\rpgdesk{\emph{rpg\_desk}}
\global\long\def\rpgmonitor{\emph{rpg\_monitor}}
\global\long\def\upennflyOne{\emph{upenn\_flying1}}
\global\long\def\upennflyThree{\emph{upenn\_flying3}}
\global\long\def\ijrrsimulation{\emph{simulation\_3planes}}
\global\long\def\studt{\emph{Student's}~$t$}

\usepackage[draft,authormarkup=none]{changes}
\definechangesauthor[color=blue]{GG}
\definechangesauthor[color=purple]{J}

\usepackage[amsmath,amsthm,thmmarks]{ntheorem}

\usepackage{adjustbox} %

\makeatletter
\def\endthebibliography{%
  \def\@noitemerr{\@latex@warning{Empty `thebibliography' environment}}%
  \endlist
}
\makeatother

\newcommand\MYhyperrefoptions{bookmarks=true,bookmarksnumbered=true,
pdfpagemode={UseOutlines},plainpages=false,pdfpagelabels=true,
colorlinks=true,citecolor={black},
pdftitle={Event-based Stereo Visual Odometry},%
pdfsubject={Robotics, Computer Vision, SLAM},%
pdfauthor={Y. Zhou, G. Gallego, S. Shen},%
pdfkeywords={Event Cameras, Bio-Inspired Vision, Asynchronous sensor, Low Latency, High Dynamic Range, Low Power}}%
\usepackage[\MYhyperrefoptions,pdftex]{hyperref}

\newif\ifclearsectionlook
\clearsectionlookfalse

\newif\ifrevised
\revisedfalse %

\newif\ifshowfigures
\showfigurestrue

\begin{document}
\bstctlcite{IEEEexample:BSTcontrol}
\title{Event-based Stereo Visual Odometry}

\author{Yi~Zhou,
Guillermo~Gallego,
Shaojie~Shen
\thanks{Yi Zhou and Shaojie Shen are with the Robotic Institute, 
the Department of Electronic and Computer Engineering at the Hong Kong University of Science and Technology, Hong Kong, China. 
E-mail: \{eeyzhou, eeshaojie\}@ust.hk. 
Guillermo Gallego is with the Technische Universit\"at Berlin and the Einstein Center Digital Future, Berlin, Germany.}
\thanks{This work was supported by the HKUST Institutional Fund. (Corresponding author: Yi Zhou.)}}

\setcounter{figure}{-2} %
\makeatletter
\g@addto@macro\@maketitle{
\vspace{4ex}
\begin{center}
  \centering
  \includegraphics[width=0.9\textwidth]{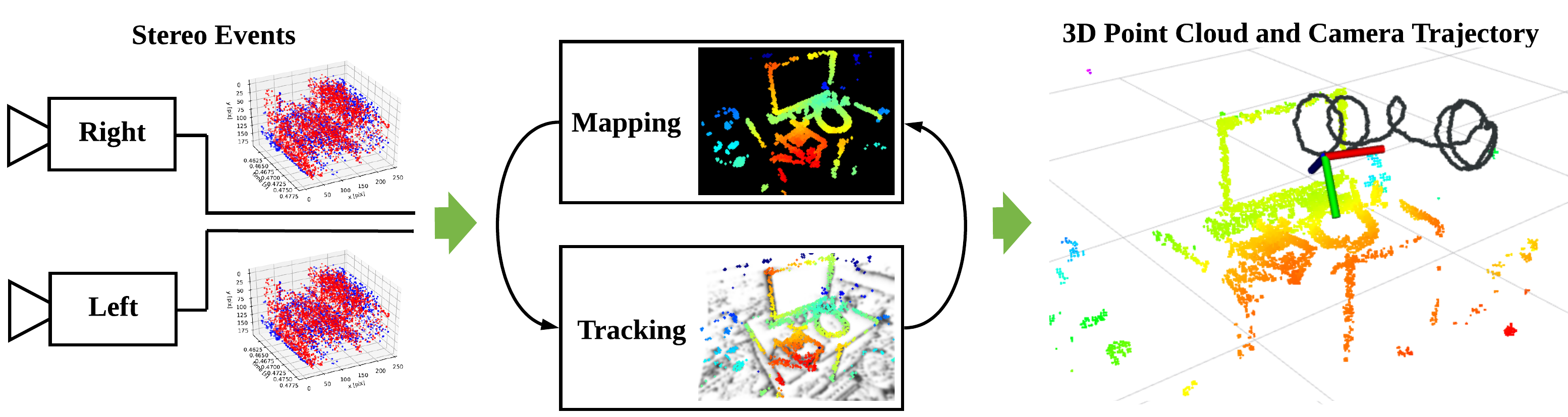}
  \captionof{figure}{The proposed system takes as input the asynchronous data acquired by a pair of event cameras in stereo configuration (Left) and recovers the motion of the cameras as well as a semi-dense map of the scene (Right).
  It exploits spatio-temporal consistency of the events across the image planes of the cameras to solve both localization (i.e., 6-DoF tracking) and mapping (i.e., depth estimation) subproblems of visual odometry (Middle). 
  The system runs in real time on a standard CPU.
  }
  \label{fig:eyecatcher}
\end{center}%

\vspace{-2ex}
}
\makeatother
\maketitle

\begin{abstract}
Event-based cameras are bio-inspired vision sensors whose pixels work independently from each other and respond asynchronously to brightness changes, with microsecond resolution. 
Their advantages make it possible to tackle challenging scenarios in robotics, such as high-speed and high dynamic range scenes. 
We present a solution to the problem of visual odometry from the data acquired by a stereo event-based camera rig.
Our system follows a parallel tracking-and-mapping approach, 
where novel solutions to each subproblem (3D reconstruction and camera pose estimation) are developed 
with two objectives in mind: being principled and efficient, for real-time operation with commodity hardware.
To this end, we seek to maximize the spatio-temporal consistency of stereo event-based data while using a simple and efficient representation.
Specifically, the mapping module builds a semi-dense 3D map of the scene by fusing depth estimates from multiple viewpoints (obtained by spatio-temporal consistency) in a probabilistic fashion.
The tracking module recovers the pose of the stereo rig by solving a registration problem that naturally arises %
due to the chosen map and event data representation.
Experiments on publicly available datasets and on our own recordings demonstrate the versatility of the proposed method in natural scenes with general 6-DoF motion.
The system successfully leverages the advantages of event-based cameras to perform visual odometry in challenging illumination conditions, such as low-light and high dynamic range, while running in real-time on a standard CPU.
We release the software and dataset under an open source licence to foster research in the emerging topic of event-based SLAM.
\end{abstract}

\ifclearsectionlook\cleardoublepage\fi \section*{Multimedia Material}
Supplemental video: {\small \url{https://youtu.be/3CPPs1gz04k}}\\
Code: {\small\url{https://github.com/HKUST-Aerial-Robotics/ESVO.git}}

\section{Introduction}
\label{sec:introduction}
Event cameras are novel bio-inspired sensors that report the pixel-wise intensity changes asynchronously at the time they occur, called ``events'' \cite{Lichtsteiner08ssc,Gallego20pami}.
Hence, they do not output grayscale images nor they operate at a fixed rate like traditional cameras.
This asynchronous and differential principle of operation suppresses temporal redundancy and therefore reduces power consumption and bandwidth.
Endowed with microsecond resolution, event cameras are able to capture high-speed motions, which would cause severe motion blur on standard cameras. 
In addition, event cameras have a very high dynamic range (HDR) (e.g., \SI{140}{\decibel} compared to \SI{60}{\decibel} of standard cameras), which allows them to be used on broad illumination conditions.
Hence, event cameras open the door to tackle challenging scenarios in robotics such as high-speed and/or HDR feature tracking~\cite{Lagorce15tnnls,Zhu17icra,Gehrig19ijcv},
camera tracking~\cite{Mueggler15rss,Gallego17pami,Gallego17ral,Bryner19icra},
control~\cite{Conradt09iscas,Delbruck13fns,Falanga20scirob} and Simultaneous Localization and Mapping (SLAM)~\cite{Kim16eccv,Rebecq17ral,Rosinol18ral,Mueggler18tro}.

The main challenge in robot perception with these sensors is to design new algorithms that process the unfamiliar stream of intensity changes (``events'') and are able to unlock the camera's potential~\cite{Gallego20pami}.
Some works have addressed this challenge by combining event cameras with additional sensors, 
such as depth sensors~\cite{Weikersdorfer14icra} or standard cameras~\cite{Censi14icra,Kueng16iros}, 
to simplify the perception task at hand.
However, this introduced bottlenecks due to the combined system being limited by the lower speed and dynamic range of the additional sensor.

In this paper we tackle the problem of stereo visual odometry (VO) with event cameras in natural scenes and arbitrary 6-DoF motion.
To this end, we design a system that processes a stereo stream of events in real time and outputs the ego-motion of the stereo rig and a map of the 3D scene (Fig.~\ref{fig:eyecatcher}).
The proposed system essentially follows a parallel tracking-and-mapping philosophy~\cite{Klein07ismar}, where the main modules operate in an interleaved fashion estimating the ego-motion and the 3D structure, respectively (a more detailed overview of the system is given in Fig.~\ref{fig:system-overview}).
In summary, our \emph{contributions} are:
\begin{itemize}
    \item A novel mapping method based on the optimization of an objective function designed to measure spatio-temporal consistency across stereo event streams (Section~\ref{sec:mapping:inverse-depth-estimation}).
    \item A fusion strategy based on the probabilistic characteristics of the estimated inverse depth to improve density and accuracy of the recovered 3D structure (Section~\ref{sec:depthfusion}).
    \item A novel camera tracking method based on $\ThreeDtoTwoD$ registration that leverages the inherent distance field nature of a compact and efficient event representation (Section~\ref{sec:tracking}).
    \item An extensive experimental evaluation, on publicly available datasets and our own, demonstrating that the system is computationally efficient, running in real time on a standard CPU (Section~\ref{sec:experiments}).
    The software, design of the stereo rig and datasets used have been open sourced.
\end{itemize}

\noindent This paper significantly extends and differs from our previous work~\cite{Zhou18eccv}, which only tackled the stereo mapping problem. 
Details of the differences are given at the beginning of Section~\ref{sec:mapping}. 
In short, we have completely reworked the mapping part due to the challenges faced for real-time operation.

Stereo visual odometry (VO) is a paramount task in robot navigation, and we aim at bringing the advantages of event-based vision to the application scenarios of this task.
To the best of our knowledge, this is the first published stereo VO algorithm for event cameras (see Section~\ref{sec:related-work}).

\paragraph*{Outline}
The rest of the paper is organized as follows.
Section~\ref{sec:related-work} reviews related work in 3D reconstruction and ego-motion estimation with event cameras.
Section~\ref{sec:system-overview} provides an overview of the proposed event-based stereo VO system,
whose mapping and tracking modules are described in Sections~\ref{sec:mapping} and~\ref{sec:tracking}, respectively.
Section~\ref{sec:experiments} evaluates the proposed system extensively on publicly available data, demonstrating its effectiveness.
Finally, Section~\ref{sec:conclusion} concludes the paper.

\ifclearsectionlook\cleardoublepage\fi \section{Related Work}
\label{sec:related-work}

Event-based stereo VO is related to several problems in structure and motion estimation with event cameras. 
These have been intensively researched in recent years, notably since event cameras such as the Dynamic Vision Sensor (DVS)~\cite{Lichtsteiner08ssc} became commercially available (2008).
Here we review some of those works. 
A more extensive survey is provided in~\cite{Gallego20pami}.

\subsection{Event-based Depth Estimation (3D Reconstruction)}
\label{sec:related:mapping}
\paragraph{Instantaneous Stereo}
The literature on event-based stereo depth estimation is dominated by methods that tackle the problem of 3D reconstruction using data from a pair of synchronized and rigidly attached event cameras during a very short time (ideally, on a per-event basis).
The goal is to exploit the advantages of event cameras to reconstruct dynamic scenes at very high speed and with low power.
These works~\cite{Kogler11isvc,Rogister12tnnls,CamunasMesa14fns} typically follow the classical two-step paradigm of finding epipolar matches and then triangulating the 3D point~\cite{Hartley03book}.
Event matching is often solved by enforcing several constraints, including temporal coherence (e.g., simultaneity) of events across both cameras.
For example, \cite{Ieng18fnins} combines epipolar constraints, temporal inconsistency, motion inconsistency and photometric error (available only from grayscale events given by ATIS cameras~\cite{Posch11ssc}) into an objective function to compute the best matches.
Other works, such as~\cite{Piatkowska14msci,Firouzi16npl,Osswald17srep}, extend cooperative stereo~\cite{Marr76Science} to the case of event cameras~\cite{Steffen19fnbot}.
These methods work well with static cameras in uncluttered scenes, so that event matches are easy to find among few moving objects.

\paragraph{Monocular}
Depth estimation with a single event camera has been shown in~\cite{Kim16eccv,Rebecq18ijcv,Gallego18cvpr}. 
Since instantaneous depth estimation is brittle in monocular setups, these methods tackle the problem of depth estimation for VO or SLAM:
hence, they require knowledge of the camera motion to integrate information from the events over a longer time interval and be able to produce a semi-dense 3D reconstruction of the scene.
Event simultaneity does not apply, hence temporal coherence is much more difficult to exploit to match events across time and therefore other techniques are devised.

\subsection{Event-based Camera Pose Estimation}
\label{sec:related:tracking}
Research on event-based camera localization has progressed by addressing scenarios of increasing complexity.
From the perspective of the type of motion, constrained motions, such as pure rotation~\cite{Cook11ijcnn,Kim14bmvc,Gallego17ral,Reinbacher17iccp} or planar motion~\cite{Weikersdorfer12robio,Weikersdorfer13icvs} have been studied before investigating the most general case of arbitrary 6-DoF motion.
Regarding the type of scenes, solutions for artificial patterns, such as high-contrast textures and/or structures (line-based or planar maps)~\cite{Weikersdorfer12robio,Mueggler14iros,Mueggler15rss}, have been proposed before solving more difficult cases: natural scenes with arbitrary 3D structure and photometric variations~\cite{Kim14bmvc,Gallego17pami,Bryner19icra}.

From the methodology point of view, probabilistic filters~\cite{Weikersdorfer12robio,Kim14bmvc,Gallego17pami} provide event-by-event tracking updates thus achieving minimal latency (\si{\micro\second}), 
whereas frame-based techniques (often non-linear optimization) trade off latency for more stable and accurate results~\cite{Gallego17ral,Bryner19icra}.

\subsection{Event-based VO and SLAM}
\label{sec:related:slam}

\paragraph{Monocular}
Two methods stand out as solving the problem of monocular event-based VO for 6-DoF motions in natural 3D scenes.
The approach in~\cite{Kim16eccv} simultaneously runs three interleaved Bayesian filters, which estimate image intensity, depth and camera pose.
The recovery of intensity information and depth regularization make the method computationally intensive, thus requiring dedicated hardware (GPU) for real-time operation. 
In contrast, \cite{Rebecq17ral} proposes a geometric approach based on the semi-dense mapping technique in~\cite{Rebecq18ijcv} (focusing events~\cite{Gallego19cvpr}) and an image alignment tracker that works on event images. It does not need to recover absolute intensity and runs in real time on a CPU.
So far, none of these methods have been open sourced to the community.

\paragraph{Stereo}
The authors are aware of the existence of a stereo VO demonstrator built by event camera manufacturer~\cite{Migliore20icraw}; however its details have not been disclosed.
Thus, to the best of our knowledge, this is the first published stereo VO algorithm for event cameras.
In the experiments (Section~\ref{sec:experiments}) we compare the proposed algorithm against an iterative-closest point (ICP) method, which is the underlying technology of the above demonstrator.

Our method builds upon our previous mapping work~\cite{Zhou18eccv}, reworked, and a novel camera tracker that re-utilizes the data structures used for mapping. 
For mapping, we do not follow the classical paradigm of event matching plus triangulation, but rather a forward-projection approach that enables depth estimation without establishing event correspondences explicitly. 
Instead, we reformulate temporal coherence using the compact representation of space-time provided by time surfaces~\cite{Lagorce17pami}.
For tracking, we use non-linear optimization on time surfaces, thus resembling the frame-based paradigm, which trades off latency for efficiency and accuracy.
Like~\cite{Rebecq17ral} our system does not need to recover absolute intensity and is efficient, able to operate in real time without dedicated hardware (GPU);  
standard commodity hardware such as a laptop's CPU suffices.

\ifclearsectionlook\cleardoublepage\fi \section{System Overview}
\label{sec:system-overview}

\begin{figure}[t]
\centering
\includegraphics[width=\columnwidth]{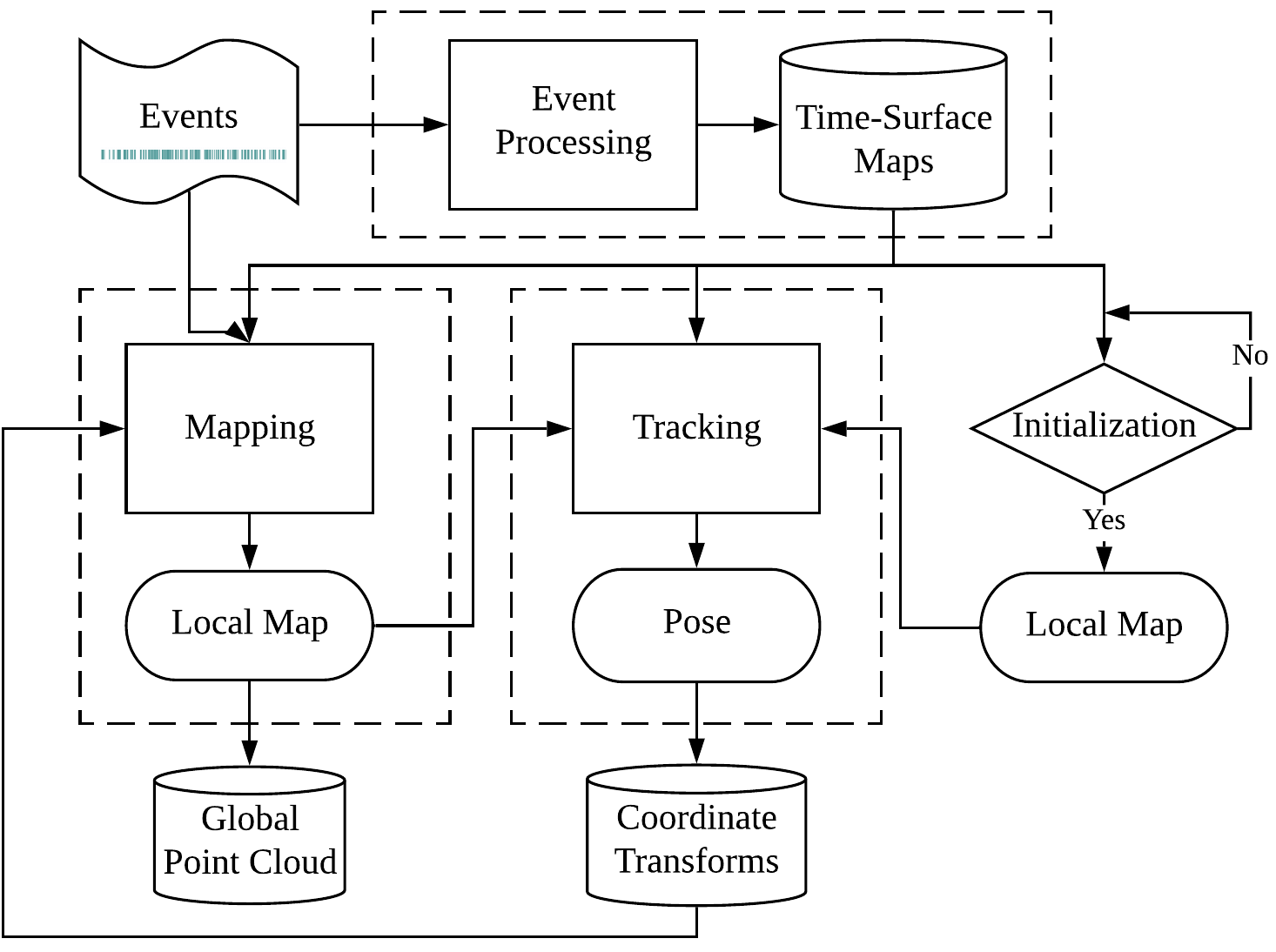}
\caption{\label{fig:system-overview}\emph{Proposed system flowchart}. 
Core modules of the system, including event pre-processing (Section~\ref{sec:timesurfaces}), mapping (Section~\ref{sec:mapping}) and tracking (Section~\ref{sec:tracking}) are marked with dashed rectangles.
The only input to the system comprises raw stereo events from calibrated cameras, 
and the output consists of camera rig poses and a point cloud of 3D scene edges.
}
\end{figure}
The proposed stereo VO system takes as input only raw events from calibrated cameras 
and manages to simultaneously estimate the pose of the stereo event camera rig while reconstructing the environment using semi-dense depth maps.
An overview of the system is given in Fig.~\ref{fig:system-overview}, in which the core modules are highlighted with dashed lines.
Similarly to classical SLAM pipelines~\cite{Klein07ismar}, the core of our system consists of two interleaved modules: mapping and tracking.
Additionally, there is a third key component: event pre-processing.

Let us briefly introduce the functionality of each module and explain how they work cooperatively.
First of all, the event processing module generates an event representation, called time-surface maps (or simply ``time surfaces'', see Section~\ref{sec:timesurfaces}), used by the other modules.
Theoretically, these time maps are updated asynchronously, with every incoming event (\si{\micro\second} resolution).
However, considering that a single event %
does not bring much information to update the state of a VO system, 
the stereo time surfaces are updated at a more practical rate: e.g., at the occurrence of a certain number of events or at a fixed rate (\eg, 100 Hz in our implementation).
A short history of time surfaces is stored in a database (top right of Fig.~\ref{fig:system-overview}) for access by other modules.
Secondly, after an initialization phase (see below), the tracking module continuously estimates the pose of the left event camera with respect to the local map.
The resulting pose estimates are stored in a database of coordinate transforms (\eg, TF in ROS~\cite{Quigley09icraoss}), which is able to return the pose at any given time by interpolation in $\text{SE}(3)$.
Finally, the mapping module takes the events, time surfaces and pose estimates and refreshes a local map (represented as a probabilistic semi-dense depth map), which is used by the tracking module.
The local maps are stored on a database of global point cloud for visualization.

\paragraph*{Initialization} 
To bootstrap the system, we apply a stereo method (a modified SGM method~\cite{Hirschmuller08pami}, as discussed in Section~\ref{sec:experiments:mapping:other-stereo}) that provides a coarse initial map.
This enables the tracking module to start working while the mapping module is also started and produces a better semi-dense inverse depth map (more accurate and dense).

\subsection{Event Representation}
\label{sec:timesurfaces}
\begin{figure}[t]
  \centering
  \captionsetup{skip=1ex}
  \includegraphics[width=0.7\columnwidth]{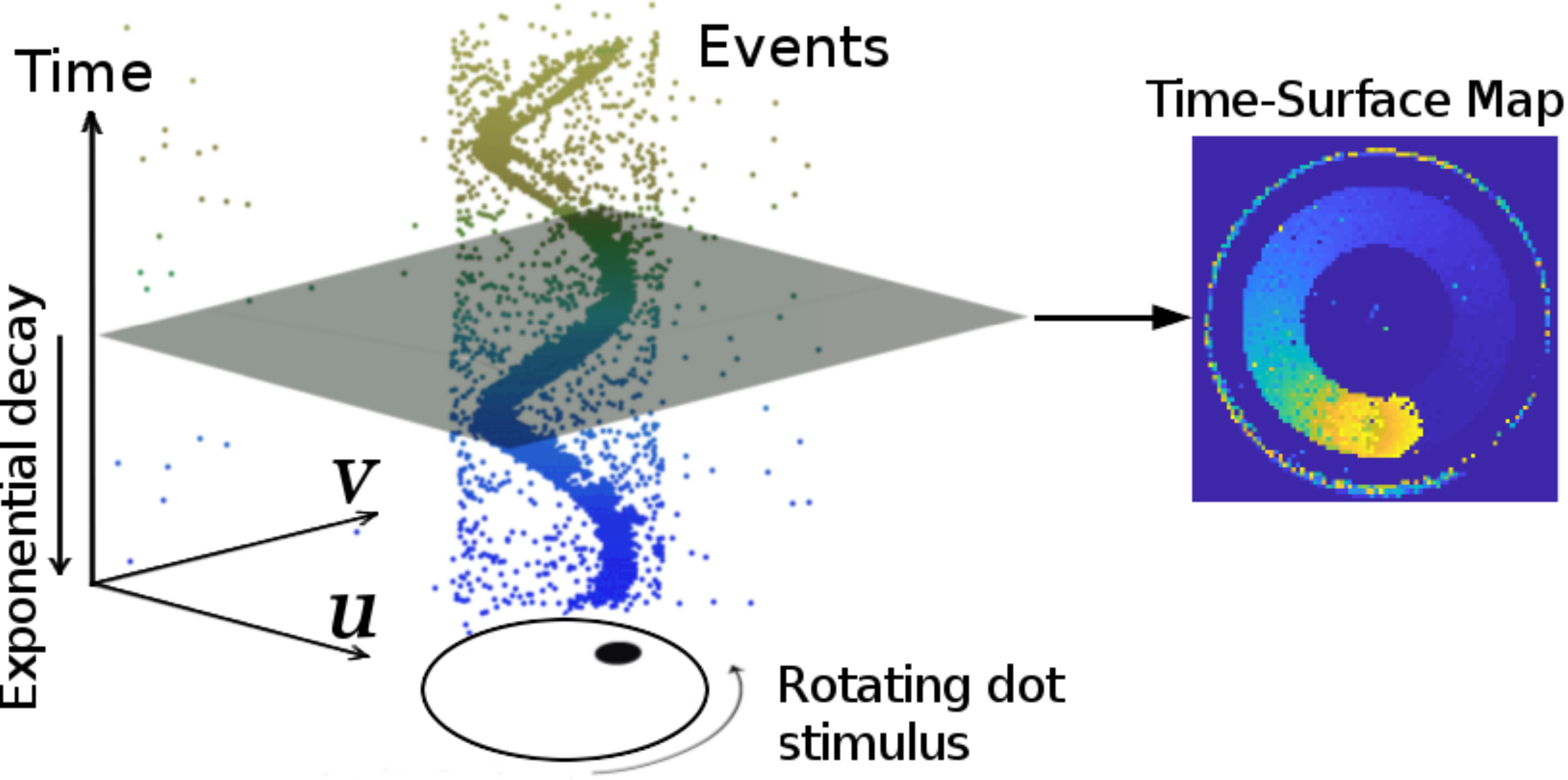}
  \caption{\emph{Event Representation}. 
  Left: Output of an event camera when viewing a rotating dot.
  Right: Time-surface map~\eqref{eq:time-surface-map} at time $t$, $\cT(\bx, t)$, which essentially measures how far in time (with respect to $t$) the last event spiked at each pixel $\bx=(u,v)^\top$.
  The brighter the color, the more recently the event was triggered.
  Figure adapted from~\cite{Liu10nb}.
  \label{fig:time-surface-rotating-dot}}
\end{figure}

As illustrated in Fig.~\ref{fig:time-surface-rotating-dot}-left, the output of an event camera is a stream of asynchronous events.
Each event $e_k=(u_k,v_k,t_k,p_k)$ consists of the space-time coordinates where an intensity change of predefined size happened and the sign (polarity $p_k\in\{+1,-1\}$) of the change.

The proposed system (Fig.~\ref{fig:system-overview}) uses both individual events and an alternative representation called Time Surface (Fig.~\ref{fig:time-surface-rotating-dot}-right).
A time surface (TS) is a 2D map where each pixel stores a single time value, e.g., the timestamp of the last event at that pixel~\cite{Delbruck08issle}.
Using an exponential decay kernel~\cite{Lagorce17pami} TSs emphasize recent events over past events. 
Specifically, if $t_\text{last}$ is the timestamp of the last event at each pixel coordinate $\bx=(u,v)^\top$ the TS at time $t\geq t_\text{last}(\bx)$ is defined by
\begin{equation}
\cT(\bx, t) \doteq \exp\left(-\frac{t - t_\text{last}(\bx)}{\eta}\right),
\label{eq:time-surface-map}
\end{equation}
where $\eta$, the decay rate parameter, is a small constant number (e.g., \SI{30}{\milli\second} in our experiments). 
As shown in Fig.~\ref{fig:time-surface-rotating-dot}-right, 
TSs represent the recent history of moving edges in a compact way (using a 2D grid). 
A discussion of several event representations (voxel grids, event frames, etc.) can be found in~\cite{Gallego20pami,Gehrig19iccv}.

We use TSs because they are memory- and computationally efficient, informative (edges are the most descriptive regions of a scene for SLAM), 
interpretable and because they have proven to be successful for motion (optical flow)~\cite{Delbruck08issle,Benosman14tnnls,Zhu18rss} and depth estimation~\cite{Zhou18eccv}.
Specifically, for mapping (Section~\ref{sec:mapping}) we propose to do pixel-wise comparisons on a stereo pair of TSs~\cite{Zhou18eccv} as a replacement for the photo-consistency criterion of standard cameras~\cite{Engel14eccv}.
Since TSs encode temporal information, comparison of TS patches amounts to measuring \emph{spatio-temporal consistency} over small data volumes on the image planes.
For tracking (Section~\ref{sec:tracking}), we exploit the fact that a TS acts like an anisotropic distance field~\cite{zhou2018canny} defined by the most recent edge locations to register events with respect to the 3D map.
For convenient visualization and processing, \eqref{eq:time-surface-map} is rescaled from $[0,1]$ to the range $[0,255]$.
\ifclearsectionlook\cleardoublepage\fi \section{Mapping: Stereo Depth Estimation\\by Spatio-temporal Consistency and Fusion}
\label{sec:mapping}

The mapping module consists of two steps: 
($i$) computing depth estimates of events (Section~\ref{sec:mapping:inverse-depth-estimation} and Algorithm~\ref{alg:mapping-algorihtm}) and 
($ii$) fusing such depth estimates into an accurate and populated depth map (Section~\ref{sec:depthfusion}).
An overview of the mapping module is provided on Fig.~\ref{fig:mapping:overall-pipeline:flowchart}.

The underlying principles often leveraged for event-based stereo depth estimation are \emph{event co-occurrence} and the epipolar constraint, 
which simply state that a 3D edge triggers two simultaneous events on corresponding epipolar lines of both cameras.
However, as shown in~\cite{benosman11tnn,Piatkowska14msci}, stereo temporal coincidence 
does not strictly hold at the pixel level because of delays, jitter and pixel mismatch (e.g., differences in event firing rates).
Hence, we define a stereo temporal consistency criterion across space-time neighborhoods of the events rather than by comparing the event timestamps at two individual pixels.
Moreover we represent such neighborhoods using time surfaces (due to their properties and natural interpretation as temporal information, Section~\ref{sec:timesurfaces}) and cast the stereo matching problem as the minimization of such a criterion.

The above two-step process and principle was used in our previous work~\cite{Zhou18eccv}.
However, we contribute some fundamental differences guided by a real-time design goal:
($i$) The objective function is built only on the temporal inconsistency across one stereo event time-surface map (Section~\ref{sec:mapping:inverse-depth-estimation}) rather than over longer time spans (thus, the proposed approach becomes closer to the strategy in~\cite{Engel14eccv} than that in~\cite{Newcombe11iccv}). 
This needs to be coupled with ($ii$) a novel depth-fusion algorithm (Section~\ref{sec:depthfusion}), which is provided after investigation of the probabilistic characteristics of the temporal residuals and inverse depth estimates, 
to enable accurate depth estimation over longer time spans than a single stereo time-surface map.
($iii$) The initial guess to minimize the objective is determined using a block matching method, which is more efficient than brute-force search~\cite{Zhou18eccv}.
($iv$) Finally, on a more technical note, non-negative per-patch residuals~\cite{Zhou18eccv} are replaced with signed per-pixel residuals, which guarantee non-zero Jacobians for valid uncertainty propagation and fusion.

\subsection{Inverse Depth Estimation for an Event}
\label{sec:mapping:inverse-depth-estimation}

\begin{figure}[t]
  \centering
  \includegraphics[width=0.95\columnwidth]{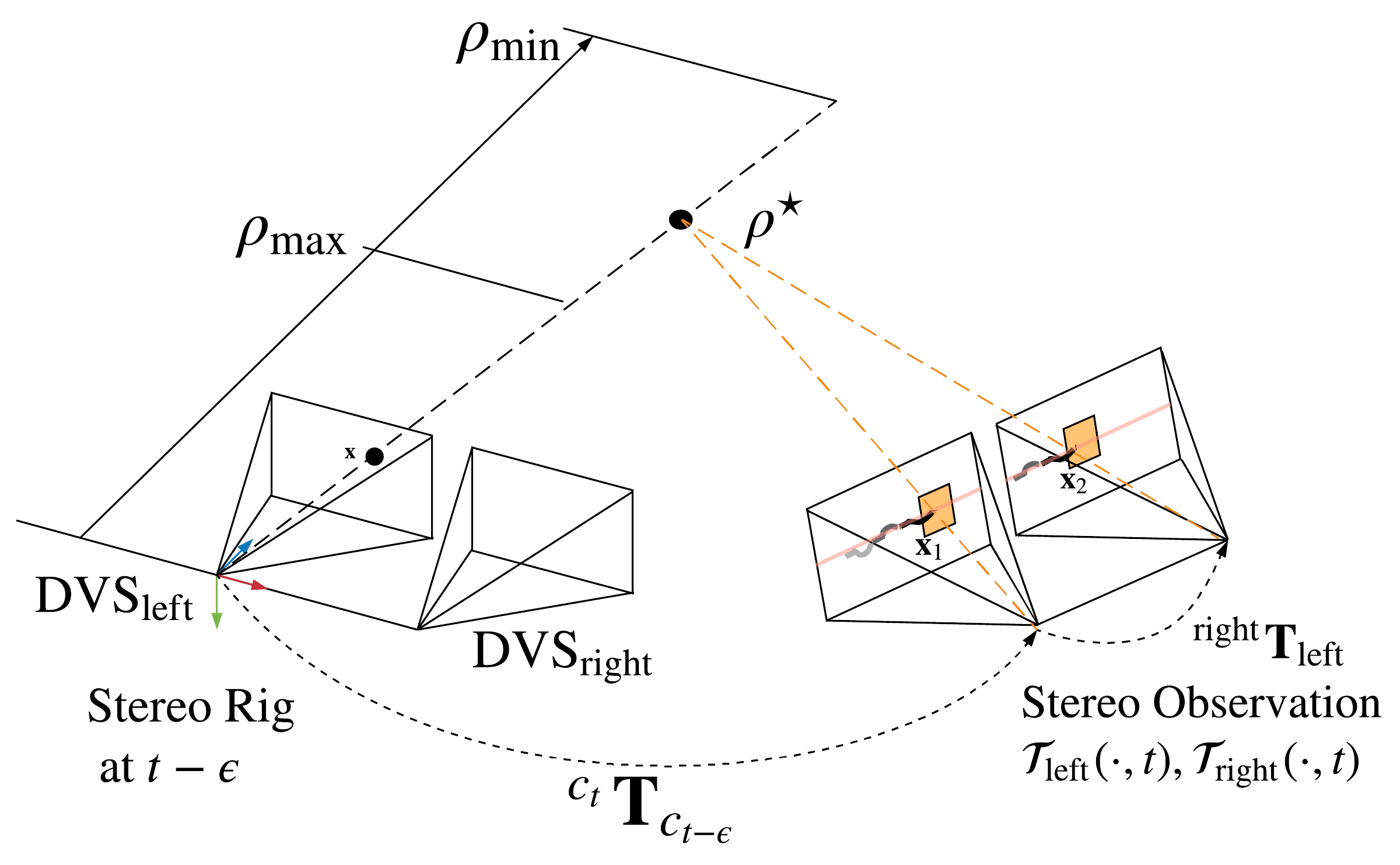}
  \caption{\emph{Mapping. Geometry of (inverse) depth estimation}.
    3D points compatible with an event $e=(\bx,t-\epsilon,p)$ on the left camera are parametrized by inverse depth $\rho$ on the viewing ray through pixel $\bx$ at time $t-\epsilon$.
    The true location of the 3D point that triggered the event corresponds to the value $\rho^{\star}$ that maximizes the temporal consistency across the stereo observation $\bigl(\cT_{\tl}(\cdot,t), \cT_{\tr}(\cdot,t)\bigr)$.
    A search interval $[\rho_{\min}, \rho_{\max}]$ is defined to bound the optimization along the viewing ray.
  \label{fig:geometry}}
\end{figure}

We follow an energy optimization framework to estimate the inverse depth of events occurred before the stereo observation at time $t$.
Fig.~\ref{fig:geometry} illustrates the geometry of the proposed approach.
Without loss of generality, we parametrize inverse depth using the left camera.
A \emph{stereo observation} at time $t$ refers to a pair of time surfaces $\bigl(\cT_{\tl}(\cdot,t), \cT_{\tr}(\cdot,t)\bigr)$ created using~\eqref{eq:time-surface-map} (see also Figs.~\ref{fig:mapping:objective:time-surface-left} and~\ref{fig:mapping:objective:time-surface-right}). 

\subsubsection{Problem Statement}
The inverse depth $\rho^{\star} \doteq 1/Z^{\star}$ of an event $e_{t-\epsilon} \equiv (\bx,t-\epsilon,p)$ (with $\epsilon \in {[0, \delta t]}$) on the left image plane, 
which follows a camera trajectory $\bT_{t-\delta t : t}$, 
is estimated by optimizing the objective function:
\begin{equation}
    \rho^{\star} = \argmin_{\rho}\, C(\bx, \rho, \cT_{\tl}(\cdot,t), \cT_{\tr}(\cdot,t), \bT_{t-\delta t : t})
    \label{eq:optimal-inv-depth}
\end{equation}
\begin{equation}
C \doteq \sum\limits_{\bx_{1,i} \in W_{1}, \bx_{2,i} \in W_{2}} r^{2}_{i}(\rho).
\label{eq:objective-function}
\end{equation}

The residual 
\begin{equation}
\label{eq:mapping-residual}
r_{i}(\rho) \doteq \cT_{\tl}(\bx_{1,i},t) - \cT_{\tr}(\bx_{2,i},t)    
\end{equation}
denotes the temporal difference between two corresponding pixels $\bx_{1,i}$ and $\bx_{2,i}$ inside neighborhoods (i.e., patches) $W_1$ and $W_2$, centered at $\bx_{1}$ and $\bx_{2}$ respectively. 
Assuming the calibration (intrinsic and extrinsic parameters) is known and the pose of the left event camera at any given time within $[t - \delta t, t]$ is available (e.g., via interpolation of $\bT_{t-\delta t : t}$ in SE(3)), the points $\bx_1$ and $\bx_2$ are given by 
\begin{subequations}
\label{eq:corresponding-image-points}
\begin{align}
\bx_{1} &= \pi\bigl({}^{c_{t}}\bT_{c_{t-\epsilon}} \cdot \pi^{-1}(\bx, \rho_k)\bigr),\\
\bx_{2} &= \pi\bigl({}^{\tr}\bT_{\tl} \cdot {}^{c_t}\bT_{c_{t-\epsilon}} \cdot \pi^{-1}(\bx, \rho_{k})\bigr).
\end{align}
\end{subequations}

Note that each event %
is warped using the camera pose at the time of its timestamp.
The function $\pi: \mathbb{R}^{3} \to \mathbb{R}^{2}$ projects a 3D point onto the camera's image plane, 
while its inverse function $\pi^{-1}: \mathbb{R}^{2} \to \mathbb{R}^{3}$ back-projects a pixel into 3D space given the inverse depth $\rho$.
${}^{\tr}\bT_{\tl}$ denotes the transformation from the left to the right event camera, which is constant.
All event coordinates $\bx$ are undistorted and stereo-rectified using the known calibration of the cameras.

\begin{figure}[t]
  \centering
  \subfigure[t][\small{Scene in dataset~\cite{Zhu18ral}.}]{
  \includegraphics[width=0.46\columnwidth]{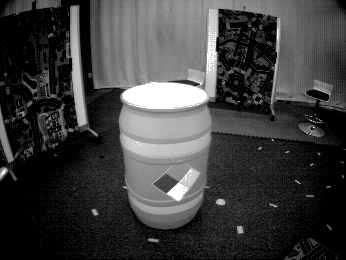}
  \label{fig:mapping:objective:scene}}
  \subfigure[t][\small{Objective function~\eqref{eq:objective-function} (in red).}]{
  \includegraphics[width=0.46\columnwidth]{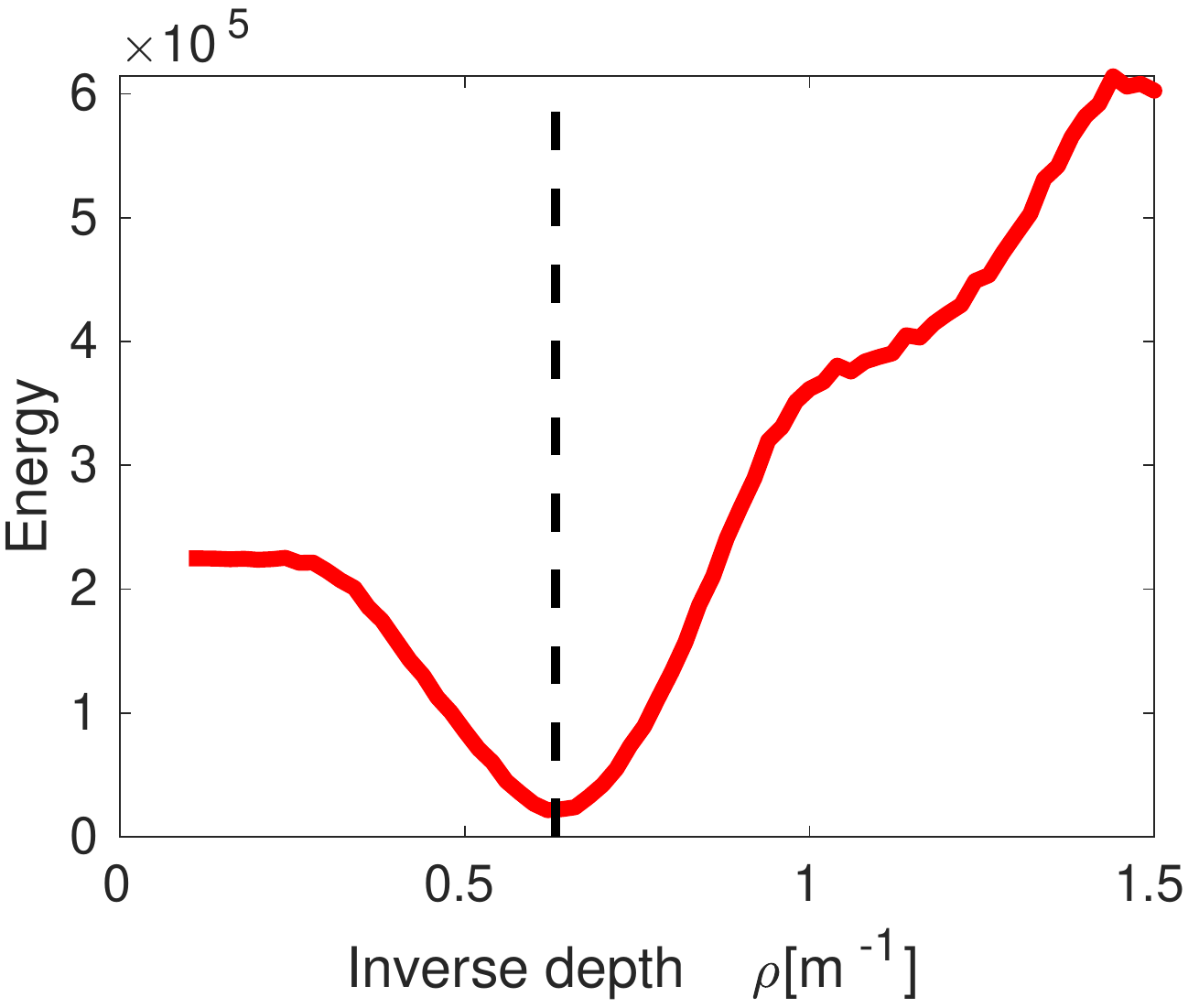}
  \label{fig:mapping:objective:energy}}\,
  \subfigure[t][\small{Time surface (left DVS).}]{
  \includegraphics[width=0.46\columnwidth]{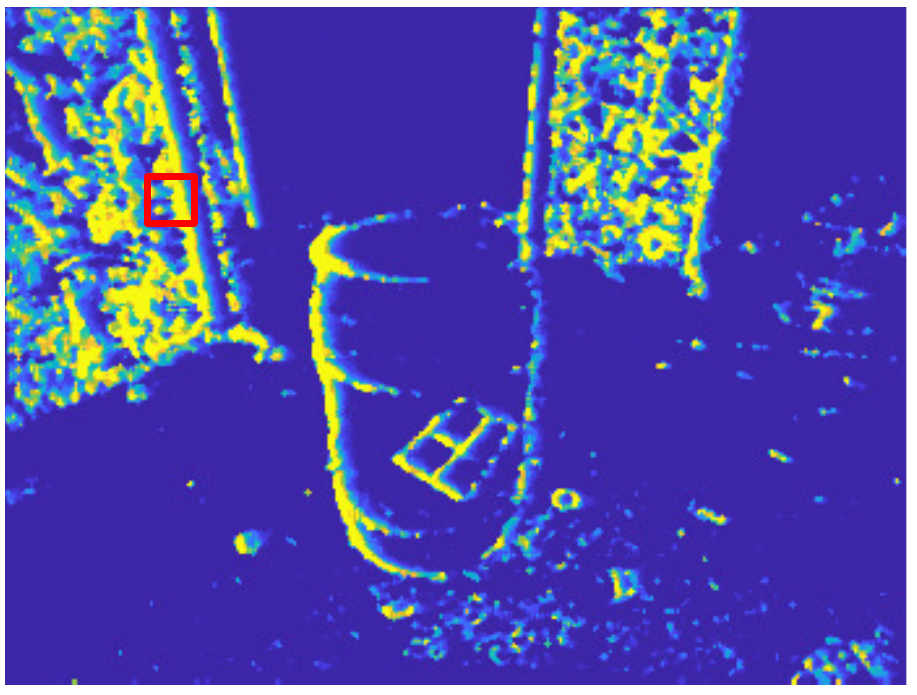}
  \label{fig:mapping:objective:time-surface-left}}
  \subfigure[t][\small{Time surface (right DVS).}]{
  \includegraphics[width=0.46\columnwidth]{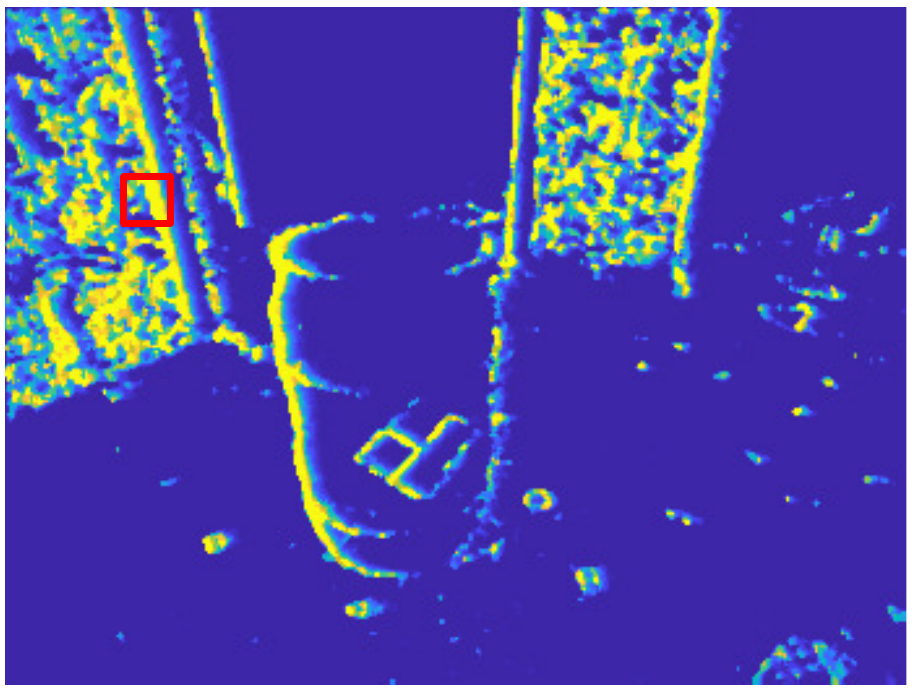}
  \label{fig:mapping:objective:time-surface-right}}
  \caption{\emph{Mapping. Spatio-temporal consistency}.
  (a) An intensity frame shows the visual appearance of the scene. 
  Our method does not use intensity frames; only events.
  (b) The objective function measures the inconsistency between the motion history content (time surfaces (c) and (d)) across left-right retinas, thus replacing the photometric error in frame-based stereo.
  Specifically, (b) depicts the variation of $C(\bx, \rho, \cT_{\tl}(\cdot,t), \cT_{\tr}(\cdot,t), \bT_{t-\delta t : t})$ with inverse depth $\rho$. 
  The vertical dashed line (black) indicates the ground truth inverse depth. 
  (c)-(d) show the time surfaces of the stereo event camera at the observation time, $\cT_{\tl}(\cdot,t)$, $\cT_{\tr}(\cdot,t)$, 
  where the pixels for measuring the temporal residual in (b) are enclosed in red.}
  \label{fig:mapping:objective}
\end{figure}

Fig.~\ref{fig:mapping:objective} shows an example of the objective function from a real stereo event-camera sequence~\cite{Zhu18ral} that has ground truth depth.
It confirms that the proposed objective function~\eqref{eq:objective-function} does lead to the optimal depth for a generic event.
It visualizes the profile of the objective function for the given event (Fig.~\ref{fig:mapping:objective:energy}) 
and the stereo observation used (Figs.~\ref{fig:mapping:objective:time-surface-left} and~\ref{fig:mapping:objective:time-surface-right}).

\paragraph*{Remark on Modeling Data Association}
Note that our approach differs from classical two-step event-processing methods~\cite{Kogler11isvc,Rogister12tnnls,CamunasMesa14fns,Ieng18fnins} that solve the stereo matching problem first and then triangulate the 3D point.
Such two-step approaches work in a ``back-projection'' fashion, mapping 2D event measurements into 3D space. 
In contrast, our approach combines matching and triangulation in a single step, operating in a forward-projection manner (3D$\rightarrow$2D). 
As shown in Fig.~\ref{fig:geometry}, an inverse depth hypothesis $\rho$ yields a 3D point, $\pi^{-1}(\bx,\rho)$, whose projection on both stereo image planes at time $t$ gives points $\bx_1(\rho)$ and $\bx_2(\rho)$ whose neighborhoods are compared in the objective function~\eqref{eq:objective-function}. 
Hence, an inverse depth hypothesis $\rho$ establishes a candidate stereo event match, 
and the best match is provided by the $\rho$ that minimizes the objective.

\subsubsection{Non-Linear Solver for Depth Estimation}
\label{sec:mapping:problemformulation}
\label{sec:mapping:nnls}
The proposed objective function \eqref{eq:optimal-inv-depth}-\eqref{eq:objective-function} is optimized using non-linear least squares methods, such as the Gauss-Newton method, which iteratively find the root of the necessary optimality condition
\begin{equation}
\frac{\partial C}{\partial \rho} = 2 \mJ^\top \br = 0,
\label{eq:newton-method}
\end{equation}
where $\br \doteq (r_1, r_2, ..., r_{N^2})^\top$, $N^2$ is the size of the patch, and $\mJ = \partial \br / \partial \rho$.
Substituting the linearization of $\br$ %
given by Taylor's formula,
$\br(\rho+\Delta\rho) \approx \br(\rho) + \mJ(\rho)\Delta\rho$,
we arrive at the normal equation $\mJ^{\top}\mJ\Delta\rho = -\mJ^{\top}\br$, 
where $\mJ^{\top}\mJ = \|\mJ\|^2$ and we omitted the dependency of $\mJ$ and $\br$ with $\rho$ for succinctness.
The inverse depth solution $\rho$ is iteratively updated by
\begin{equation}
\rho \leftarrow \rho + \Delta\rho
\quad\text{with}\quad
\Delta \rho = -(\mJ^{\top}\br) / \|\mJ\|^2.
\label{eq:update-rho}
\end{equation}
Analytical derivatives are used to speed up computations. %

\subsubsection{Initialization of the Non-Linear Solver}
\label{sec:mapping:nnls-initialization}
Successful convergence of the inverse depth estimator~\eqref{eq:update-rho} relies on a good initial guess $\rho_0$.
For this, instead of carrying out an exhaustive search over an inverse depth grid~\cite{Zhou18eccv}, 
we apply a more efficient strategy exploiting the canonical stereo configuration: 
block matching along epipolar lines of the stereo observation $\bigl(\cT_{\tl}(\cdot,t), \cT_{\tr}(\cdot,t)\bigr)$ using an integer-pixel disparity grid. 
That is, we maximize the Zero-Normalized Cross-Correlation (ZNCC) 
using patch centers $\bx'_1=\bx$ (pixel coordinates of event $e_{t-\epsilon}$) and $\bx'_2=\bx'_1+(d,0)^\top$, where $d$ is the disparity, 
as an approximation to the true patch centers \eqref{eq:corresponding-image-points}.
Note that temporal consistency is slightly violated here because the relative motion ${}^{c_t}\bT_{c_{t-\epsilon}}$ in \eqref{eq:corresponding-image-points}, corresponding to the time of the event, $t-\epsilon$, is not compensated for in $\bx'_1,\bx'_2$.
Nevertheless, this approximation provides a reasonable and efficient initial guess $\rho$ (using $d$) whose temporal consistency is refined in the subsequent nonlinear optimization.

\subsubsection{Summary}
Inverse depth estimation for a given event on the left camera is summarized in Algorithm~\ref{alg:mapping-algorihtm}. 
The inputs of the algorithm are: 
the event $e_{t-\epsilon}$ (space-time coordinates), 
a stereo observation (time surfaces at time $t$) $\cT_\text{left/right}(\cdot,t)$, 
the incremental motion ${}^{c_t}\bT_{c_{t-\epsilon}}$ of the stereo rig between the times of the event and the stereo observation,
and the constant extrinsic parameters between both event cameras, ${}^{\tr}\bT_{\tl}$.
The inverse depth of each event considered is estimated independently; thus computations are parallelizable. 
\begin{algorithm}[t]
\caption{Inverse Depth Estimation}
\label{alg:mapping-algorihtm}
\begin{algorithmic}[1]
\State \textbf{Input}: event $e_{t-\epsilon}$, stereo event observation $\cT_\tl(\cdot,t)$, $\cT_\tr(\cdot,t)$ and relative transformation ${}^{c_t}\bT_{c_{t-\epsilon}}$.
    	\State Initialize $\rho$: ZNCC-block matching on $\cT_\tl(\cdot,t)$, $\cT_\tr(\cdot,t)$. %
    	\While{not converged}
        	\State Compute residuals $\br(\rho)$ in~\eqref{eq:mapping-residual}.
            \State Compute Jacobian $\mathbf{J}(\rho)$ (analytical derivatives).
            \State Update: $\rho \leftarrow \rho + \Delta\rho$, using~\eqref{eq:update-rho}.
        \EndWhile
	\State \textbf{return} Converged inverse depth $\rho$ (i.e., $\rho^\star$ in Fig.~\ref{fig:geometry}).
\end{algorithmic}
\end{algorithm}

\ifclearsectionlook\cleardoublepage\fi \subsection{Semi-Dense Reconstruction}
\label{sec:depthfusion}

The 3D reconstruction method presented in Section~\ref{sec:mapping:inverse-depth-estimation} (Algorithm~\ref{alg:mapping-algorihtm}) produces inverse depth estimates for individual events, 
and according to the parametrization (Fig.~\ref{fig:geometry}), each estimate has a different timestamp.
This section develops a probabilistic approach for fusion of inverse depth estimates 
to produce a semi-dense depth map at the current time (Fig.~\ref{fig:mapping:overall-pipeline}), which is later used for tracking.
Depth fusion is crucial since it allows us to refer all depth estimates to a common time, 
reduces uncertainty of the estimated 3D structure and improves density of the reconstruction.
In the following, we first study the probabilistic characteristics of inverse depth estimates (Section~\ref{sec:prob-model-inverse-depth}).
Based on these characteristics, the fusion strategy is presented and incrementally applied as depth estimates on new stereo observations are obtained (Sections~\ref{sec:inv-depth-filters} and \ref{sec:prob-inv-depth-fusion}).
Our fused reconstruction approaches a semi-dense level, producing depth values for most edge pixels.

\subsubsection{Probabilistic Model of Estimated Inverse Depth}
\label{sec:prob-model-inverse-depth}
We model inverse depth at a pixel on the reference view not with a number $\rho$ but with an actual probability distribution. 
Algorithm~\ref{alg:mapping-algorihtm} provides an ``average'' value $\rho^{\star}$ (also in \eqref{eq:optimal-inv-depth}). 
We now present how uncertainty (i.e., spread around the average) is propagated
and carry out an empirical study to determine the distribution of inverse depth.

In the last iteration of Gauss-Newton's method~\eqref{eq:update-rho}, the inverse depth is updated by
\begin{equation}
\rhos \leftarrow \rho + \Delta \rho(\br),
\label{eq:uncertainty-function}
\end{equation}
where $\Delta \rho$ is a function of the residuals \eqref{eq:mapping-residual} $\br$.
Using events, ground truth depth and poses from two datasets we computed a large number of residuals~\eqref{eq:mapping-residual} to empirically determine their probabilistic model.
Fig.~\ref{fig:temporal-residual-distribution} shows the resulting histogram of the residuals $r$ together with a fitted parametric model.
In the experiment, we found that a \studt{} distribution fits the histogram well. %
The resulting probabilistic model of $r$ is denoted by $r \sim  St(\mu_r, s_r^{2}, \nu_r)$, where $\mu_r, s_r, \nu_r$ are the model parameters, namely the mean, scale and degree of freedom, respectively.
The residual histograms in Fig.~\ref{fig:temporal-residual-distribution} seem to be well centered at zero (compared to their spread and to the abscissa range), and so we may set $\mu_r\approx 0$.
Parameters of the fitted \studt{} distributions are given in Table.~\ref{tab:mapping:empirical-study-distribution}
for the two sequences used from two different datasets.

Since Generalised Hyperbolic distributions (GH) are closed under affine transformations and the \studt{} distribution is a particular case of GH, 
we conclude that the affine transformation $\mathbf{z} = \mathbf{Ax} + \mathbf{b}$ (with non-singular matrix $\mathbf{A}$ and vector $\mathbf{b}$) 
of a random vector $\mathbf{x} \sim St(\boldsymbol{\mu}, \boldsymbol{S}, \nu)$ that follows a multivariate \studt{} distribution 
(with mean vector $\boldsymbol{\mu}$, scale matrix $\boldsymbol{S}$ and degree of freedom $\nu$),
also follows a \studt{} distribution~\cite{kotz2004multivariate}, in the form
$\mathbf{z} \sim St(\mathbf{A}\boldsymbol{\mu} + \mathbf{b},\, \mathbf{A}\boldsymbol{S}\mathbf{A}^{\top},\, \nu)$.

Applying this theorem to \eqref{eq:update-rho}, 
with $r\sim St(\mu_r,s^2_r,\nu_r)$ and $\mathbf{A}\equiv -\sum_i J_{i}/\|\mJ\|^2$, $\mathbf{b}\equiv\mathbf{0}$, 
we have that the update $\Delta \rho$ approximately follows a \studt{} distribution
\begin{equation}
\Delta \rho \sim St\left(-\frac{\sum J_{i}}{\|\mJ\|^{2}}\mu_{r},\, \frac{s_{r}^{2}}{\|\mJ\|^{2}},\, \nu_{r}\right).
\label{eq:uncertainty-propagation}
\end{equation}
Next, applying the theorem to the affine function~\eqref{eq:uncertainty-function}
and assuming $\mu_r\approx 0$ (Fig.~\ref{fig:temporal-residual-distribution}) 
we obtain the approximate distribution
\begin{equation}
\rho \sim St\left(\rho^{\star},\, \frac{s_{r}^{2}}{\|\mJ\|^{2}},\, \nu_{r}\right),
\label{eq:uncertainty-propagation-rho}
\end{equation}
with $\mJ\equiv\mJ(\rho^{\star})$.
The resulting variance is given by
\begin{equation}
\sigma_{\rhos}^{2} = \frac{\nu_{r}}{\nu_{r}-2}\, \frac{s_{r}^{2}}{\|\mJ\|^{2}}.
\end{equation}

\paragraph*{Robust Estimation}
The obtained probabilistic model can be used for robust inverse depth estimation in the presence of noise and outliers, since the heavy tails of the \studt{} distribution account for them.
To do so, each squared residual in~\eqref{eq:objective-function} is re-weighted by a factor $\omega(r_i)$, which is a function of the probabilistic model $p(r)$.
The resulting optimization problem is solved using the Iteratively Re-weighted Least Squares (IRLS) method, replacing the Gauss-Newton solver in Algorithm~\ref{alg:mapping-algorihtm}.
Details about the derivation of the weighting function are provided in~\cite{Kerl13icra,zhou2018canny}.

\ifshowfigures
\begin{figure}[t!] %
\centering
    \subfigure[t][\small{\ijrrsimulation{} \cite{Mueggler17ijrr}.}]{
    \includegraphics[width=0.46\columnwidth]{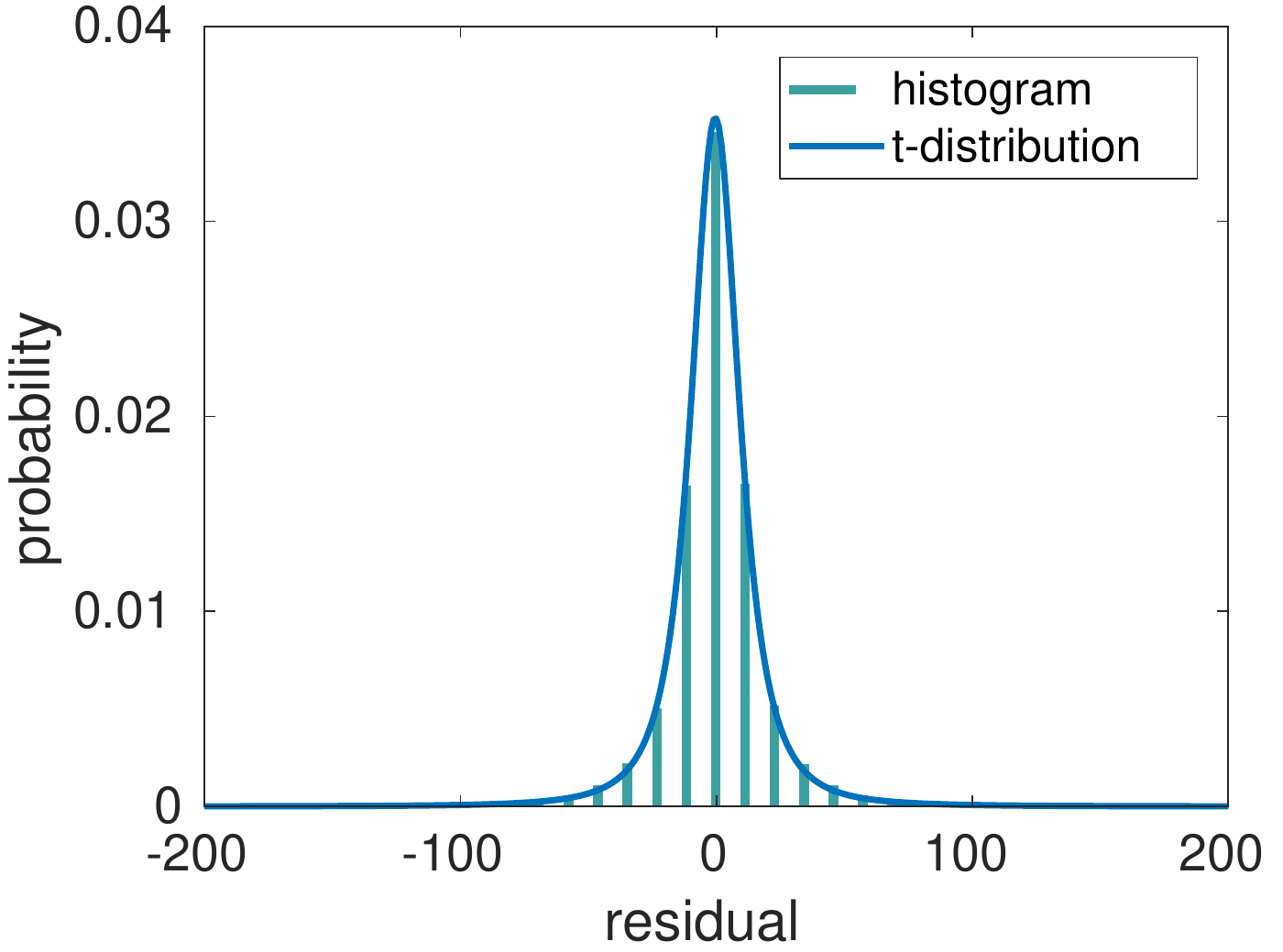}
    \label{fig:temporal-residual-distribution:blender}}\,
    \subfigure[t][\small{\upennflyOne{} \cite{Zhu18ral}.}]{
    \includegraphics[width=0.46\columnwidth]{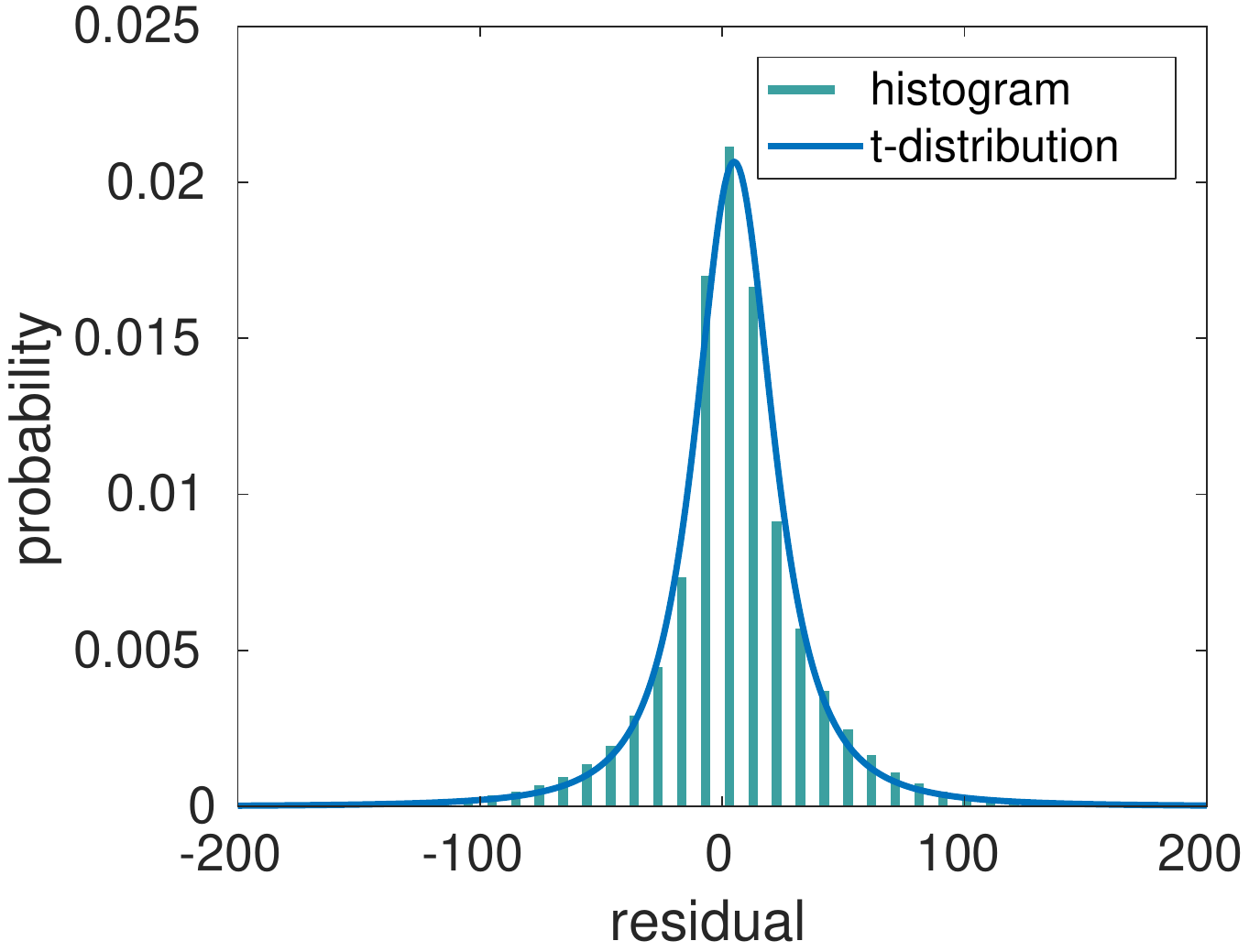}
    \label{fig:temporal-residual-distribution:upenn}}
    \caption{Probability distribution (PDF) of the temporal residuals $r_i$: empirical (green histogram) and \studt{} fit (blue curve).}
    \label{fig:temporal-residual-distribution}
    \captionof{table}{Parameters of the fitted \studt{} distribution.} %
    \vspace{0.5ex}
    \centering %
    \begin{adjustbox}{max width=\linewidth}
    \begin{tabular}{@{}lllll@{}} %
    \toprule
     & Mean ($\mu$) & Scale ($s$) & DoF ($\nu$) & Std. ($\sigma$) \\[0.5ex]
    \midrule
    \ijrrsimulation{} \cite{Mueggler17ijrr} & -0.423 & 10.122 & 2.207 & 33.040 \\[0.6ex]
    \upennflyOne{} \cite{Zhu18ral} & 4.935 & 17.277 & 2.182 & 59.763 \\
    \bottomrule %
    \end{tabular}
    \end{adjustbox}
    \label{tab:mapping:empirical-study-distribution}
\end{figure}

\fi

\subsubsection{Inverse Depth Filters}
\label{sec:inv-depth-filters}
The fusion of inverse depth estimates from several stereo pairs is performed in two steps.
First, inverse depth estimates are propagated from the time of each event to the time of a stereo observation (i.e., the current time).
This is simply done similarly to the uncertainty propagation operation in~\eqref{eq:uncertainty-propagation}-\eqref{eq:uncertainty-propagation-rho}.
Second, the propagated inverse depth estimate is fused (updated) with prior estimates at this pixel coordinate.
The update step is performed using robust Bayesian filter for \studt{} distribution.
A \studt{} filter %
is derived in~\cite{roth2017robust}:
given a prior $St(\mu_a, s_a, \nu_a)$ and a measurement $St(\mu_b, s_b, \nu_b)$, 
the posterior is approximated by a $St(\mu, s, \nu)$ distribution with parameters
\begin{subequations}
\label{eq:t-filter-update}
\begin{align}
    \nu^{\prime} &= \min(\nu_a, \nu_b), \\
    \mu &= \frac{s_{a}^{2}\mu_{b} +s_{b}^{2}\mu_{a}}{s_{a}^{2} + s_{b}^{2}}, \\ 
    s^2 &= \frac{\nu^{\prime} + \frac{(\mu_a - \mu_b)^{2}}{s_{a}^{2} + s_{b}^{2}}}{\nu^{\prime} + 1} \cdot \frac{s_{a}^{2}s_{b}^{2}}{s_{a}^{2} + s_{b}^{2}}, \\
    \nu &= \nu^{\prime} + 1.
\end{align}
\end{subequations}

\ifshowfigures

\begin{figure*}[t!]
  \centering
  \subfigure[Flowchart of mapping module.\label{fig:mapping:overall-pipeline:flowchart}]{\includegraphics[trim={0 0.5cm 11.2cm 0},clip,width=0.5\linewidth]{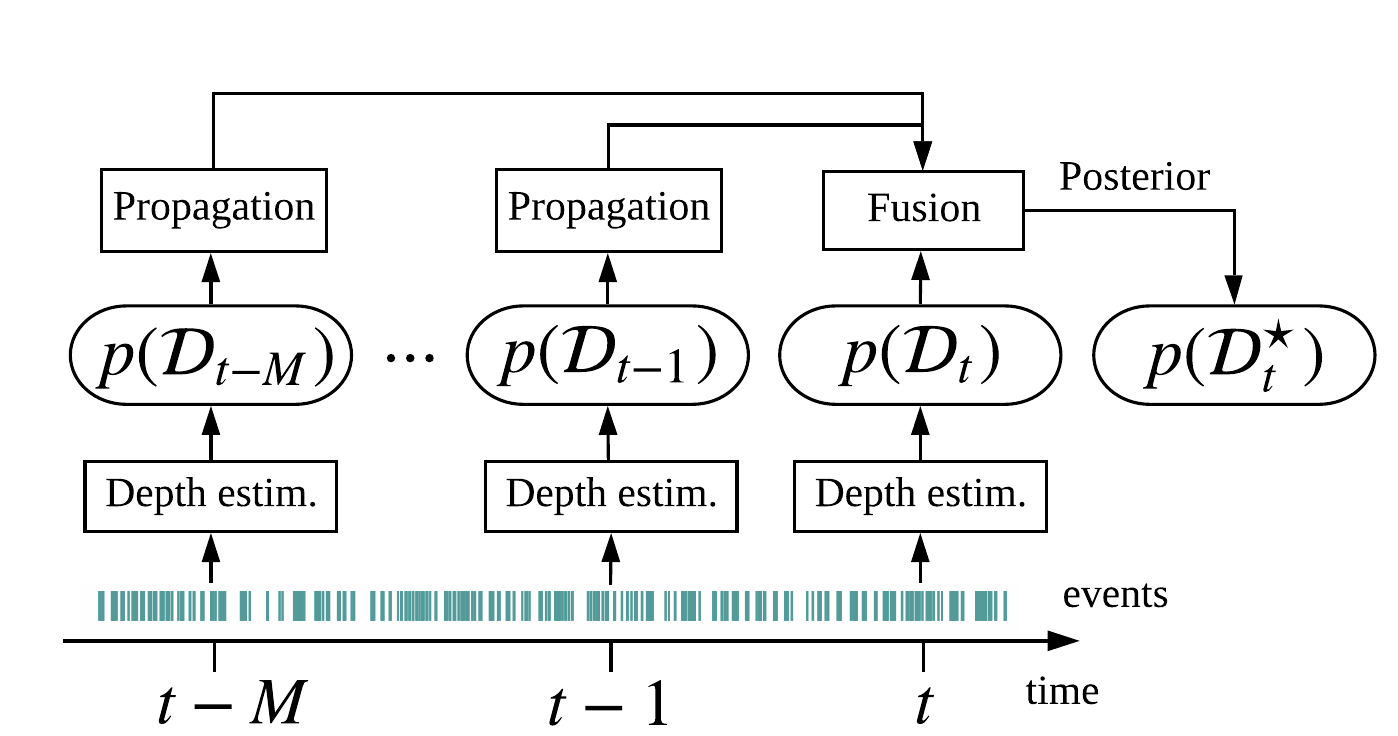}}\;\;\;
  \subfigure[Depth fusion rules at locations on a $3\times 3$ pixel grid.\label{fig:mapping:overall-pipeline:fusion-rules}]{\includegraphics[trim={14cm 0.5cm 0 0},clip,width=0.4\linewidth]{pic/fusion/EPTAM_fusion_strategy_new_flowchart.pdf}}
\caption{\emph{Mapping module}:
(\textbf{a}) Stereo observations (time surfaces) are created at selected timestamps $t, \cdots, t-M$ (\eg, \SI{20}{\Hz}) and fed to the mapping module along with the events and camera poses.
Inverse depth estimates, represented by probability distributions $p(\mathcal{D}_{t-k})$, are propagated to a common time $t$ and fused to produce an inverse depth map, $p(\mathcal{D}^{\star}_{t})$.
We fuse estimates from 20 stereo observations (i.e., $M=19$) to create $p(\mathcal{D}^{\star}_{t})$.
(\textbf{b}) Taking the fusion from $t-1$ to $t$ as an example, the fusion rules are indicated in the dashed rectangle, which represents a $3\times 3$ region of the image plane (pixels are marked by a grid of gray dots).
A 3D point corresponding to the mean depth of $p(\mathcal{D}_{t-1})$ projects on the image plane at time $t$ at a blue dot. 
Such a blue dot and $p(\mathcal{D}_{t-1})$ influence (i.e., assign, fuse or replace) the distributions $p(\mathcal{D}^{\star}_{t})$ estimated at the four closest pixels.
}
\label{fig:mapping:overall-pipeline}
\end{figure*}
\fi

\subsubsection{Probabilistic Inverse Depth Fusion}
\label{sec:prob-inv-depth-fusion}
Assuming the propagated inverse depth follows a distribution $St(\mu_a, s_{a}^2, \nu_a)$, its corresponding location in the target image plane is typically a non-integer coordinate $\bx^{\text{float}}$. 
Hence, the propagated inverse depth will have an effect on the distributions at the four nearest pixel locations $\{\bx_{j}^{\text{int}}\}_{j=1}^{4}$ (see Fig.~\ref{fig:mapping:overall-pipeline:fusion-rules}). 
Using $\bx_{1}^{\text{int}}$ as an example, the fusion is performed based on the following rules:
\begin{enumerate}
\item If no previous distribution exists at $\bx_{1}^{\text{int}}$, initialize it with $St(\mu_a, s_{a}^2, \nu_a)$. 
\item If there is already an inverse depth distribution at $\bx_{1}^{\text{int}}$, e.g., $St(\mu_b, s_{b}^2, \nu_b)$, the compatibility between the two inverse depth hypotheses is checked to decide whether they may be fused. 
The compatibility of two hypotheses $\rho_a, \rho_b$ is evaluated by checking
\begin{equation}
    \mu_b - 2 \sigma_b \, \leq \mu_a \leq \, \mu_b + 2 \sigma_b,
\label{eq:consistency criterion}
\end{equation}
where $\sigma_b = s_b \sqrt{\nu_b/(\nu_b-2)}$.
If the two hypotheses are compatible, they are fused into a single inverse depth distribution using \eqref{eq:t-filter-update}, otherwise the distribution with the smallest variance remains.
\end{enumerate}
The fusion strategy is illustrated in the dashed rectangle of Fig.~\ref{fig:mapping:overall-pipeline:fusion-rules} using as example the propagation and update from estimates of $\cD_{t-1}$ to the inverse depth map $\cD_t$.

\subsubsection{Summary}
Together with the inverse depth estimation introduced in Sec.~\ref{sec:mapping:inverse-depth-estimation}, the overall mapping procedure is illustrated in Fig~\ref{fig:mapping:overall-pipeline}.
The inverse depth estimation at a given timestamp $t$, using the stereo observation $\cT_\text{left/right}(\cdot,t)$ and involved events as input, is tackled via nonlinear optimization (IRLS).
Probabilistic estimates at different timestamps are propagated and fused to the inverse depth map distribution at the most recent timestamp $t$, $p(\cD^{\star}_t)$.
The proposed fusion leads to a semi-dense inverse depth map $\cD^{\star}_t$ with reasonably good signal-noise ratio, which is required by the tracking method discussed in the following Section.

\paragraph*{Remarks}
All events are involved in creating time surfaces, which are used for tracking and mapping. 
However, depth is not estimated for every event because it is expensive and we aim at achieving real-time operation with limited computational resources (see Section~\ref{sec:experiments:timing}).

The number of fused stereo observations, $M+1=20$ in Fig.~\ref{fig:mapping:overall-pipeline}, was determined empirically as a sensible choice for having a good density of the semi-dense depth map in most sequences tested (Section~\ref{sec:experiments}). 
A more theoretical approach would be to have an adaptive number based on statistical criteria, such as the apparent density of points or the decrease of uncertainty in the fused depth, but this is left as future work.

\ifclearsectionlook\cleardoublepage\fi \section{Camera Tracking}
\label{sec:tracking}
Let us now present the tracking module in Fig.~\ref{fig:system-overview}, which takes events and a local map as input and computes the pose of the stereo rig with respect to the map.
In principle, each event has a different timestamp and hence also a different camera pose~\cite{Mueggler18tro} as the stereo rig moves. 
Since it is typically not necessary to compute poses with microsecond resolution, we consider the pose of a stereo observation (\ie, time surfaces).

Two approaches are now considered before presenting our solution.
($i$) Assuming a semi-dense inverse depth map is available in a reference frame and 
a subsequent stereo observation is temporally (and thus spatially) close to the reference frame, 
the relative pose (between the reference frame and the stereo observation) could be characterized as being the one that, 
transferring the depth map to both left and right frames of the stereo observation, yields minimal spatio-temporal inconsistency.
However, this characterization is only a necessary condition for solving the tracking problem rather than a sufficient one. 
The reason is that a wrong relative pose might transfer the semi-dense depth map to the ``blank'' regions of both left and right time surfaces, which would produce an undesired minimum.
($ii$) An alternative approach to the spatio-temporal consistency criterion would be to consider only the left time surface of the stereo observation (since the right camera is rigidly attached) and use the edge-map alignment method from the monocular system~\cite{Rebecq17ral}. 
However, this requires the creation of additional event images.

Instead, our solution consists of taking full advantage of the time surfaces already defined for mapping.
To this end we present a novel tracking method based on global image-like registration using time surface ``negatives''.
It is inspired by an edge-alignment method for RGB-D cameras using distance fields~\cite{zhou2018canny}.
In the following, we intuitively and formally define the tracking problem (Sections \ref{sec:tracking:inspiration} and \ref{sec:tracking:problem-statement}), 
and solve it using the forward compositional Lucas-Kanade method~\cite{Baker04ijcv} (Section \ref{sec:tracking:compositional-method}).
Finally, we show how to improve tracking robustness while maintaining a high throughput (Section \ref{sec:tracking:robust}).

\subsection{Exploiting Time Surfaces as Distance Fields}
\label{sec:tracking:inspiration}

Time surfaces (TS) (Section~\ref{sec:timesurfaces}) encode the motion history of the edges in the scene.
Large values of the TS~\eqref{eq:time-surface-map} correspond to recently triggered events, i.e., the current location of the edge.
Typically those large values have a ramp on one side (signaling the previous locations of the edge) and a ``cliff'' on the other one.
This can be interpreted as an anisotropic distance field: following the ramp, one may smoothly reach the current location of the edge.
Indeed, defining the ``negative'' (as in image processing) of a TS $\cT(\bx,t)$ by
\begin{equation}
    \bar{\cT}(\bx,t) = 1 - \cT(\bx,t),
\label{eq:negative-time-surface}
\end{equation}
allows us to interpret the small values as the current edge location and the ramps as a distance field to the edge.
This negative transformation also allows us to formulate the registration problem as a minimization one rather than a maximization one.
Like the TS, \eqref{eq:negative-time-surface} is rescaled to the range $[0,255]$.

\begin{figure}[t]
  \centering
  \subfigure[t][\small{Depth map in the reference viewpoint with known pose.}]{
  \includegraphics[width=0.46\columnwidth]{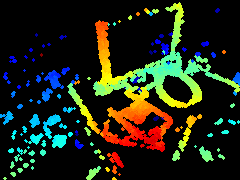}
  \label{fig: ref inverse depth map}}\,
  \subfigure[t][\small{Warped depth map overlaid on the time surface negative at the current time.}]{
  \frame{\includegraphics[width=0.46\columnwidth]{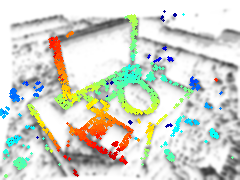}
  \label{fig: registration on negative SAE}}}
  \caption{\emph{Tracking}.
  The point cloud recovered from the inverse depth map in (a) is warped to the time surface negative at the current time (b) using the estimated relative pose. 
  The result (b) is a good alignment between the projection of the point cloud and the minima (dark areas) of the time surface negative.}
  \label{fig:tracking:registration}
\end{figure}

The essence of the proposed tracking method is to align the dark regions of the TS negative and the support of the inverse depth map when warped to the TS frame by a candidate pose. 
Thus the method is posed as an image-like alignment method, with the images representing time information and the scene edges being represented by ``zero time''.
A successful tracking example showing edge-map alignment is given in Fig.~\ref{fig: registration on negative SAE}.
Building on the findings of semi-dense direct tracking for frame-based cameras~\cite{Engel14eccv}, we only use the left TS for tracking because incorporating the right TS does not significantly increase accuracy while it doubles the computational cost.

\subsection{Tracking Problem Statement}
\label{sec:tracking:problem-statement}
More specifically, the problem is formulated as follows. 
Let $\cS^{\cFref} = \{ \bx_i \}$ be a set of pixel locations with valid inverse depth $\rho_i$ in the reference frame $\cFref$ (i.e., the support of the semi-dense depth map $\cD^{\cFref}\equiv \cD^{\star}$). 
Assuming the TS negative at time $k$ is available, denoted by $\bar{\cT}_{\tl}(\cdot, k)$, the goal is to find the pose $T$ such that the support of the warped semi-dense map $T(\cS^{\cFref})$ aligns well with the minima of $\bar{\cT}_{\tl}(\cdot, k)$, as shown in Fig.~\ref{fig:tracking:registration}.
The overall objective of the registration is to find
\begin{equation}
    \btheta^{\star} = \argmin\limits_{\btheta} \sum_{\bx \in \cS^{\cFref}} 
    \bigl(\bar{\cT}_{\tl}(W(\bx, \rho; \btheta), k) \bigr)^{2},
\label{eq:tracking-objective}
\end{equation}
where the warping function
\begin{equation}
W(\bx, \rho; \btheta) \doteq \pi_{\tl}(T(\pi_{\text{ref}}^{-1}(\bx, \rho), G(\btheta))),
\label{eq:warp-function}
\end{equation}
transfers points from $\cFref$ to the current frame. 
It consists of a chain of transformations: back-projection from $\cFref$ into 3D space given the inverse depth, change of coordinates in space (using candidate motion parameters), and perspective projection onto the current frame.
The function $G(\btheta): \mathbb{R}^{6} \to \text{SE}(3)$ gives the transformation matrix corresponding to the motion parameters $\btheta \doteq (\bc^\top,\bt^\top)^\top$, where $\bc = (c_1, c_2, c_3)^\top$ are the Cayley parameters~\cite{cayleyparameter} for orientation $\Rot$, and $\bt = (t_x, t_y, t_z)^\top$ is the translation.
The function $\pi_{\text{ref}}^{-1}(\cdot)$ back-projects a pixel $\bx$ into space using the known inverse depth $\rho$, while $\pi_{\tl}(\cdot)$ projects the transformed space point onto the image plane of the left camera.
$T(\cdot)$ performs a change of coordinates, transforming the 3D point with motion $G(\btheta)$ from $\cFref$ to the left frame ${\cF_{k}}$ of the current stereo observation (time $k$).
We assume rectified and undistorted stereo configuration, which simplifies the operations by using homogeneous coordinates.

\subsection{Compositional Algorithm}
\label{sec:tracking:compositional-method}
\begin{figure}[t]
  \centering
  \subfigure[t][\small{Objective w.r.t $c_1$.}]{
  \includegraphics[width=0.14\textwidth]{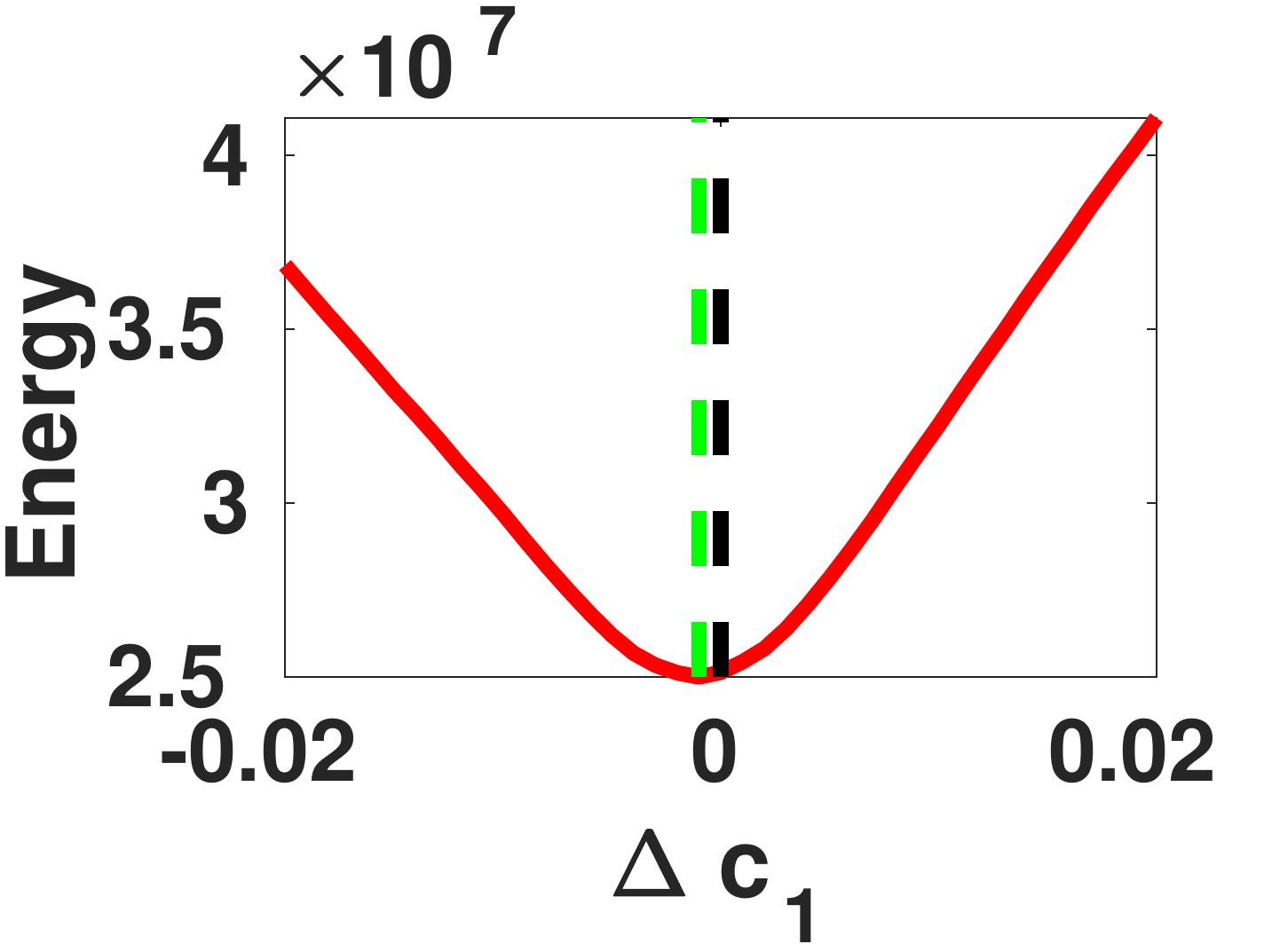}
  \label{fig: Objective function w.r.t $c_1$}}\,
  \subfigure[t][\small{Objective w.r.t $c_2$.}]{
  \includegraphics[width=0.14\textwidth]{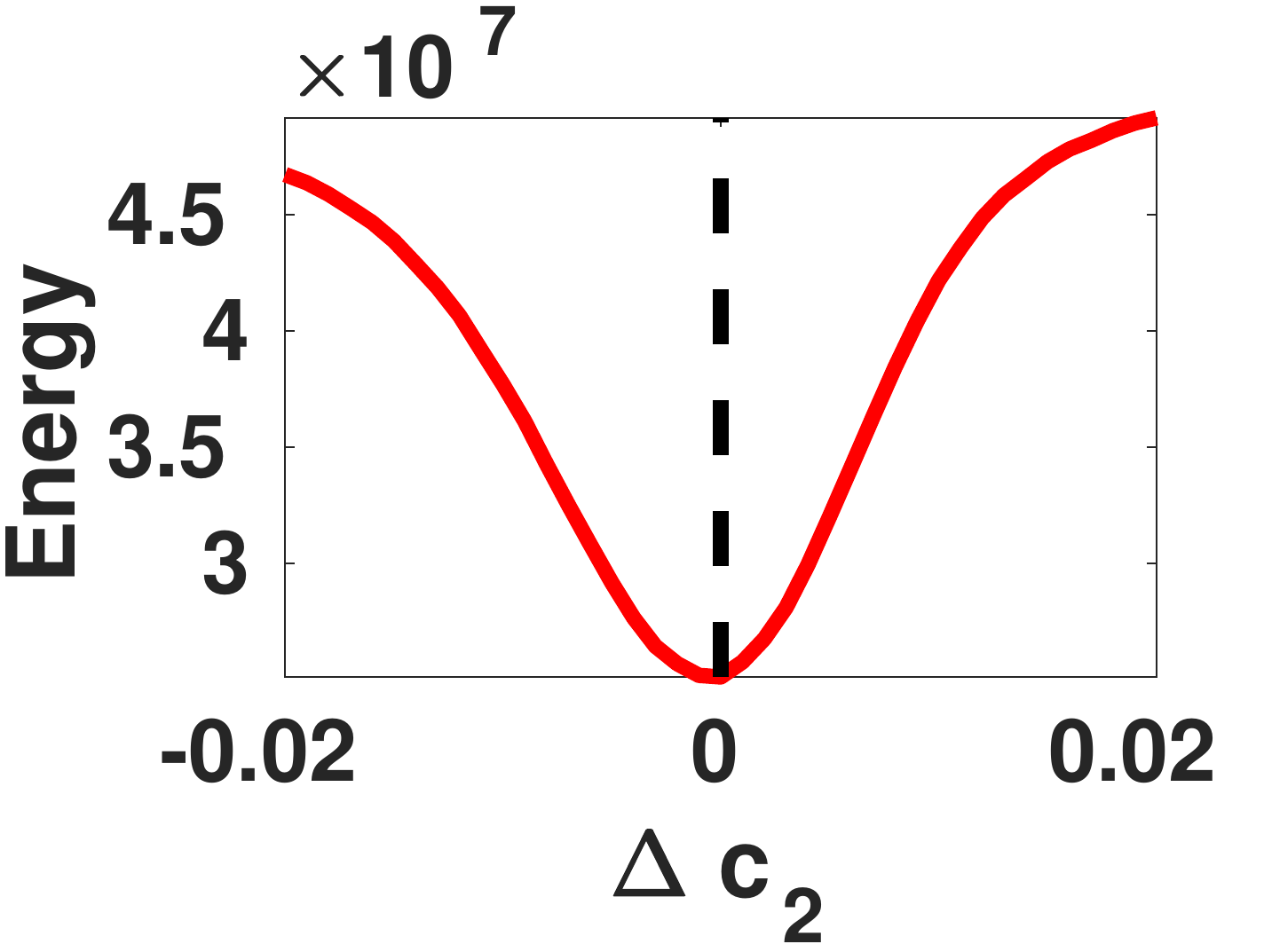}
  \label{fig: Objective function w.r.t $c_2$}}\,
  \subfigure[t][\small{Objective w.r.t $c_3$.}]{
  \includegraphics[width=0.14\textwidth]{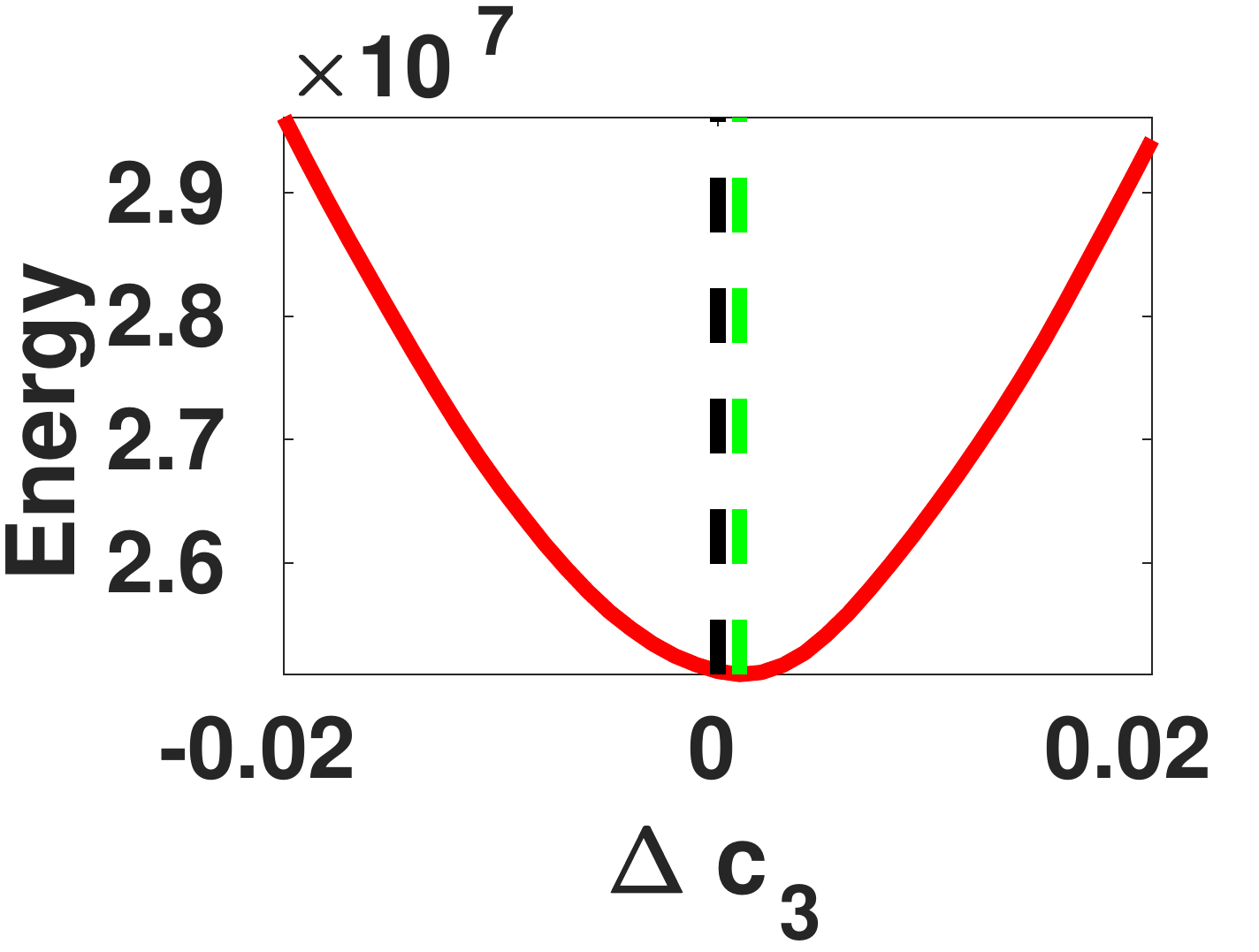}
  \label{fig: Objective function w.r.t $c_3$}}\,
  \subfigure[t][\small{Objective w.r.t $t_x$.}]{
  \includegraphics[width=0.14\textwidth]{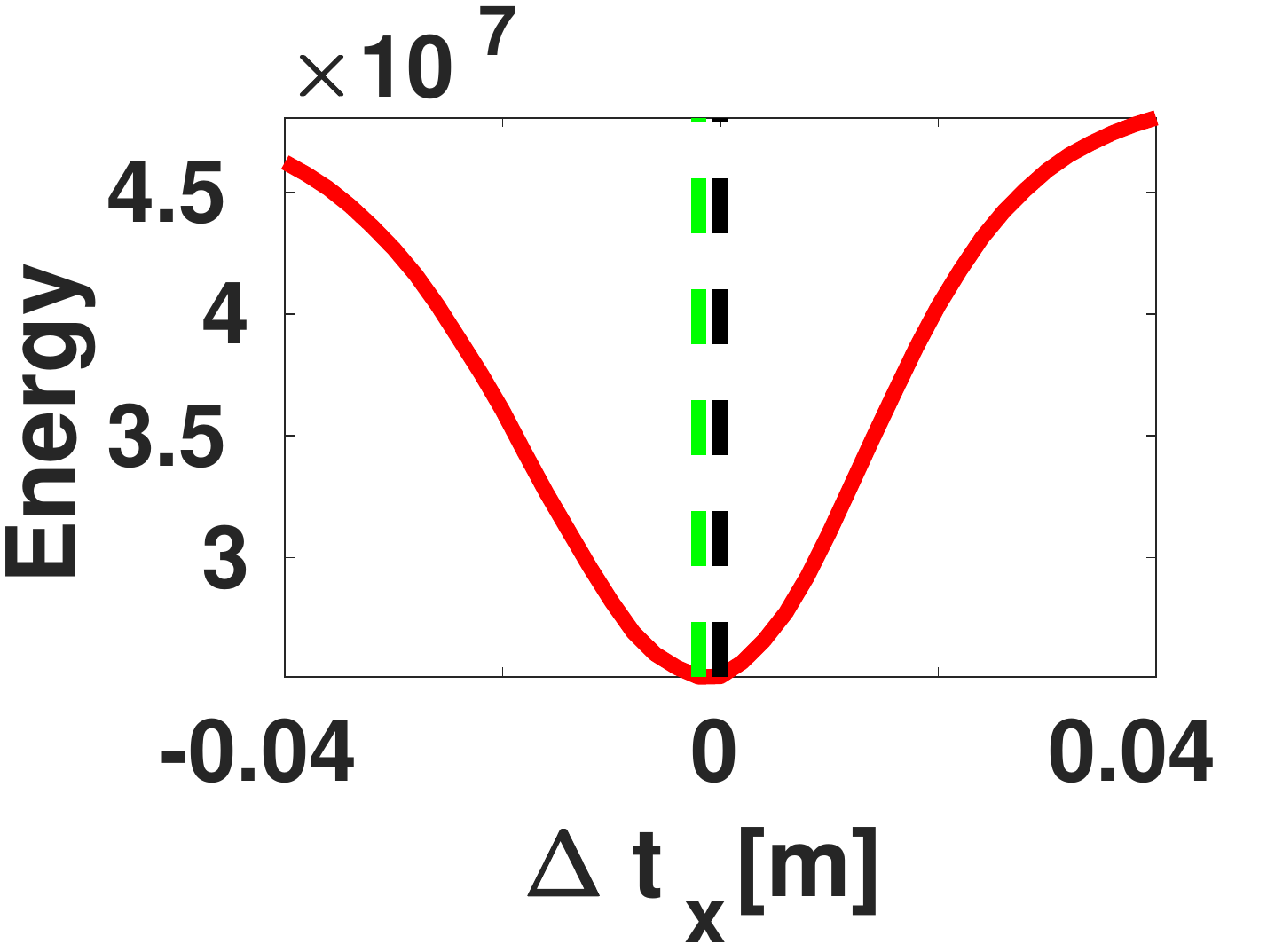}
  \label{fig: Objective function w.r.t $t_x$}}\,
  \subfigure[t][\small{Objective w.r.t $t_y$.}]{
  \includegraphics[width=0.14\textwidth]{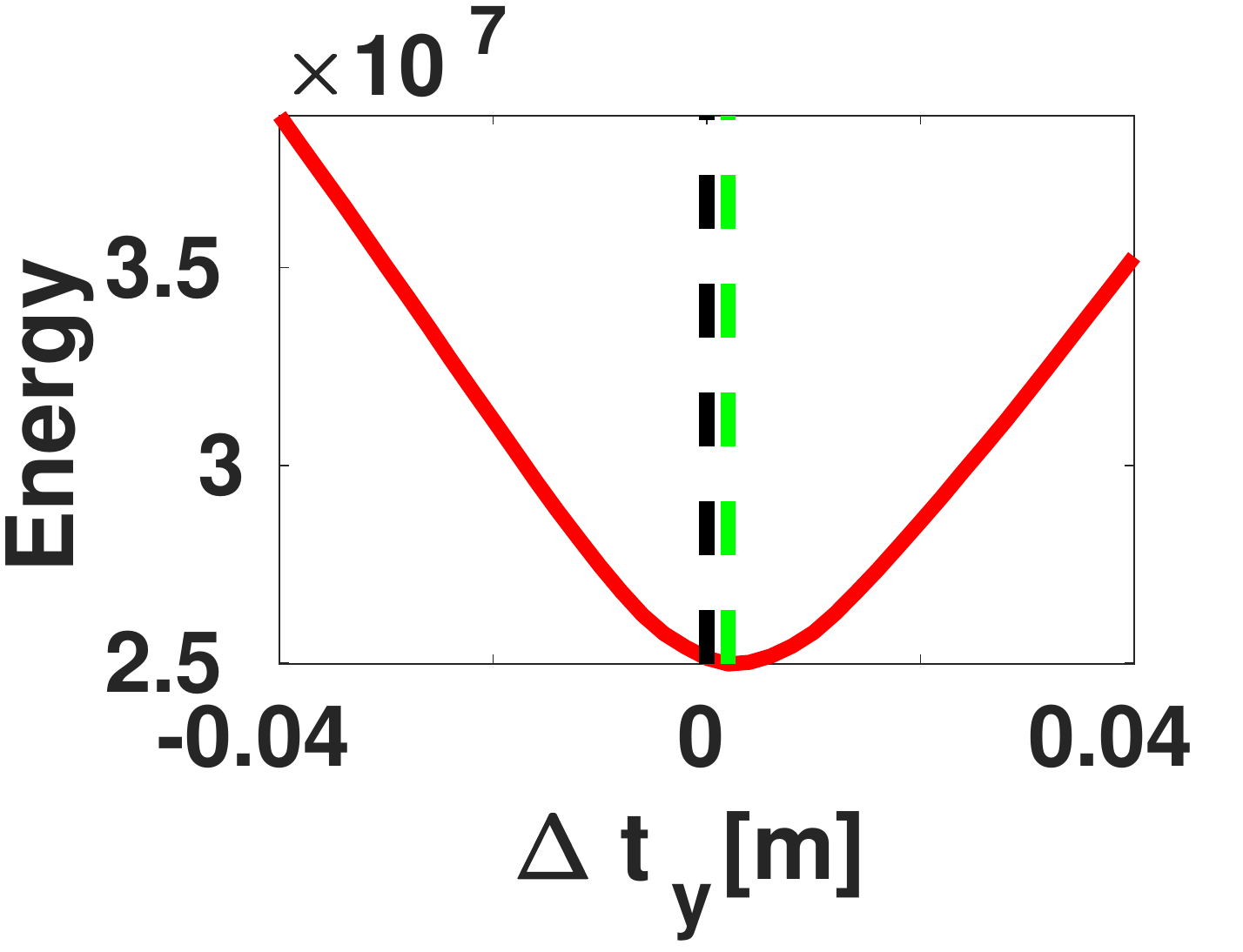}
  \label{fig: Objective function w.r.t $t_y$}}\,
  \subfigure[t][\small{Objective w.r.t $t_z$.}]{
  \includegraphics[width=0.14\textwidth]{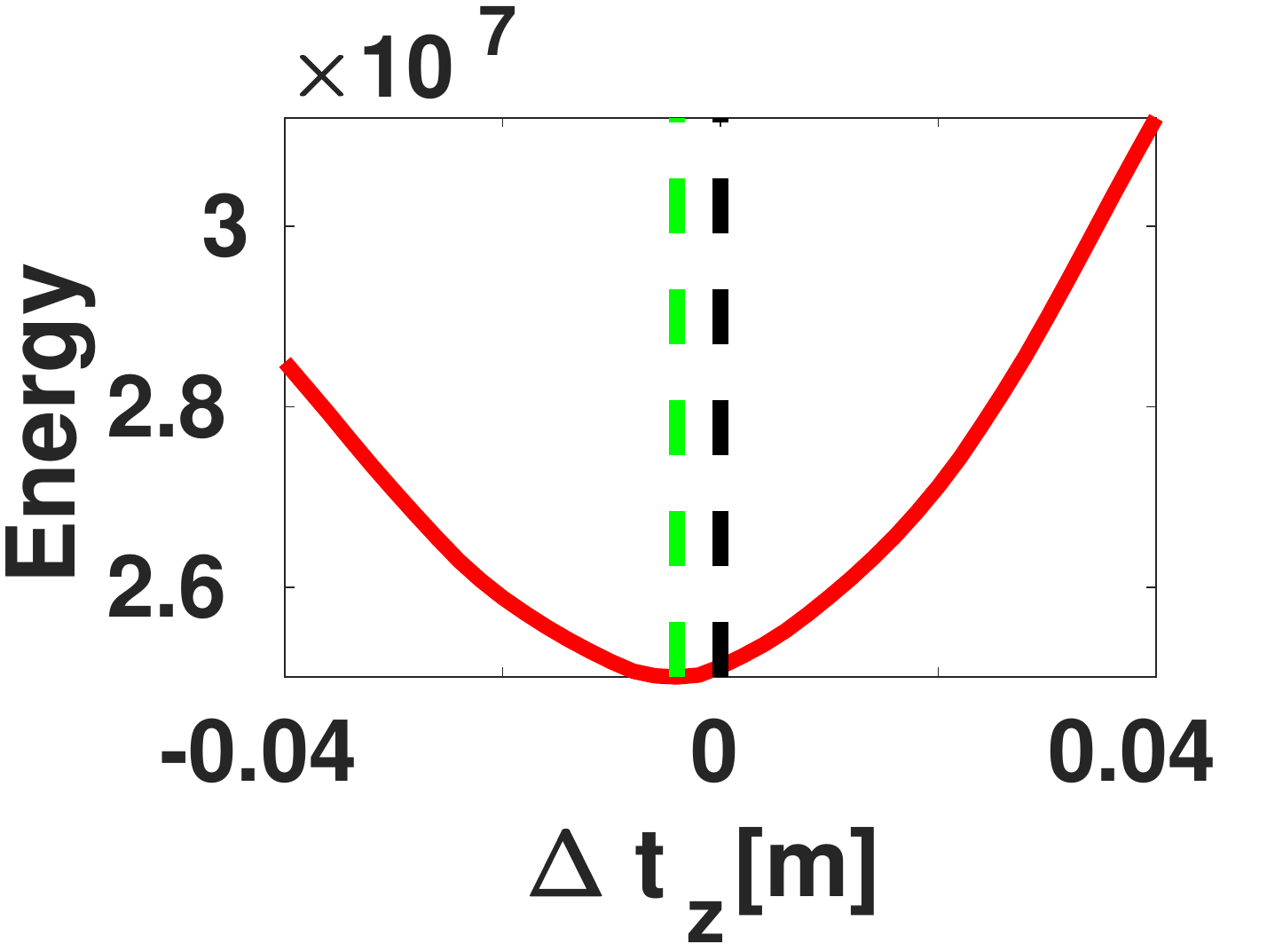}
  \label{fig: Objective function w.r.t $t_z$}}
  \caption{\emph{Tracking}. Slices of the objective function \eqref{eq:tracking-objective}. 
  Plots (a)-(c) and (d)-(f) show the variation of the objective function with respect to each DoF in orientation and translation, respectively.
  The vertical black dashed line indicates the ground truth pose, while the green one depicts the function's minimizer.}
  \label{fig:tracking:one-dim-slices-dofs}
\end{figure}

We reformulate the problem~\eqref{eq:tracking-objective} using the forward compositional Lucas-Kanade method~\cite{Baker04ijcv}, which iteratively refines the incremental pose parameters.
It minimizes
\begin{align}
    F(\Delta \btheta) \doteq
    \sum_{\bx \in \cS^{\cFref}} 
    \bigl(\bar{\cT}_{\tl}(W(W(\bx, \rho; \Delta \btheta);\btheta), k)\bigr)^{2},
\label{eq:tracking:compositional-algorithm}
\end{align}
with respect to $\Delta \btheta$ in each iteration and then updates the estimate of the warp as:
\begin{equation}
    W(\bx, \rho; \btheta) \leftarrow W(\bx, \rho; \btheta) \,\circ\, W(\bx, \rho; \Delta \btheta).
\label{eq:warping update}
\end{equation}
The compositional approach is more efficient than the additive method~\eqref{eq:tracking-objective} because some parts of the Jacobian remain constant throughout the iteration and can be precomputed.
This is due to the fact that linearization is always performed at the position of zero increment. 
As an example, Fig.~\ref{fig:tracking:one-dim-slices-dofs} shows slices of the objective function with respect to each degree of freedom of $\btheta$, evaluated around ground-truth relative pose $\Delta \btheta = \bzero$.
It is clear that the objective function formulated using the compositional method is smooth, differentiable and has unique local optimum near the ground truth.
To enlarge the width of the convergence basin, a Gaussian blur (kernel size of 5 pixels) is applied to the TS negative.

\subsection{Robust and Efficient Motion Estimation}
\label{sec:tracking:robust}
As far as we have observed, the non-linear least-squares solver is already accurate enough. 
However, to improve robustness in the presence of noise and outliers in the inverse depth map, a robust norm is considered. 
For efficiency, the Huber norm is applied and the iteratively reweighted least squares (IRLS) method is used to solve the resulting problem.

To speed up the optimization, we solve the problem using the Levenberg-Marquardt (LM) method with stochastic sampling strategy (as in~\cite{Rebecq17ral}).
At each iteration, only a batch of $N_p$ 3D points are randomly picked in the reference frame and used for evaluating the objective function (typically $N_p=300$).
The LM method can deal with the non-negativeness of the residual $\bar{\cT}_{\tl}(\cdot,k)$ and it is run only one iteration per batch.
We find that five iterations are often enough for a
successful convergence because the initial pose is typically close to the optimum.

\ifclearsectionlook\cleardoublepage\fi \section{Experiments}
\label{sec:experiments}
Let us now evaluate the proposed event-based stereo VO system.
First we present the datasets and stereo camera rig used as source of event data (Section \ref{sec:experiments:setup}).
Then, we evaluate the performance of the method with two sets of experiments. 

In the first set, we show the effectiveness of the mapping module alone by using ground truth poses provided by an external motion capture system.
We show that the proposed \studt{} probabilistic approach leads to more accurate inverse depth estimates than standard least squares (Section~\ref{sec:experiments:mapping:solvers}), 
and then we compare the proposed mapping method against three stereo 3D reconstruction baselines (Section \ref{sec:experiments:mapping:other-stereo}).

In the second set of experiments, we evaluate the performance of the full system by feeding only events and comparing the estimated camera trajectories against the ground truth ones (Section \ref{sec:experiments:full-system}). 
We further demonstrate the capabilities of our approach to unlock the advantages of event-based cameras in order to perform VO in difficult illumination conditions, such as low light and HDR (Section \ref{sec:experiments:HDR}).
Finally, we analyze the computational performance of the VO system (Section \ref{sec:experiments:timing}), 
discuss its limitations (Section \ref{sec:experiments:missing-edges}),
and motivate research on difficult motions for space-time consistency (Section \ref{sec:experiments:STC Degeneracy Analysis}).

\subsection{Experimental Setup and Datasets Used}
\label{sec:experiments:setup}
\begin{figure}[t!] %
\centering
    \includegraphics[trim={1cm 3.5cm 0 0},clip,width=0.7\columnwidth]{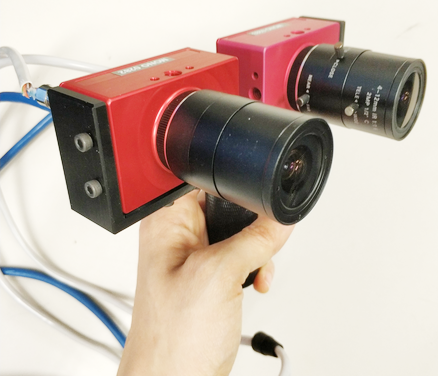}
    \caption{Custom stereo event-camera rig consisting of two DAVIS346 cameras with a horizontal baseline of \SI{7.5}{cm}.}
    \label{fig:stereo-rig}
    \vspace{0.5ex}
    \captionof{table}{Parameters of various stereo event-camera rigs used in the experiments.}
    \vspace{0.5ex}
    \label{tab:stereo-rig-params}
    \begin{adjustbox}{max width=\linewidth}
    \begin{tabular}{@{}lllll@{}}
    \toprule
    Dataset           & Cameras & Resolution (\si{pix}) & Baseline (\si{\centi\meter}) & FOV (\si{\degree}) \\
    \midrule
    \cite{Zhou18eccv} & DAVIS240C & $240 \times 180$ & 14.7 & 62.9 \\[0.5ex]
    \cite{Zhu18ral}   & DAVIS346  & $346 \times 260$ & 10.0 & 74.8  \\[0.5ex]
    \cite{Mueggler17ijrr} & Simulator & $346 \times 260$ & 10.7 & 74.0 \\[0.5ex]
    Ours              & DAVIS346  & $346 \times 260$ & 7.5 & 66.5   \\
    \bottomrule %
    \end{tabular}
    \end{adjustbox}
\end{figure}
To evaluate the proposed stereo VO system we use sequences from publicly available datasets and simulators~\cite{Zhou18eccv,Zhu18ral,Mueggler17ijrr}.
Data provided by \cite{Zhou18eccv} was collected with a hand-held stereo event camera in an indoor environment.
Sequences used from \cite{Zhu18ral} were collected using a stereo event camera mounted on a drone flying in a capacious indoor environment.
The simulator~\cite{Mueggler17ijrr} provides synthetic sequences with simple structure (\eg, front-to-parallel planar structures, geometric primitives, etc.) and an ``ideal'' event camera model.
Besides the above datasets, we collect several sequences using the stereo event-camera rig in Fig.~\ref{fig:stereo-rig}.
The stereo rig consists of two Dynamic and Active Pixel Vision Sensors (DAVIS 346) of $346 \times 260$ pixel resolution, which are calibrated intrinsically and extrinsically. 
The DAVIS comprises a frame camera and an event sensor (DVS) on the same pixel array, 
thus calibration can be done using standard methods on the intensity frames and applied to the events.
Our algorithm works on undistorted and stereo-rectified coordinates, which are precomputed given the camera calibration.
The parameters of the stereo event-camera setup in each dataset used are listed in Table~\ref{tab:stereo-rig-params}.

\subsection{Comparison of Mapping Optimization Criteria: IRLS vs LS}
\label{sec:experiments:mapping:solvers}
\begin{figure}[t!] %
\centering
  \subfigure[b][\small{Standard LS solver.}]{
  \includegraphics[width=0.8\columnwidth]{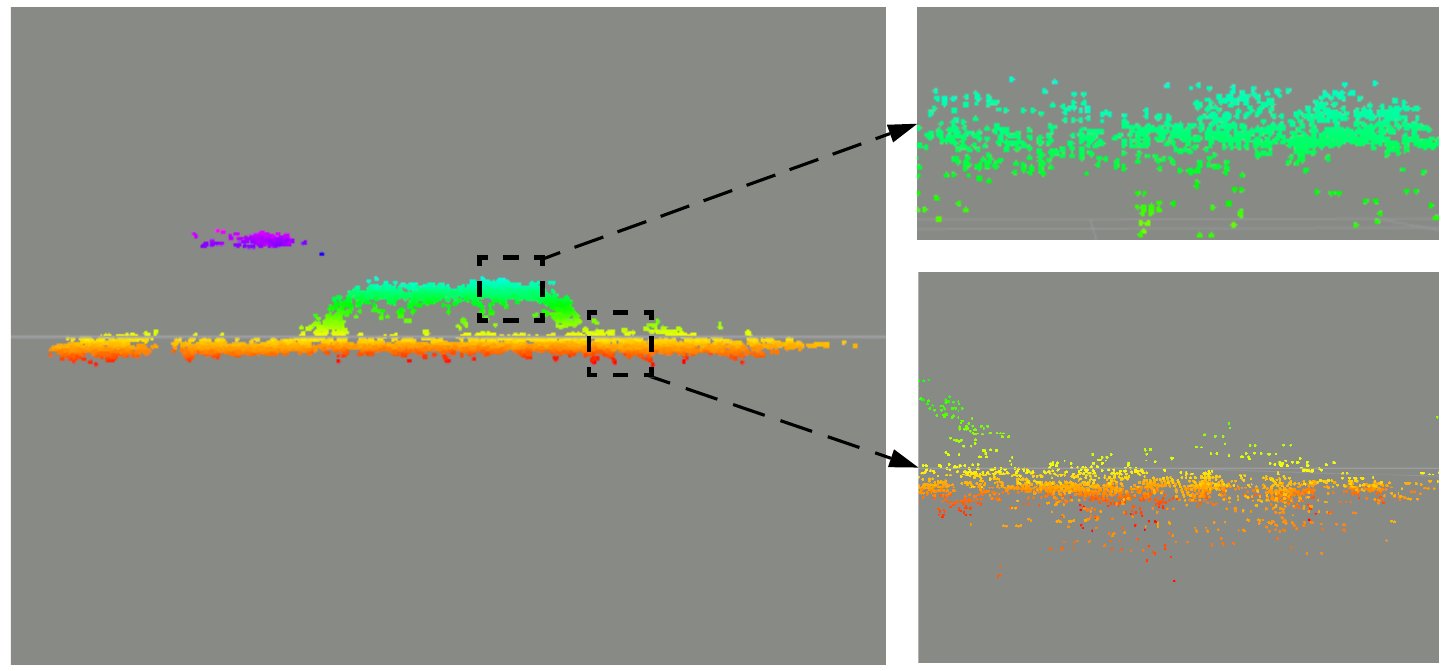}
  \label{fig:experim:solvers}}\\
  \subfigure[b][\small{$\emph{Student's t}$ distribution based IRLS solver.}]{
  \includegraphics[width=0.8\columnwidth]{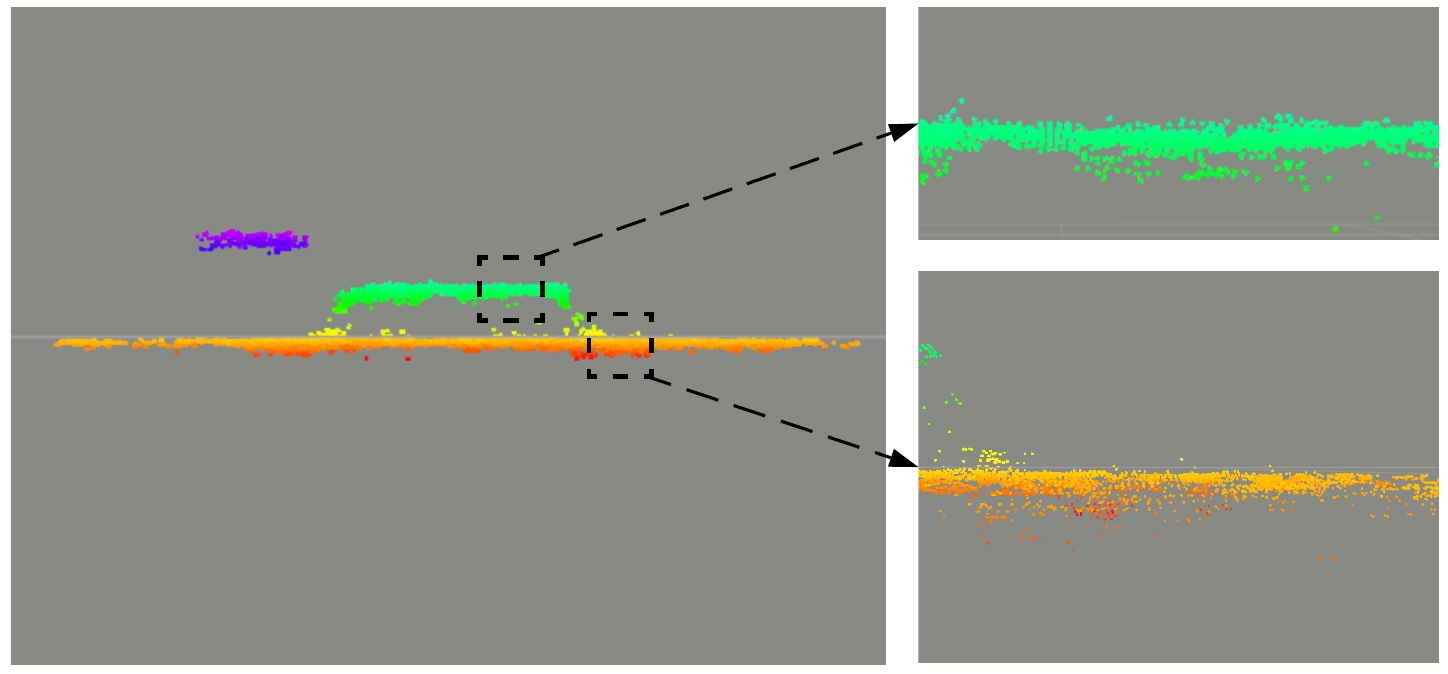}
  \label{fig:mapping:student-t-reconstruction}}
  \caption{\emph{Mapping}. 
  Qualitative comparison between standard least-squares (LS) solver and \emph{Student's}~$t$ distribution-based iteratively reweighted LS (IRLS) solver. 
  Regions highlighted with dashes are zoomed in for better visualization of details.}
  \label{fig:mapping:LS-vs-IRLS}
  \vspace{0.5ex}
  
  \captionof{table}{Comparison between standard least-squares (LS) solver and \emph{Student's}~$t$ distribution-based IRLS solver.}
  \vspace{0.5ex}
  \label{tab:mapping:LS-vs-IRLS}
  \renewcommand{\arraystretch}{1.1}
  \setlength{\tabcolsep}{0.5em}
  \begin{adjustbox}{max width=0.7\linewidth}
    \begin{tabular}{@{}lccc@{}} %
    \toprule
    & \#Fusions $\uparrow$ & Mean error $\downarrow$ & Std. $\downarrow$ \\
    \midrule
    $L_{2}$ norm & 3.33$\cdot 10^5$ & 2.76 cm & 2.94 cm\\[0.5ex]
    \emph{Student's t} & \textbf{5.07}$\cdot 10^5$ & \textbf{2.15} cm & \textbf{1.29} cm\\
    \bottomrule
    \end{tabular}
  \end{adjustbox}
\end{figure}
With this experiment we briefly justify the probabilistic inverse depth model derived from empirical observations of the distribution of time-surface residuals (Fig.~\ref{fig:temporal-residual-distribution}); two very different but related quantities~\eqref{eq:uncertainty-propagation-rho}.
Using synthetic data from~\cite{Mueggler17ijrr}, Fig.~\ref{fig:mapping:LS-vs-IRLS} and Table~\ref{tab:mapping:LS-vs-IRLS} show that the proposed probabilistic approach leads to more accurate 3D reconstructions than the standard least-squares (LS) objective criterion.
The synthetic scene in Fig.~\ref{fig:mapping:LS-vs-IRLS} consists of three planes parallel to the image planes of the cameras at different depths.
The reconstruction results in Fig.~\ref{fig:mapping:student-t-reconstruction} shows more accurate planar structures than those in Fig. \ref{fig:experim:solvers}.
As quantified in Table~\ref{tab:mapping:LS-vs-IRLS}, the depth error's standard deviation of the \studt{} distribution-based objective is 2-3 times smaller than that of the standard LS objective, which explains the more compact planar reconstructions in Fig.~\ref{fig:mapping:student-t-reconstruction} over Fig. \ref{fig:experim:solvers}.

\subsection{Comparison of Stereo 3D Reconstruction Methods}
\label{sec:experiments:mapping:other-stereo}
To prove the effectiveness of the proposed mapping method, we compare against three stereo methods and ground truth depth when available.
The baseline methods are abbreviated by \GTS{} \cite{Ieng18fnins}, SGM~\cite{Hirschmuller08pami} and CopNet~\cite{Piatkowska17cvprw}.

\paragraph{Description of Baseline Methods}
The method in~\cite{Ieng18fnins} proposes to match events by using a per-event time-based consistency criterion that also works on grayscale events from the ATIS~\cite{Posch11ssc} camera; after that, classical triangulation provides the 3D point location.
Since the code for this method is not available, we implement an abridged version of it, without the term for grayscale events because they are not available with the DAVIS.
The semiglobal matching (SGM) algorithm~\cite{Hirschmuller08pami}, available in OpenCV, is originally designed to solve the stereo matching problem densely on frame-based inputs.
We adapt it to our problem by running it on the stereo time surfaces and masking the produced depth map so that depth estimates are only given at pixels where recent events happened.
The method in~\cite{Piatkowska17cvprw} (CopNet) applies a cooperative stereo strategy~\cite{Marr76Science} in an asynchronous fashion.
We use the implementation in~\cite{Zhu18eccv}, where identical parameters are applied.

For a fair comparison against our method, which incrementally fuses successive depth estimates, we also propagate the depth estimates produced by \GTS{} and SGM. 
Since the baselines do not provide uncertainty estimates, we simply warp depth estimates from the past to the present time (i.e., the time where fusion is triggered in our method).
All methods start and terminate at the same time, and use ground truth poses to propagate depth estimates in time so that the evaluation does not depend on the tracking module.
Due to software incompatibility, propagation was not applied to CopNet. 
Therefore, CopNet is called only at the evaluation time; however, the density of its resulting inverse depth map is satisfactory when fed with enough number of events (\SI{15000}{\text{events}}~\cite{Zhu18eccv}).

\global\long\def\figWidth{0.17\linewidth}
\begin{figure*}[!ht]
	\centering
    {\small
    \setlength{\tabcolsep}{2pt}
	\begin{tabular}{
	>{\centering\arraybackslash}m{0.5cm} 
	>{\centering\arraybackslash}m{\figWidth} 
	>{\centering\arraybackslash}m{\figWidth}
	>{\centering\arraybackslash}m{\figWidth} 
	>{\centering\arraybackslash}m{\figWidth}
	>{\centering\arraybackslash}m{\figWidth}}
		& Scene & Inverse depth by \GTS~\cite{Ieng18fnins} & Inverse depth by SGM~\cite{Hirschmuller08pami} & Inverse depth by CopNet~\cite{Piatkowska17cvprw} & Inverse depth (Ours)
		\\\addlinespace[1ex]

		\rotatebox{90}{\makecell{\rpgreader{}}}
		&\includegraphics[width=\linewidth]{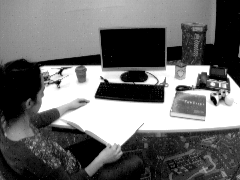}
		&\includegraphics[width=\linewidth]{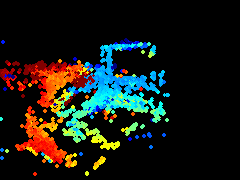}
		&\includegraphics[width=\linewidth]{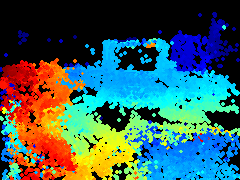}
		&\includegraphics[width=\linewidth]{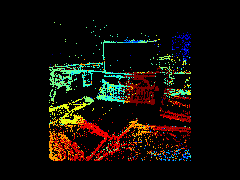}
		&\includegraphics[width=\linewidth]{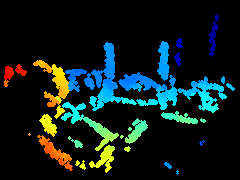}
		\\

		\rotatebox{90}{\makecell{\rpgbox{}}}
		&\includegraphics[width=\linewidth]{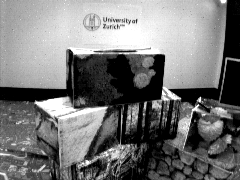}
		&\includegraphics[width=\linewidth]{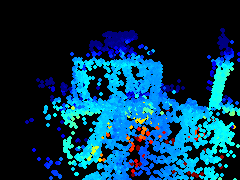}
		&\includegraphics[width=\linewidth]{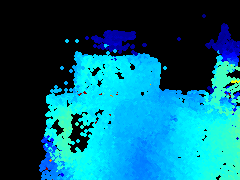}
		&\includegraphics[width=\linewidth]{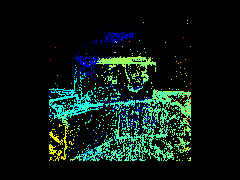}
		&\includegraphics[width=\linewidth]{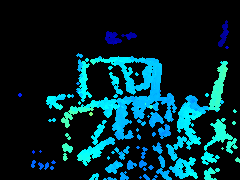}
		\\

		\rotatebox{90}{\makecell{\rpgmonitor{}}}
		&\includegraphics[width=\linewidth]{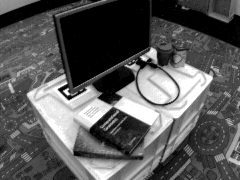}
		&\includegraphics[width=\linewidth]{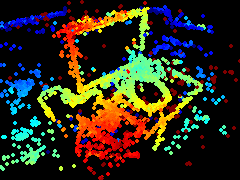}
		&\includegraphics[width=\linewidth]{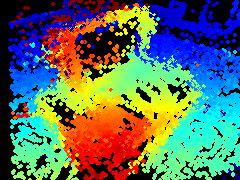}
		&\includegraphics[width=\linewidth]{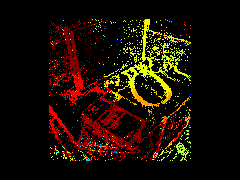}
		&\includegraphics[width=\linewidth]{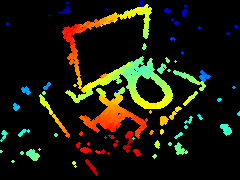}
		\\
		
		\rotatebox{90}{\makecell{\rpgbin{}}}
		&\includegraphics[width=\linewidth]{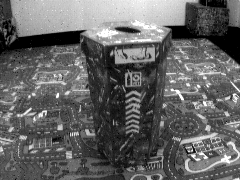}
		&\includegraphics[width=\linewidth]{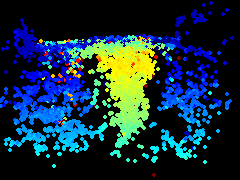}
		&\includegraphics[width=\linewidth]{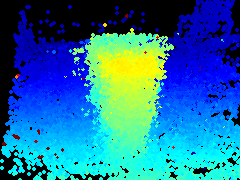}
		&\includegraphics[width=\linewidth]{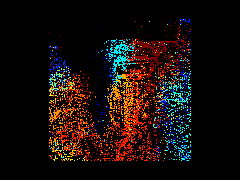}
		&\includegraphics[width=\linewidth]{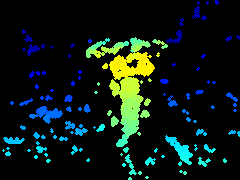}
		\\
  
		\rotatebox{90}{\makecell{\upennflyOne{}}}
		&\includegraphics[width=\linewidth]{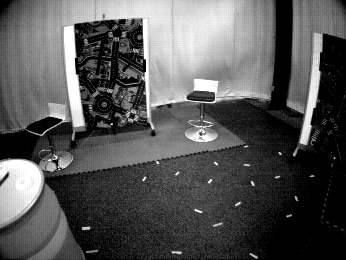}
		&\includegraphics[width=\linewidth]{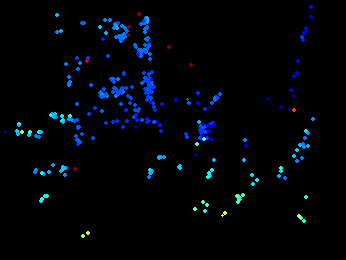}
		&\includegraphics[width=\linewidth]{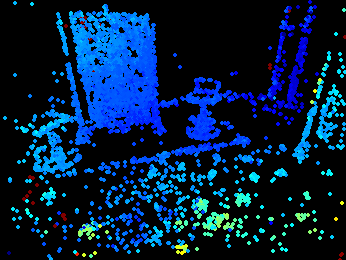}
		&\includegraphics[width=\linewidth]{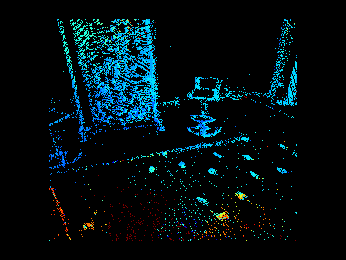}
		&\includegraphics[width=\linewidth]{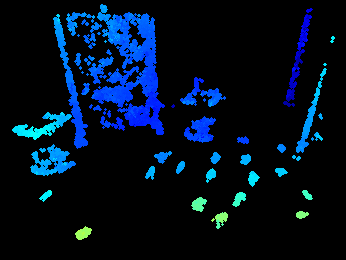}
		\\

		\rotatebox{90}{\makecell{\upennflyThree{}}}
		&\includegraphics[width=\linewidth]{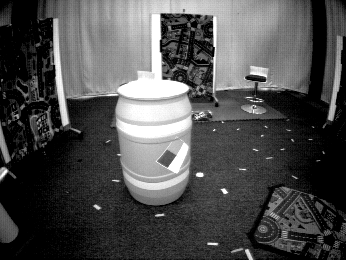}
		&\includegraphics[width=\linewidth]{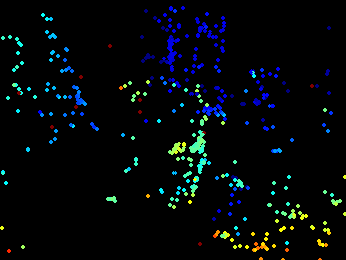}
		&\includegraphics[width=\linewidth]{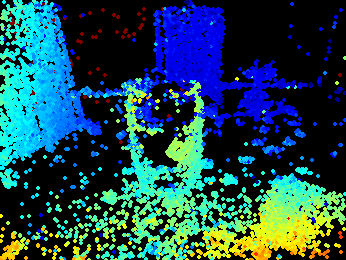}
				&\includegraphics[width=\linewidth]{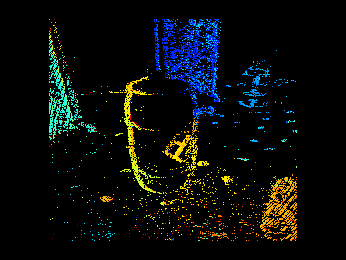}
		&\includegraphics[width=\linewidth]{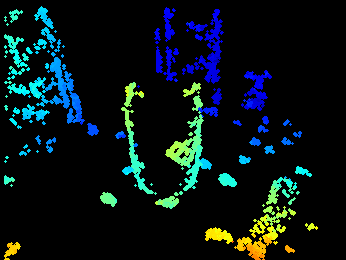}
		\\
	\end{tabular}
	}
	\caption{\label{fig:mapping:depthmaps-grid}\emph{Mapping}. 
	Qualitative comparison of mapping results (depth estimation) on several sequences using various stereo algorithms.
	The first column shows intensity frames from the DAVIS camera (not used, just for visualization).
	Columns 2 to 5 show inverse depth estimation results of \GTS~\cite{Ieng18fnins}, SGM~\cite{Hirschmuller08pami}, CopNet~\cite{Piatkowska17cvprw} and our method, respectively.
	Depth maps are color coded, from red (close) to blue (far) over a black background, 
	in the range \SIrange{0.55}{6.25}{\meter} for the top four rows (sequences from~\cite{Zhou18eccv})
	and the range \SIrange{1}{6.25}{\meter} for the bottom two rows (sequences from~\cite{Zhu18ral}).
	}
\end{figure*}
\paragraph{Results}
Fig.~\ref{fig:mapping:depthmaps-grid} compares the inverse depth maps produced by the above stereo methods.
The first column shows the raw grayscale frames from the DAVIS~\cite{Brandli13fns}, which only illustrate the appearance of the scenes because the methods do not use intensity information.
The second to the last columns show inverse depth maps produced by \GTS{}, SGM, CopNet and our method, respectively.
As expected because event cameras respond to the apparent motion of edges, the methods produce semi-dense depth maps that represent the 3D scene edges. 
This is more apparent in \GTS{}, CopNet and our method than in SGM because the regularizer in SGM helps to hallucinate depth estimates in regions where the spatio-temporal consistency is ambiguous, thus leading to the most dense depth maps.
Though CopNet produces satisfactory density results, it performs worse than our method in terms of depth accuracy.
This may be due to the fact that CopNet's output disparity is quantized to pixel accuracy.
In addition, the relatively large neighborhood size used (suggested by its creators~\cite{Piatkowska17cvprw}) introduces over-smoothing effects.
Finally, it can be observed that our method gives the best results in terms of compactness and signal-to-noise ratio.
This is due to the fact that we model both (inverse) depth and its uncertainty, which enables a principled multi-view depth fusion and pruning of unreliable estimates.
Since our method incrementally fuses successive depth estimates, the density of the resulting depth maps remains stable even though the streaming rate of events may vary, as is noticeable in the accompanying video.

An interesting phenomenon regarding the \GTS{} method is found: the density of the \GTS{}'s results on \emph{upenn} sequences are considerably lower than in \emph{rpg} sequences.
\emph{upenn} sequences differ from \emph{rpg} sequences in two aspects: 
($i$) they have larger depth range, 
and ($ii$) the motion is different (\emph{upenn} cameras are mounted on a drone which moves in a dominantly translating manner, while \emph{rpg} sequences are acquired with hand-held cameras performing general motions in 3D space).
The combination of both factors yields a smaller apparent motion of edges on the image plane in \emph{upenn} sequences; this may produce large times between corresponding events (originated by the same 3D edge).
To improve the density of the \GTS{}'s result, one may relax the maximum time distance used for event matching, which, however, would lead to less accurate and nosier depth estimation results.

We observe that the results of \rpgreader{} and \rpgbin{} are less sharp compared to those of \rpgbox{} and \rpgmonitor{}.
This is due to the different quality of the ground truth poses provided; we found that poses provided in \rpgreader{} and \rpgbin{} are less globally consistent than in other sequences.

Finally, table~\ref{tab:mapping:mvsec} quantifies the depth errors for the last two sequences of Fig.~\ref{fig:mapping:depthmaps-grid}, which are the ones where ground truth depth is available (acquired using a LiDAR \cite{Zhu18ral}).
Our method outperforms the baseline methods in all criteria: mean, median and relative error (with respect to the depth range).
\begin{table}[t]
    \caption{Quantitative evaluation of mapping on sequences with ground truth depth.} 
    \centering %
    \begin{adjustbox}{max width=\linewidth}
    \begin{tabular}{@{}llll@{}} %
    \toprule
     & Sequence \cite{Zhu18ral} & \upennflyOne{} & \upennflyThree{} \\[0.5ex]
    & Depth range [m] & 5.48 m & 6.03 m \\
    \midrule
    \GTS~\cite{Ieng18fnins} & Mean error & 0.31 m & 0.44 m \\
    & Median error & 0.18 m & 0.21 m \\
    & Relative error & 5.64 \% & 7.26 \% \\[0.5ex]
    SGM~\cite{Hirschmuller08pami} & Mean error & 0.31 m & 0.20 m \\
    & Median error & 0.15 m & 0.10 m \\
    & Relative error & 5.58 \% & 3.28 \% \\[0.5ex]
    CopNet~\cite{Piatkowska17cvprw} & Mean error & 0.59 m & 0.53 m \\
    & Median error & 0.49 m & 0.44 m \\
    & Relative error & 10.93 \% & 8.87 \% \\[0.5ex]
    Our Method~ & Mean error & $\mathbf{0.16}$ m & $\mathbf{0.19}$ m \\
    & Median error & $\mathbf{0.12}$ m & $\mathbf{0.09}$ m \\
    & Relative error & $\mathbf{3.05}$ \% & $\mathbf{3.13}$ \% \\
    \bottomrule
    \end{tabular}
    \end{adjustbox}
    \label{tab:mapping:mvsec}
\end{table}

\subsection{Full System Evaluation}
\label{sec:experiments:full-system}
To show the performance of the full VO system, we report ego-motion estimation results using two standard metrics: relative pose error (RPE) and absolute trajectory error (ATE)~\cite{Sturm12iros}.
Since no open-source event-based VO/SLAM projects is yet available, we implement a baseline that leverages commonly applied methods of depth and rigid-motion estimation in computer vision.
Additionally, we compare against a state-of-the-art frame-based SLAM pipeline (ORB-SLAM2~\cite{MurArtal17tro}) running on the grayscale frames acquired by the stereo DAVIS.

More specifically, the baseline solution, called ``SGM+ICP'', consists of combining the SGM method~\cite{Hirschmuller08pami} for dense depth estimation and the iterative closest point (ICP) method~\cite{Besl92pami} for estimating the relative pose between successive depth maps (i.e., point clouds). 
The whole trajectory is obtained by sequentially concatenating relative poses.

The evaluation is performed on six sequences with ground truth trajectories and the evaluation results can be found in Tables~\ref{tab:RPE-RMS} and~\ref{tab:ATE-RMS}.
The best results per sequence are highlighted in bold.
It is clear that our method outperforms the event-based baseline solution on all sequences.
To make the comparison against ORB-SLAM2 fair, global bundle adjustment (BA) was disabled; 
nevertheless, the results with global BA enabled are also reported in the tables, for reference.
Our system is slightly less accurate than ORB-SLAM2 on \emph{rpg} dataset, while shows a better performance on \emph{upenn\_indoor\_flying} dataset.
This is due to a flickering effect in the \emph{rpg} dataset induced by the motion capture system, which slightly deteriorates the performance of our method but does not appear on the grayscale frames used by ORB-SLAM2.

The trajectories produced by event-based methods are compared in Fig.~\ref{fig:tracking:dof-plots}. %
Our method significantly outperforms the event-based baseline SGM+ICP.
The evaluation of the full VO system using Fig.~\ref{fig:tracking:dof-plots} assesses whether the mapping and tracking remain consistent with each other. 
This requires the mapping module to be robust to the errors induced by the tracking module, and vice versa.
Our system does a remarkable job in this respect.

As a result of the above-mentioned flickering phenomena in \emph{rpg} datasets, the spatio-temporal consistency across stereo time-surface maps may not hold well all the time.
We find that our system performs robustly under this challenging scenario as long as it does not occur during initialization.
Readers can get a better understanding of the flickering phenomena by watching the accompanying video.

The VO results on the \emph{upenn} dataset show worse accuracy compared to those on the \emph{rpg} dataset.
This may be attributed to the following two reasons.
First, the motion pattern (dominant translation with slight rotation) determines that no structures parallel to the baseline of the stereo rig are reconstructed (as will be discussed in Fig.~\ref{fig:mapping:missing-edges:after-thresholding}).
These missing structures may lead to less accurate motion estimation in the corresponding degree of freedom.
Second, the accuracy of the system (tracking and mapping) is limited by the relatively small spatial resolution of the sensor.
Using event cameras with higher resolution (\eg, VGA~\cite{Son17isscc}) would improve the accuracy of the system.
Finally, note that when the drone stops and hovers few events are generated and, thus, time surfaces triggered at constant rate become unreliable.
This would cause our system to reinitialize.
It could be mitigated by using more complex strategies to signal the creation of time surfaces, such as a constant or adaptive number of events~\cite{Liu18bmvc}. 
However this is out of scope of the present work.
Thus, we only evaluate on the dynamic section of the dataset.

\begin{figure*}[!h]
  \centering
  \includegraphics[width=0.99\textwidth]{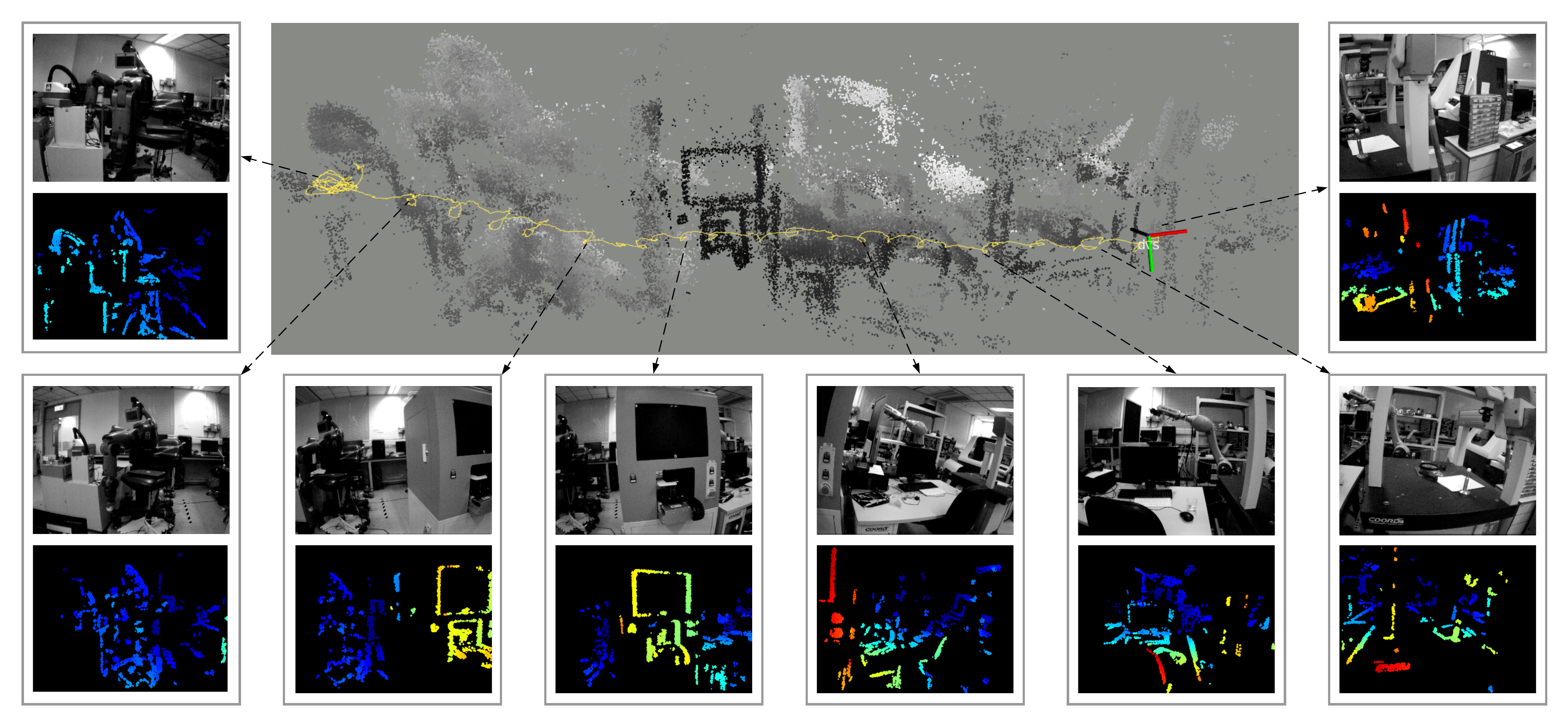}
  \vspace{-1ex}
\caption{Estimated camera trajectory and 3D reconstruction of $\emph{hkust\_lab}$ sequence. 
Computed inverse depth maps at selected viewpoints are visualized sequentially, from left to right. 
Intensity frames are shown for visualization purpose only.}
\label{fig:system-hkust}
\end{figure*}

\begin{table}[t]
\caption{Relative Pose Error (RMS) {[$\mathbf{R}$: \si{\degree/\second}, $\bt$: \si{\centi\meter/\second}]}
}
\centering
\begin{adjustbox}{max width=\linewidth}
\setlength{\tabcolsep}{3pt}
\begin{tabular}{@{}lllllllll@{}}
\toprule
& \multicolumn{2}{c}{ORB\_SLAM2 (Stereo)} && \multicolumn{2}{c}{SGM + ICP} && \multicolumn{2}{c}{Our Method} \\[0.4ex] 
\cmidrule{2-3} \cmidrule{5-6} \cmidrule{8-9}
Sequence & $\mathbf{R}$ & $\mathbf{t}$ 
&~& $\mathbf{R}$ & $\mathbf{t}$
&~& $\mathbf{R}$ & $\mathbf{t}$ \\[0.3ex] 
\midrule
\rpgbin{} & $\mathbf{0.6}$ (0.5) & $\mathbf{1.5}$ (1.2) && 7.6 & 13.3 && 1.2 & 3.1  \\ 
\rpgbox{} & $\mathbf{1.8}$ (1.7) & $\mathbf{5.1}$ (2.7) && 7.9 & 15.5 && 3.4 & 7.2 \\ 
\rpgdesk{} & $\mathbf{2.4}$ (1.7) & $\mathbf{3.3}$ (2.8) && 10.1 & 14.6 && 3.1 & 4.5 \\ 
\rpgmonitor{} & $\mathbf{1.0}$ (0.6) & $\mathbf{1.8}$ (1.0) && 8.1 & 10.7 && 1.7 & 3.2 \\
\upennflyOne{} & 5.4 (5.8) & 20.4 (16.2) && 4.8  & 31.6 && $\mathbf{1.0}$ & $\mathbf{6.5}$ \\ 
\upennflyThree{} & 5.6 (3.0) & 22.0 (20.1) && 7.3 & 26.3 && $\mathbf{1.2}$  & $\mathbf{7.1}$ \\
\bottomrule
\end{tabular}
\end{adjustbox}
\begin{tablenotes}
\small
\item The numbers in parentheses in ORB\_SLAM2 represent the RMS errors with bundle adjustment enabled.
\end{tablenotes}
\label{tab:RPE-RMS}
\end{table}
\begin{table}[t]
\caption{Absolute Trajectory Error (RMS) {[$\bt$: \si{\centi\meter}]}}
\centering
\begin{adjustbox}{max width=\linewidth}
\begin{tabular}{@{}llll@{}}
\toprule
 & ORB\_SLAM2 & SGM + ICP & Our Method \\[0.4ex]
\midrule
\rpgbin{}       & $\mathbf{0.9}$ (0.7) & 13.8 & 2.8 \\ 
\rpgbox{}       & $\mathbf{2.9}$ (1.6) & 19.8 & 5.8 \\ 
\rpgdesk{}      & 7.7 (1.8) & 8.5 & $\mathbf{3.2}$ \\ 
\rpgmonitor{}   & $\mathbf{2.5}$ (0.8) & 29.5 & 3.3 \\ 
\upennflyOne{} & 49.8 (41.7) & 95.8 & $\mathbf{13.9}$ \\ 
\upennflyThree{} & 50.2 (36.5) & 55.7 & $\mathbf{11.1}$ \\ 
\bottomrule
\end{tabular}
\end{adjustbox}
\label{tab:ATE-RMS}
\end{table}
\global\long\def\dofPlotsWidth{0.2\textwidth}
\begin{figure*}[ht!]
	\centering
    {\small
    \setlength{\tabcolsep}{6pt}
	\begin{tabular}{
	>{\centering\arraybackslash}m{\dofPlotsWidth} 
	>{\centering\arraybackslash}m{\dofPlotsWidth}
	>{\centering\arraybackslash}m{\dofPlotsWidth} 
	>{\centering\arraybackslash}m{\dofPlotsWidth}}
		Translation $X$ & Translation $Y$ & Translation $Z$ & Orientation error
		\\\addlinespace[1ex]
		
   \includegraphics[trim={0 0 0 1.18cm},clip,width=\linewidth]{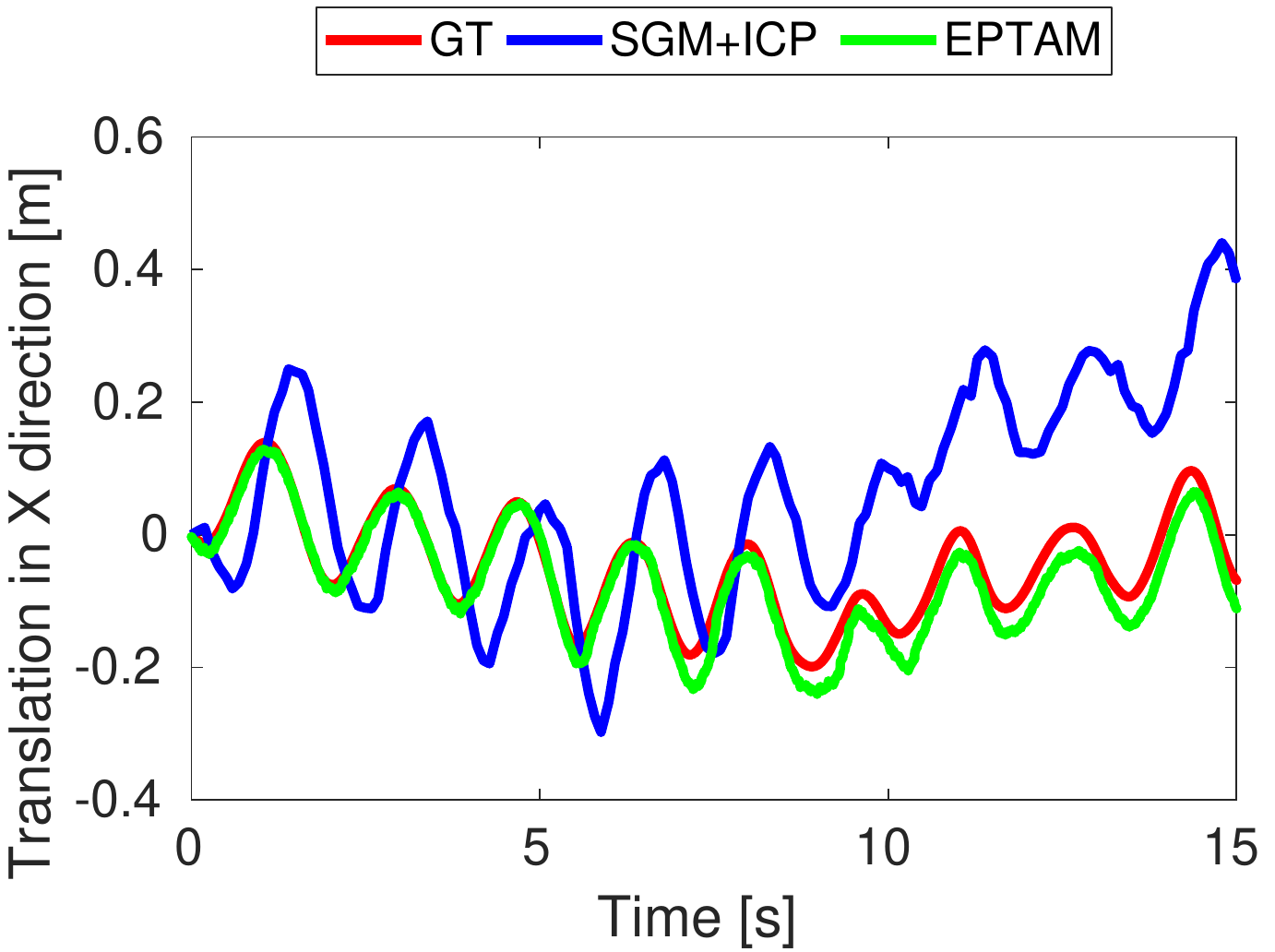}
  &\includegraphics[trim={0 0 0 1.18cm},clip,width=\linewidth]{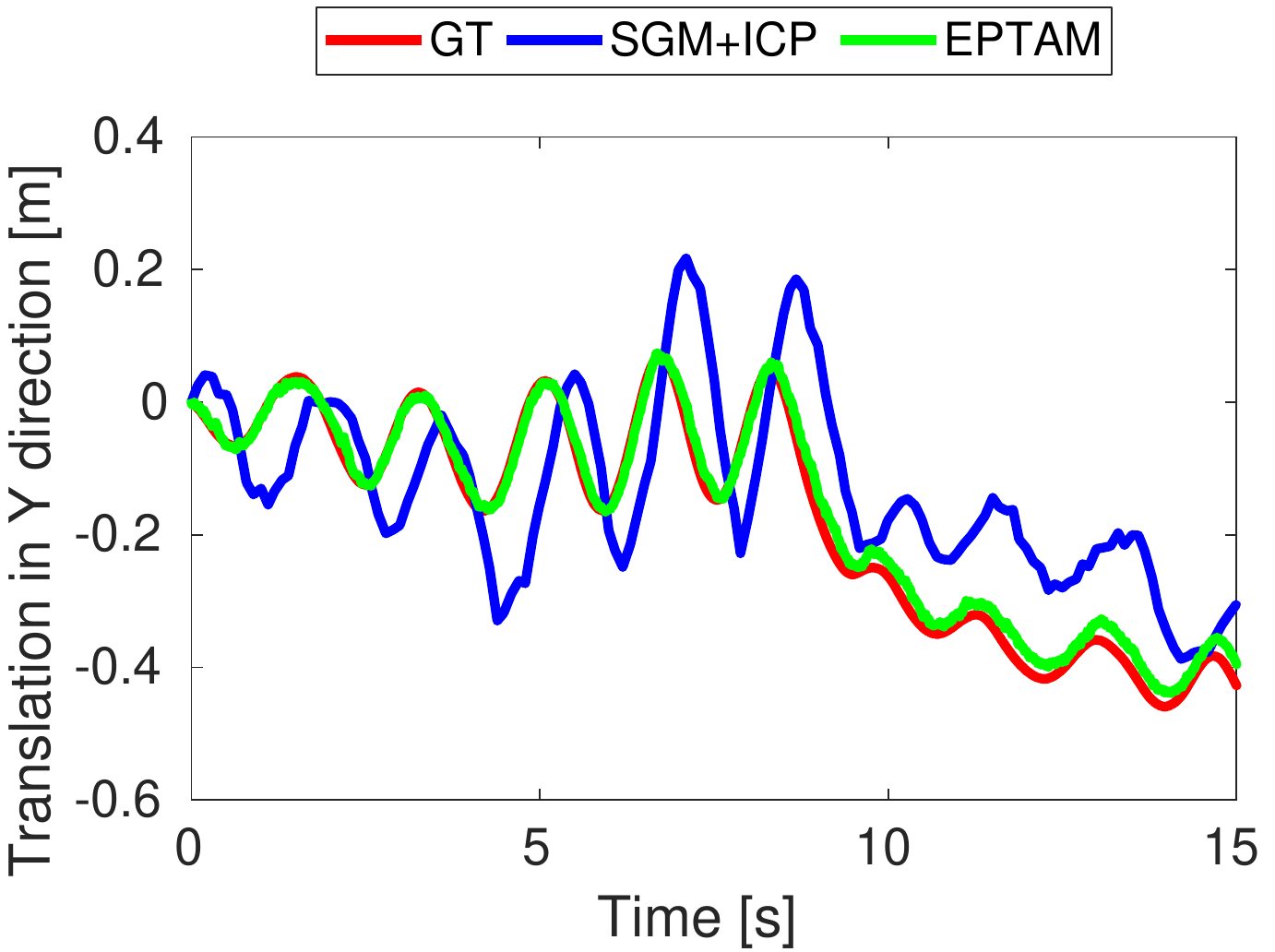}
  &\includegraphics[trim={0 0 0 1.18cm},clip,width=\linewidth]{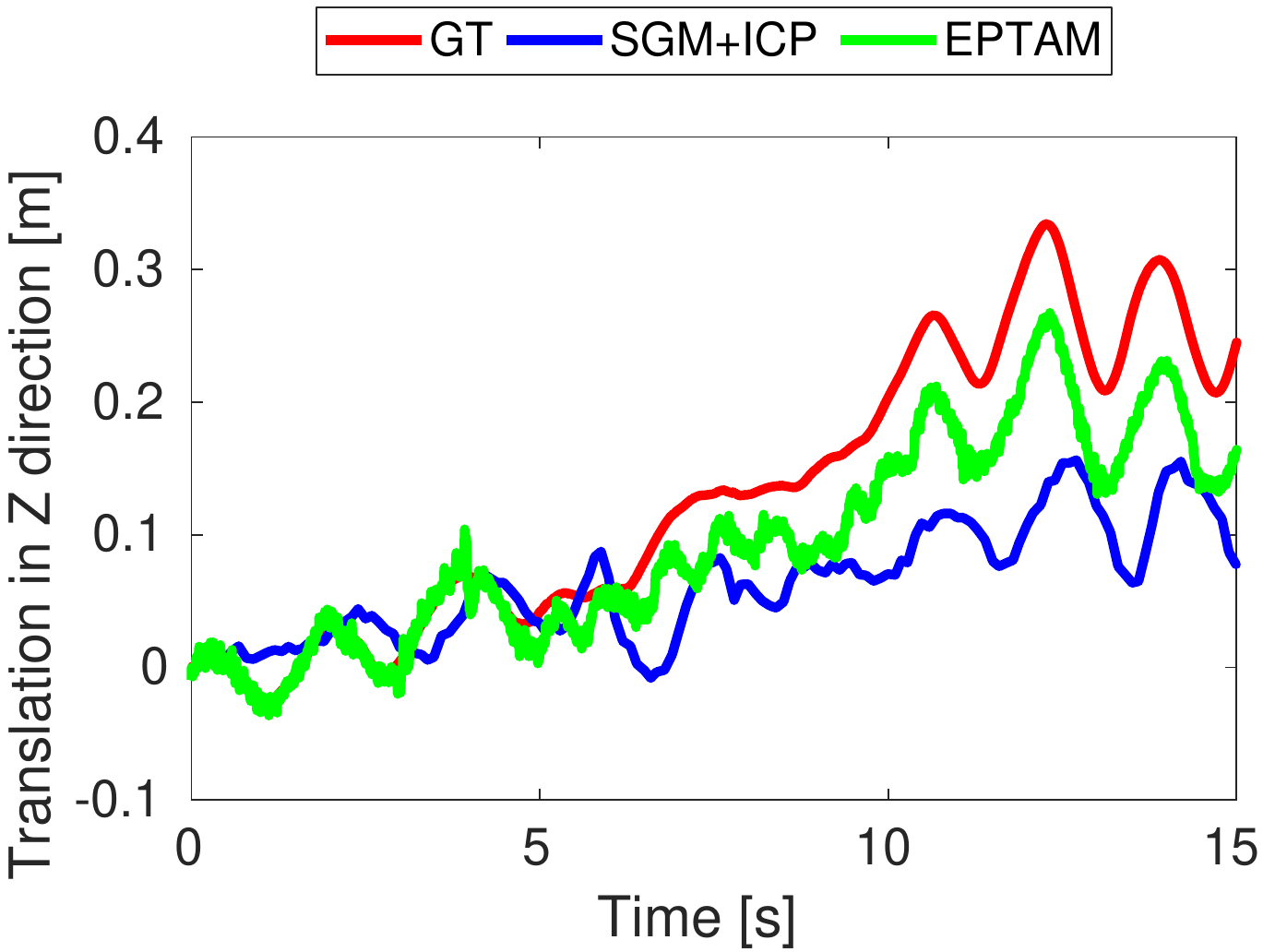}
  &\includegraphics[trim={0 0 0 1.18cm},clip,width=\linewidth]{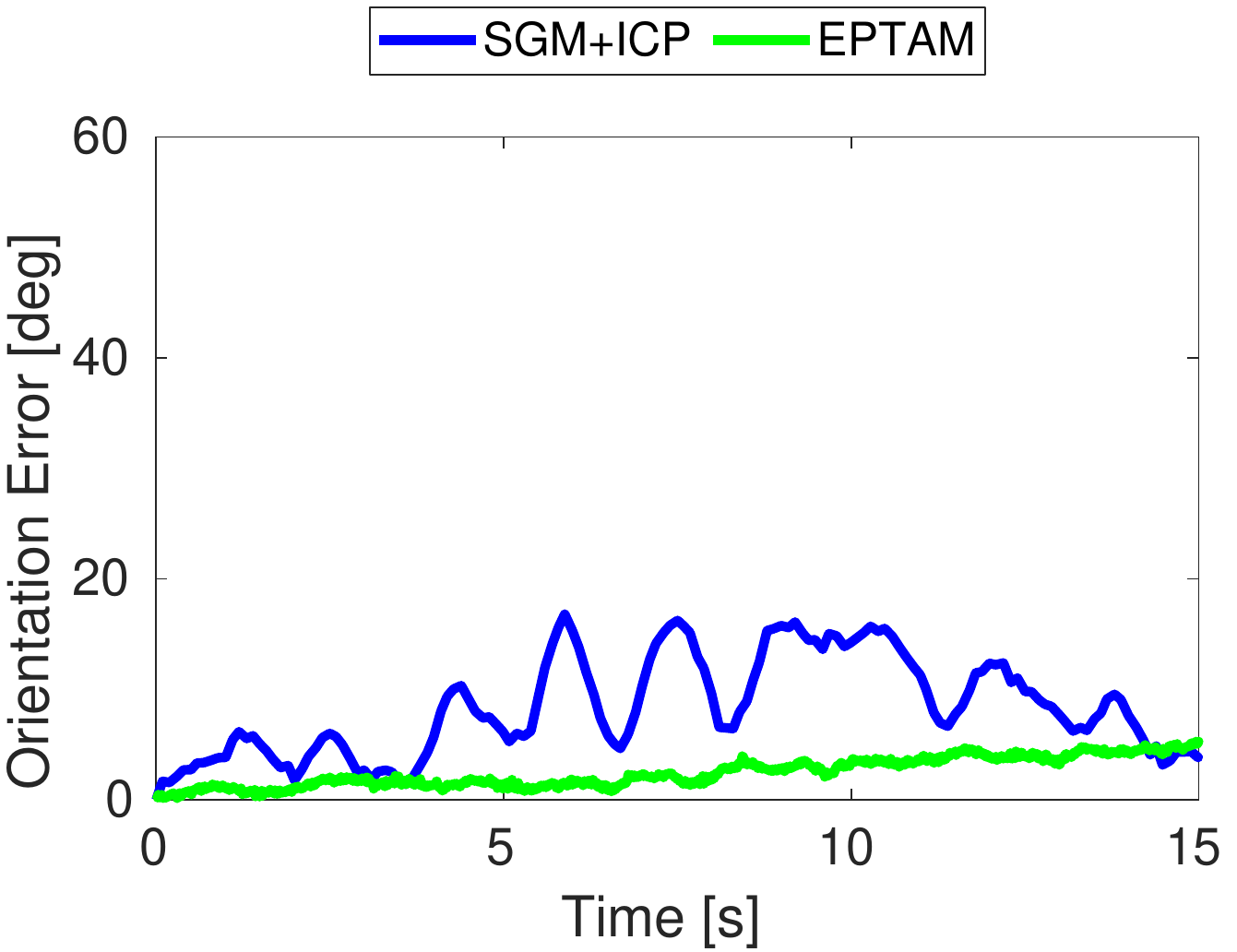}
  \\
   \includegraphics[trim={0 0 0 1.18cm},clip,width=\linewidth]{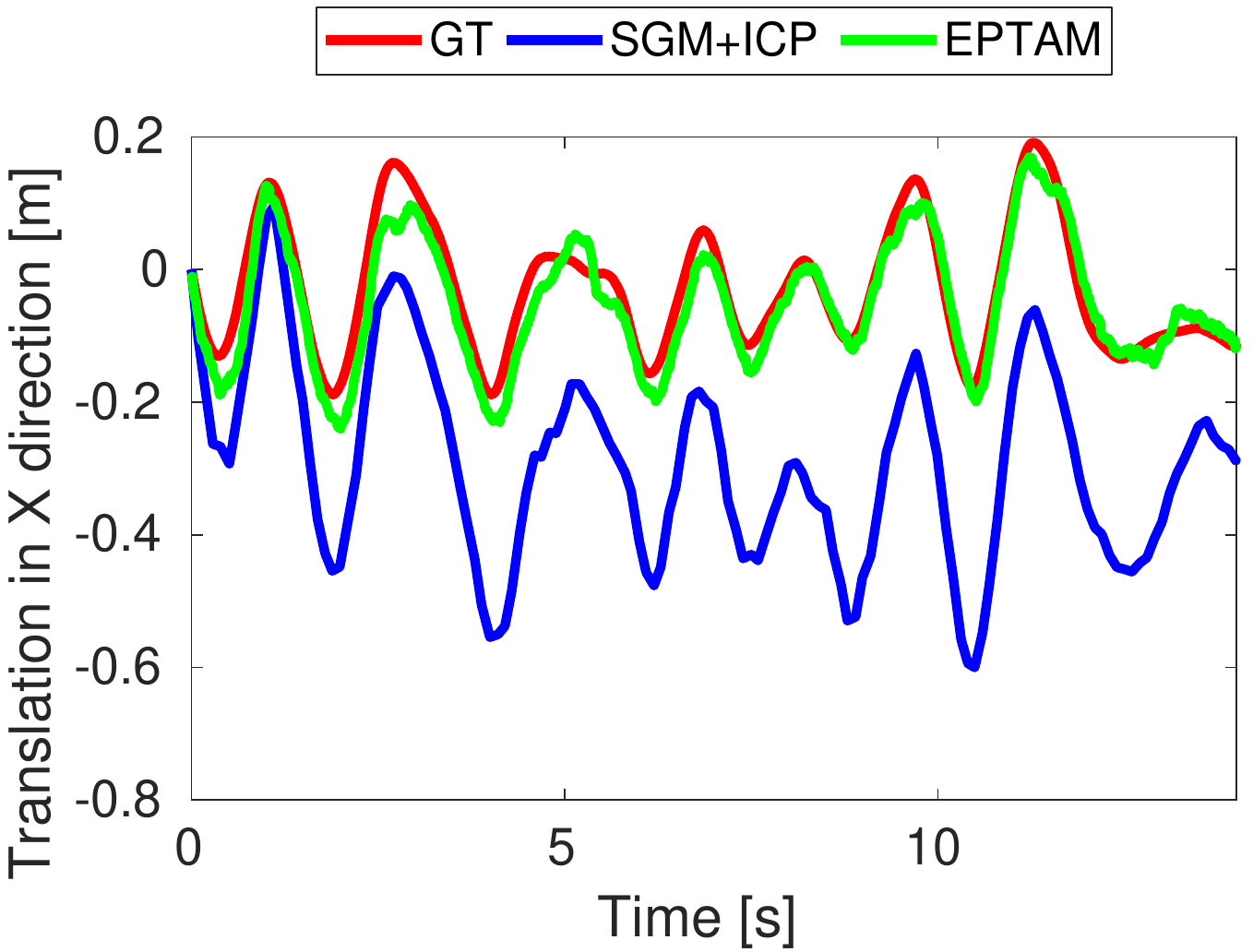}
  &\includegraphics[trim={0 0 0 1.18cm},clip,width=\linewidth]{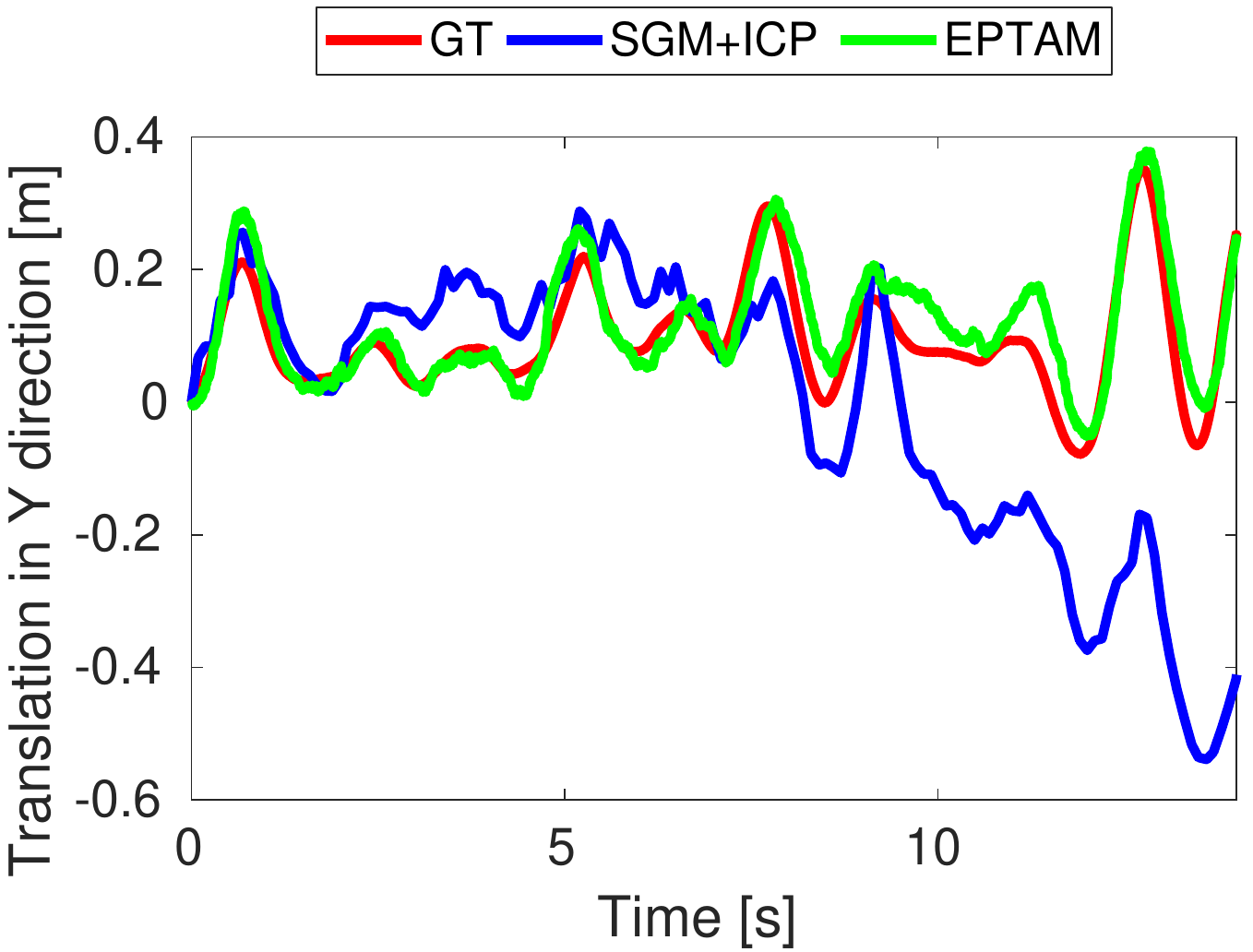}
  &\includegraphics[trim={0 0 0 1.18cm},clip,width=\linewidth]{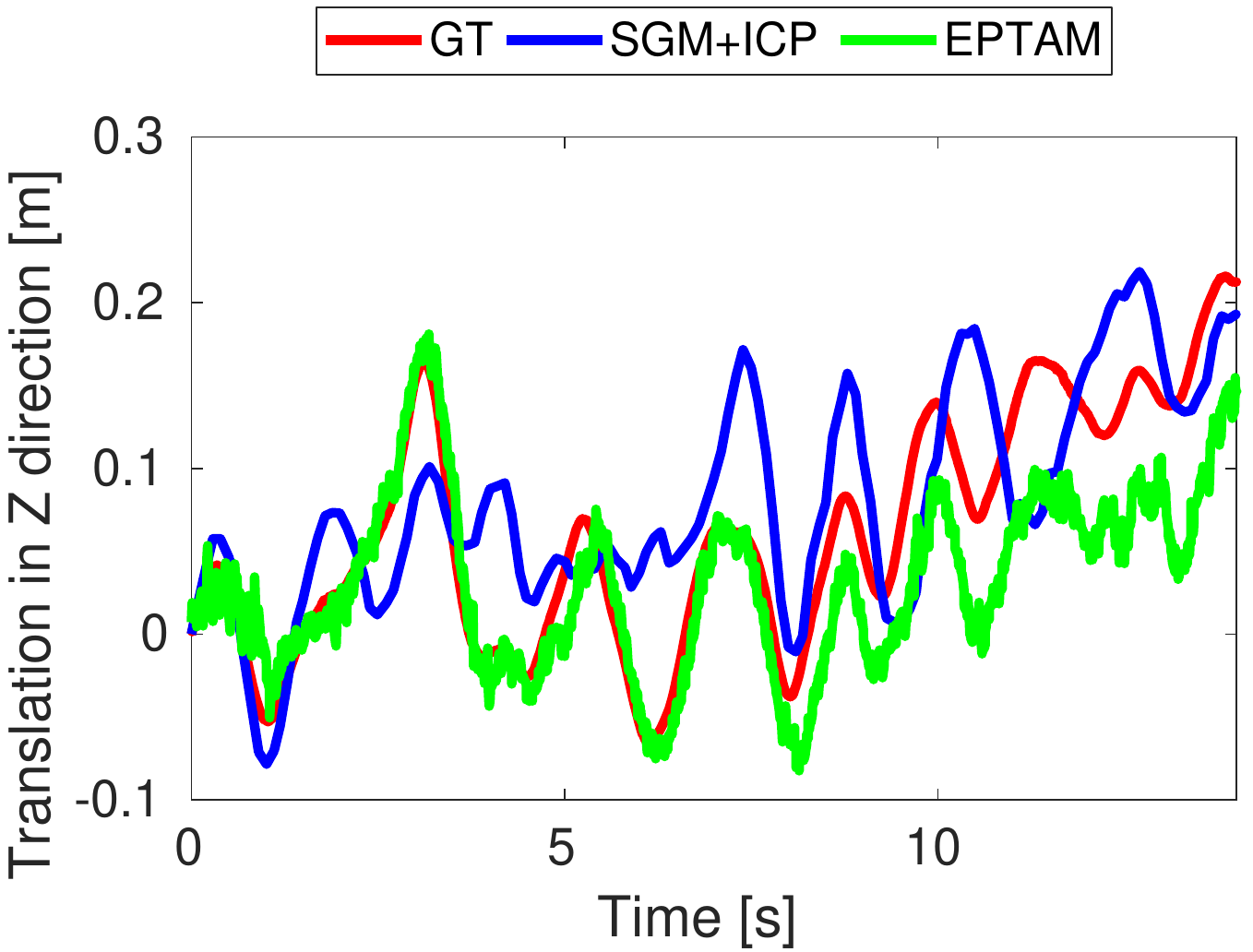}
  &\includegraphics[trim={0 0 0 1.18cm},clip,width=\linewidth]{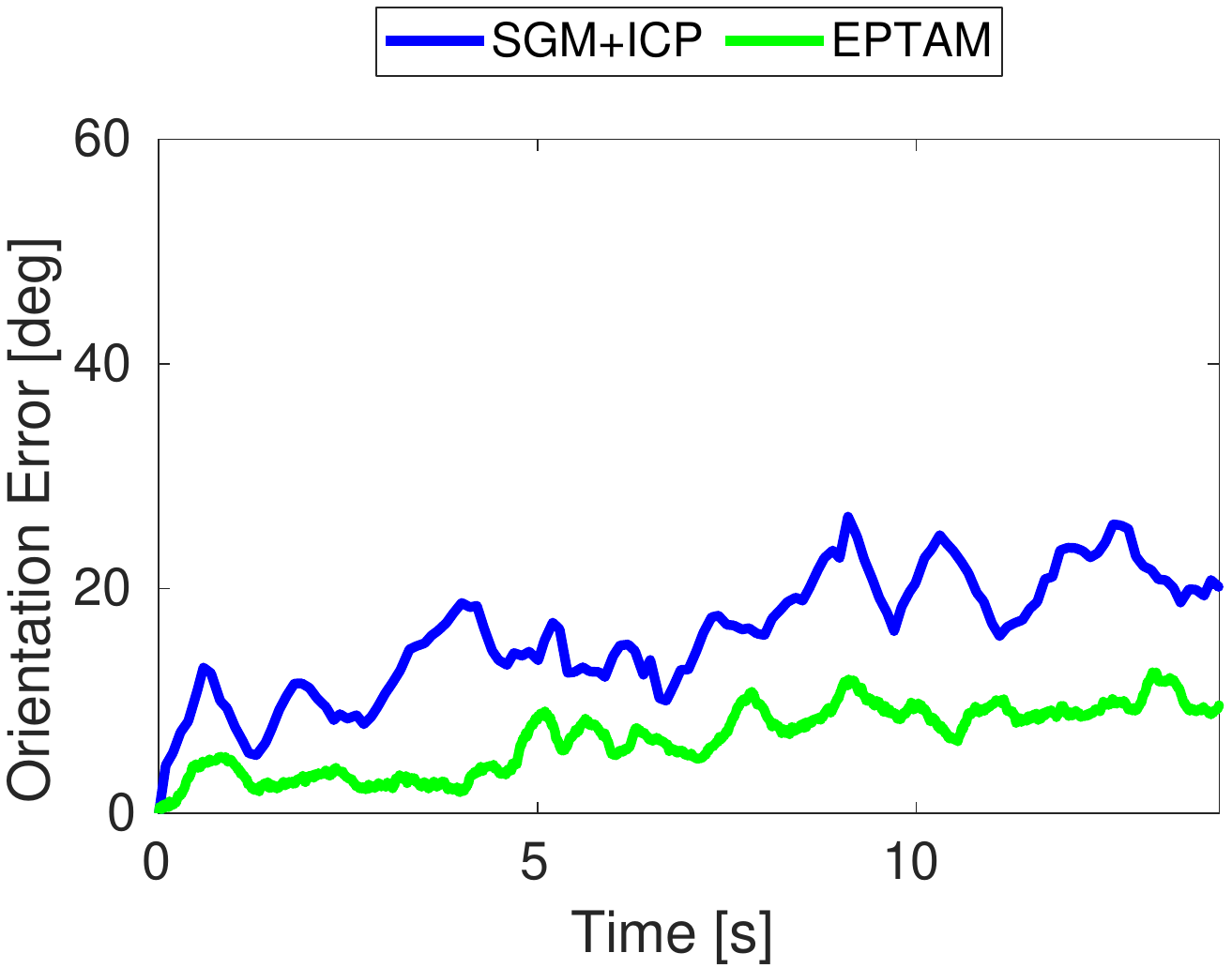}
  \\
   \includegraphics[trim={0 0 0 1.18cm},clip,width=\linewidth]{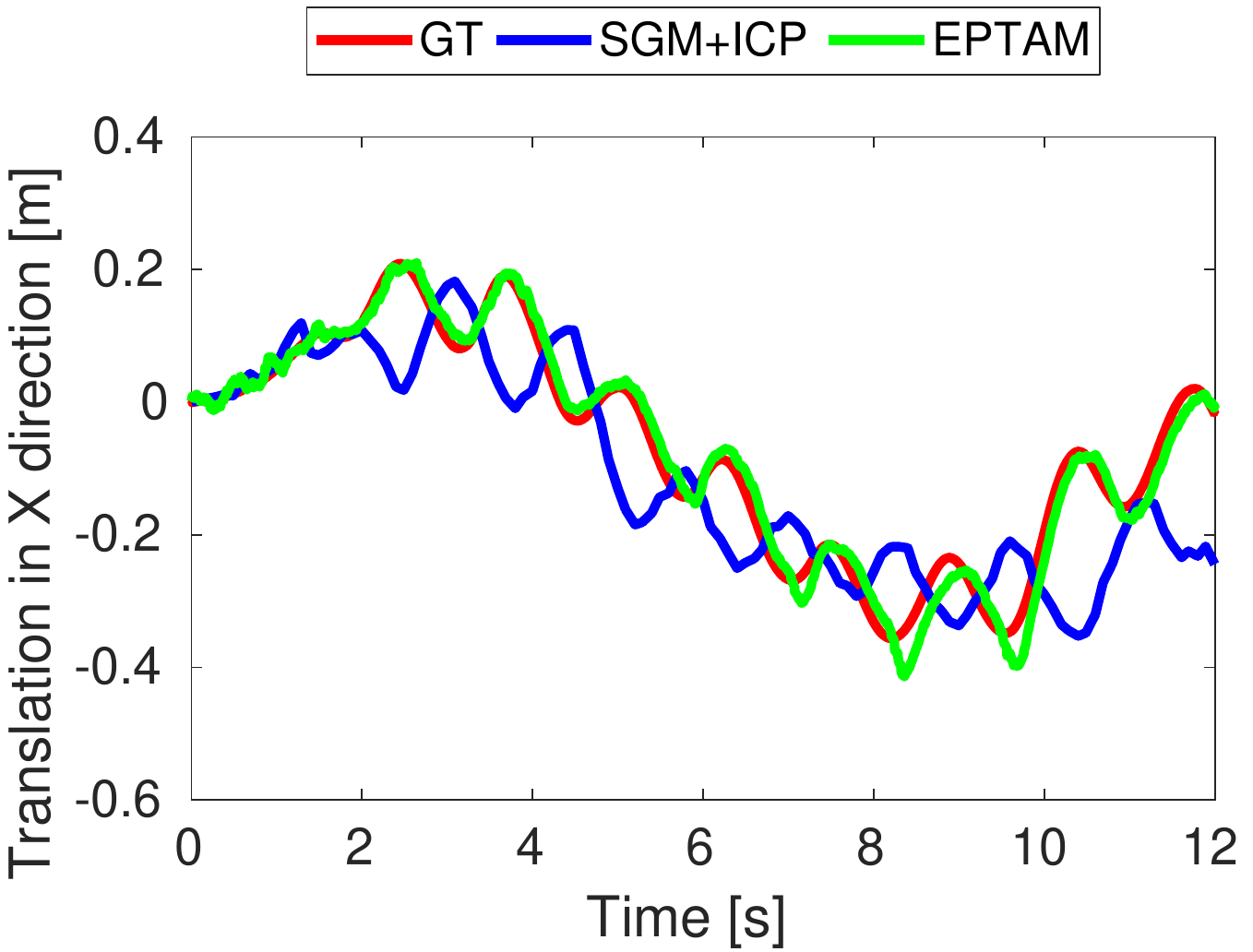}
  &\includegraphics[trim={0 0 0 1.18cm},clip,width=\linewidth]{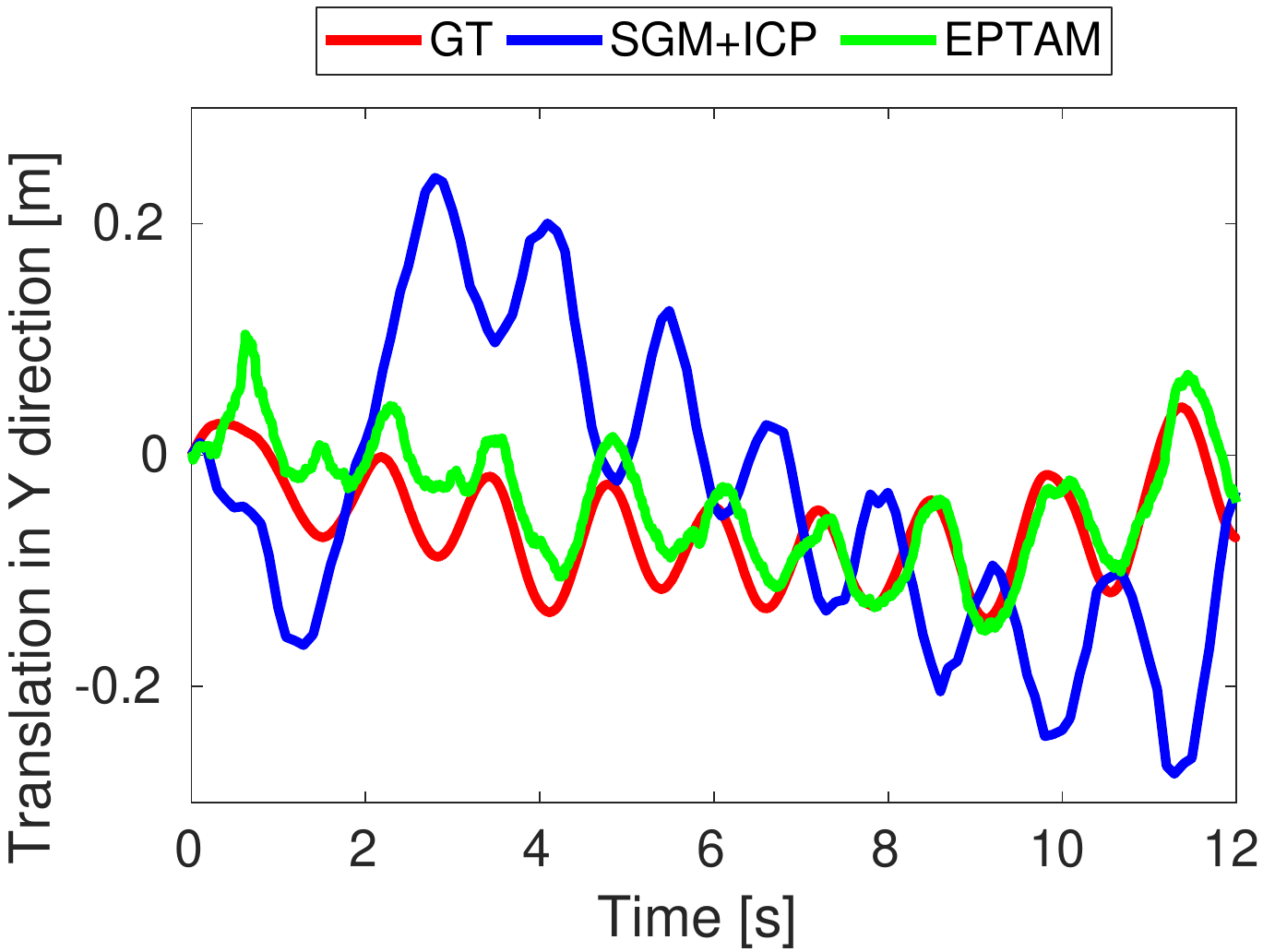}
  &\includegraphics[trim={0 0 0 1.18cm},clip,width=\linewidth]{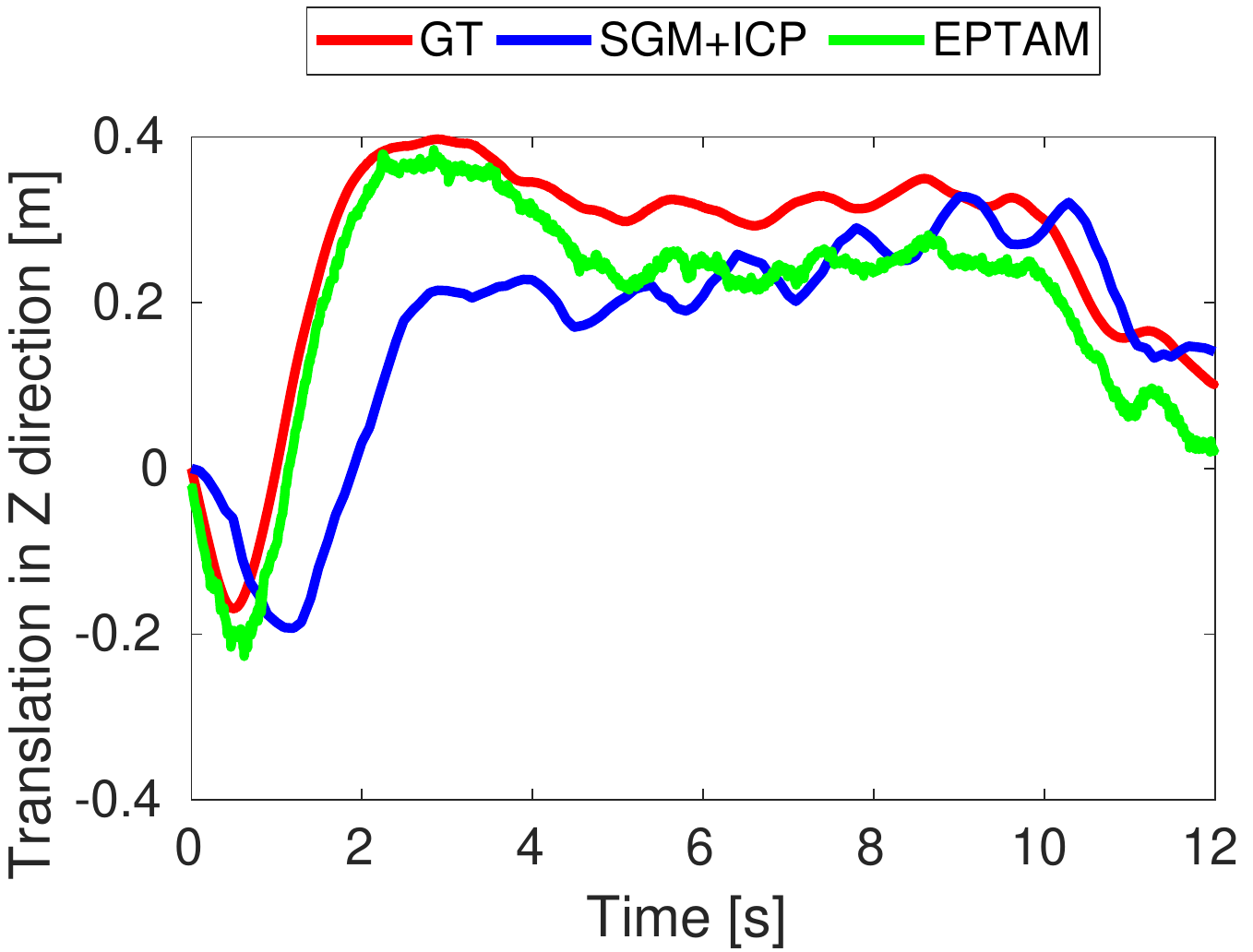}
  &\includegraphics[trim={0 0 0 1.18cm},clip,width=\linewidth]{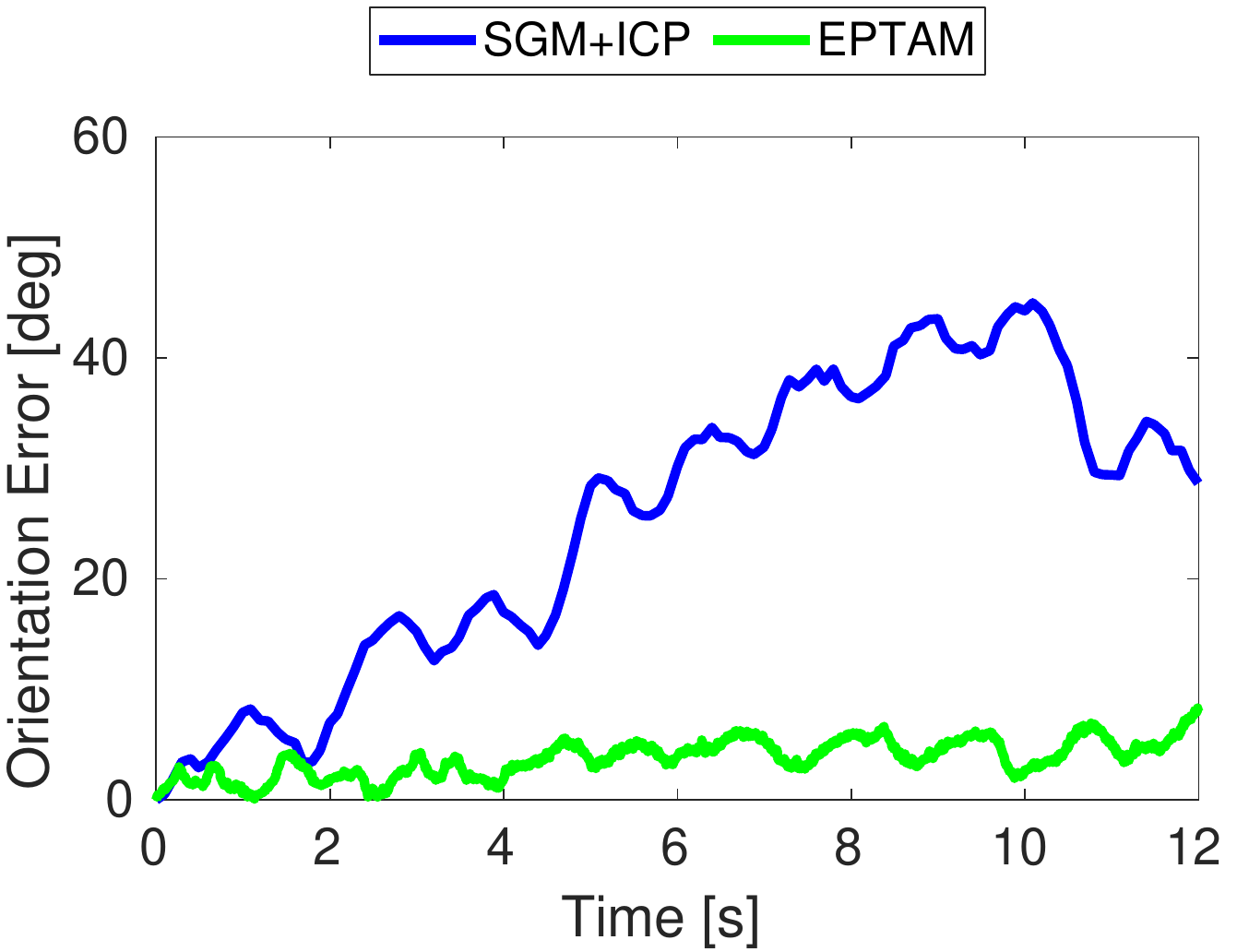}
  \\
   \includegraphics[trim={0 0 0 1.18cm},clip,width=\linewidth]{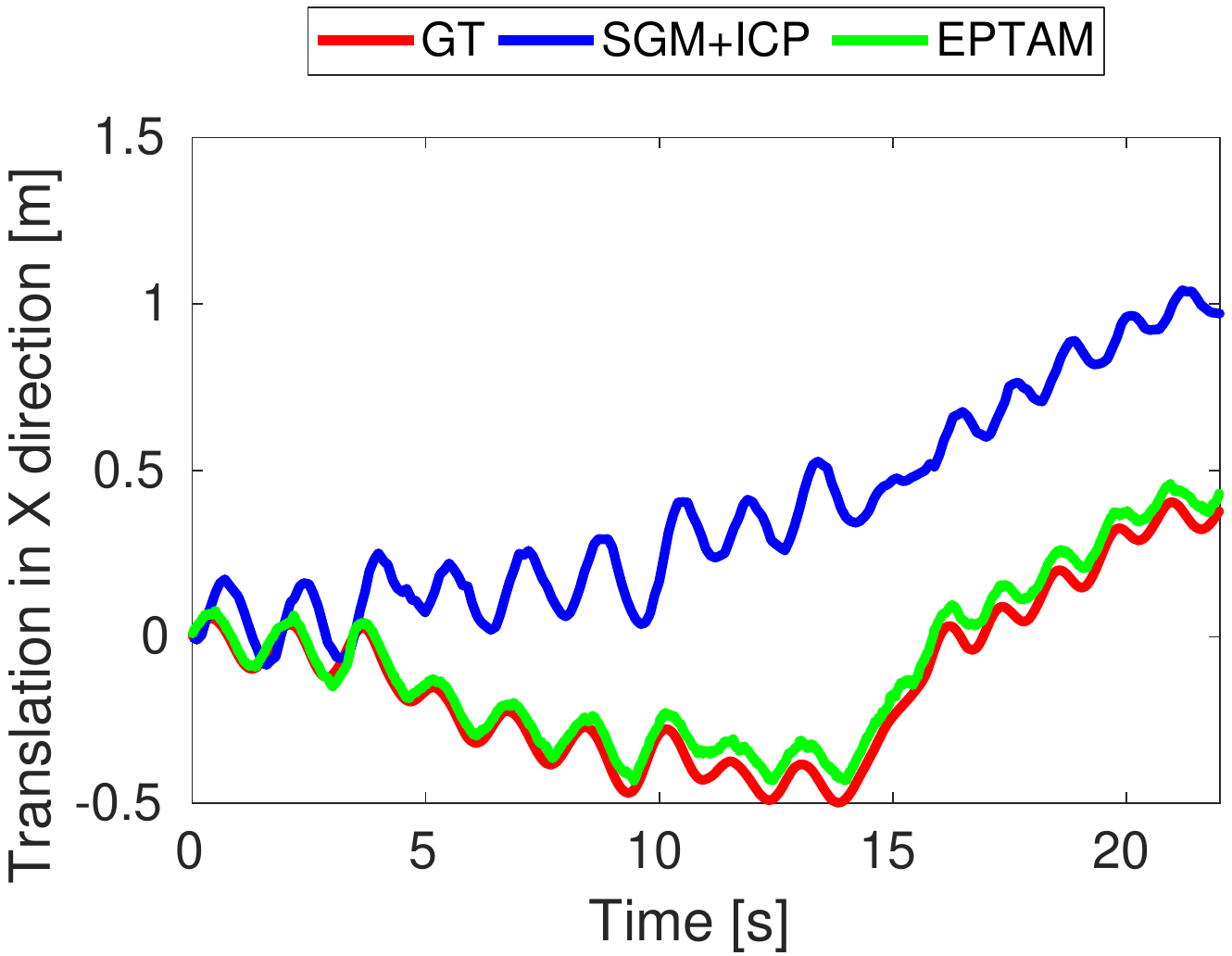}
  &\includegraphics[trim={0 0 0 1.18cm},clip,width=\linewidth]{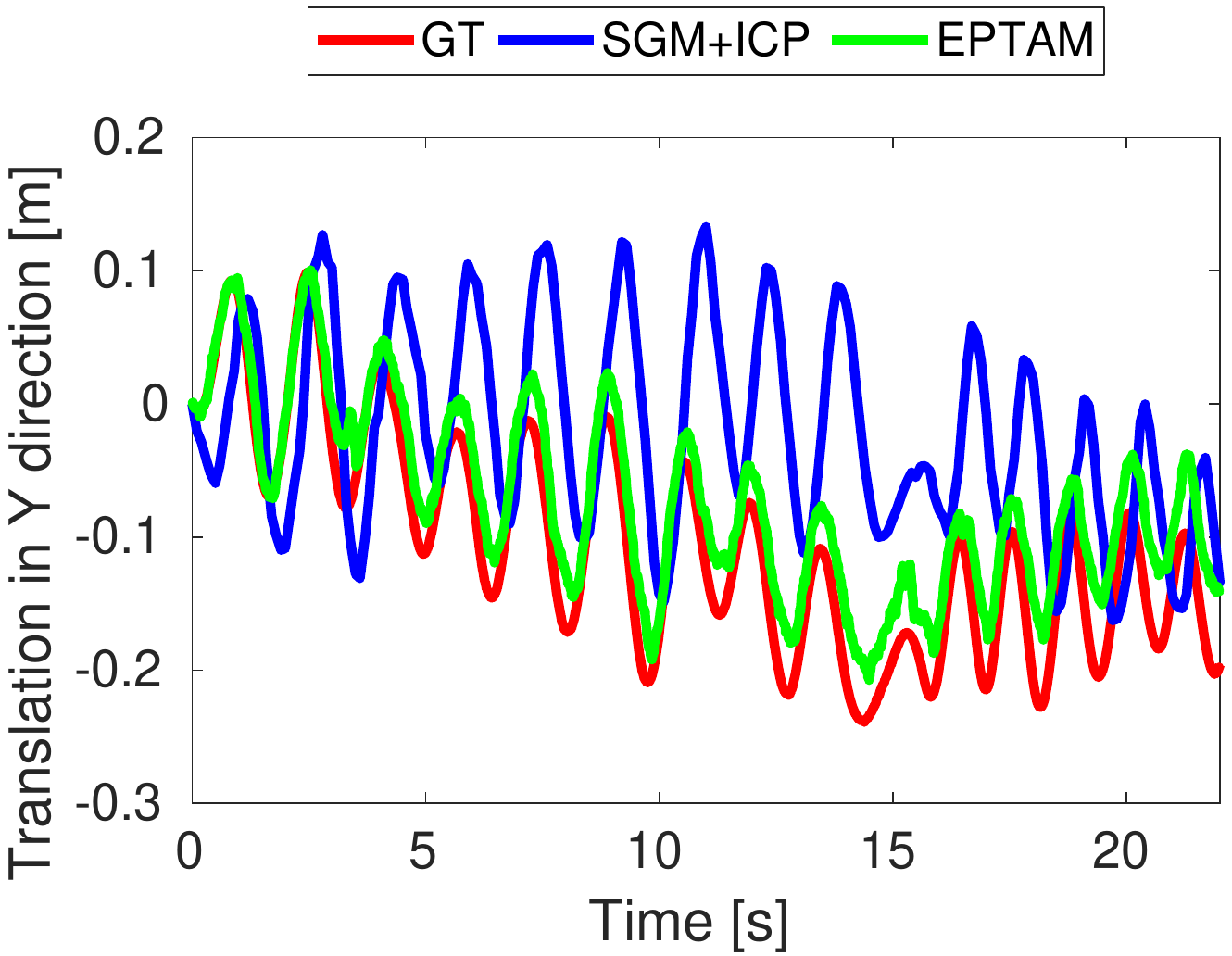}
  &\includegraphics[trim={0 0 0 1.18cm},clip,width=\linewidth]{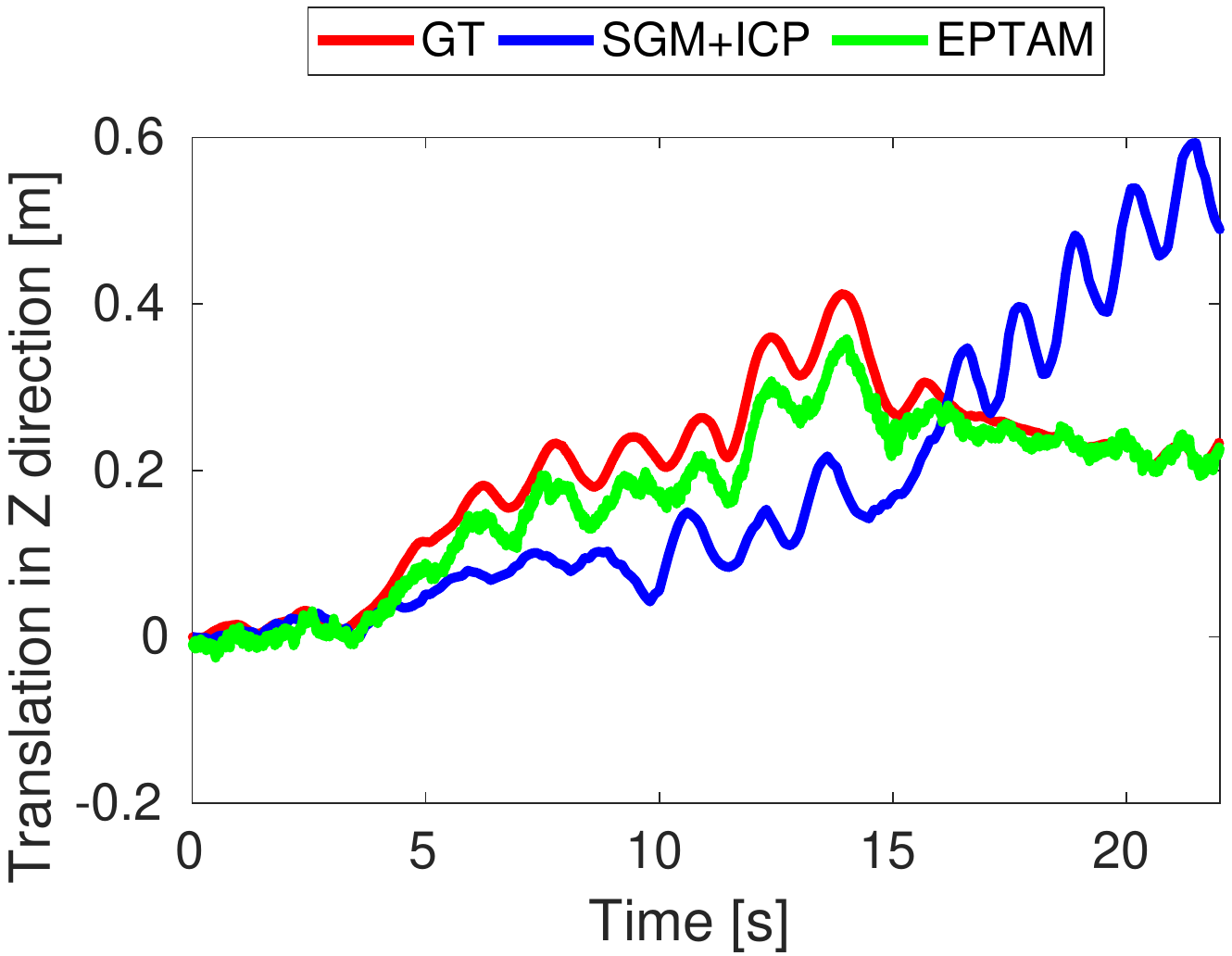}
  &\includegraphics[trim={0 0 0 1.18cm},clip,width=\linewidth]{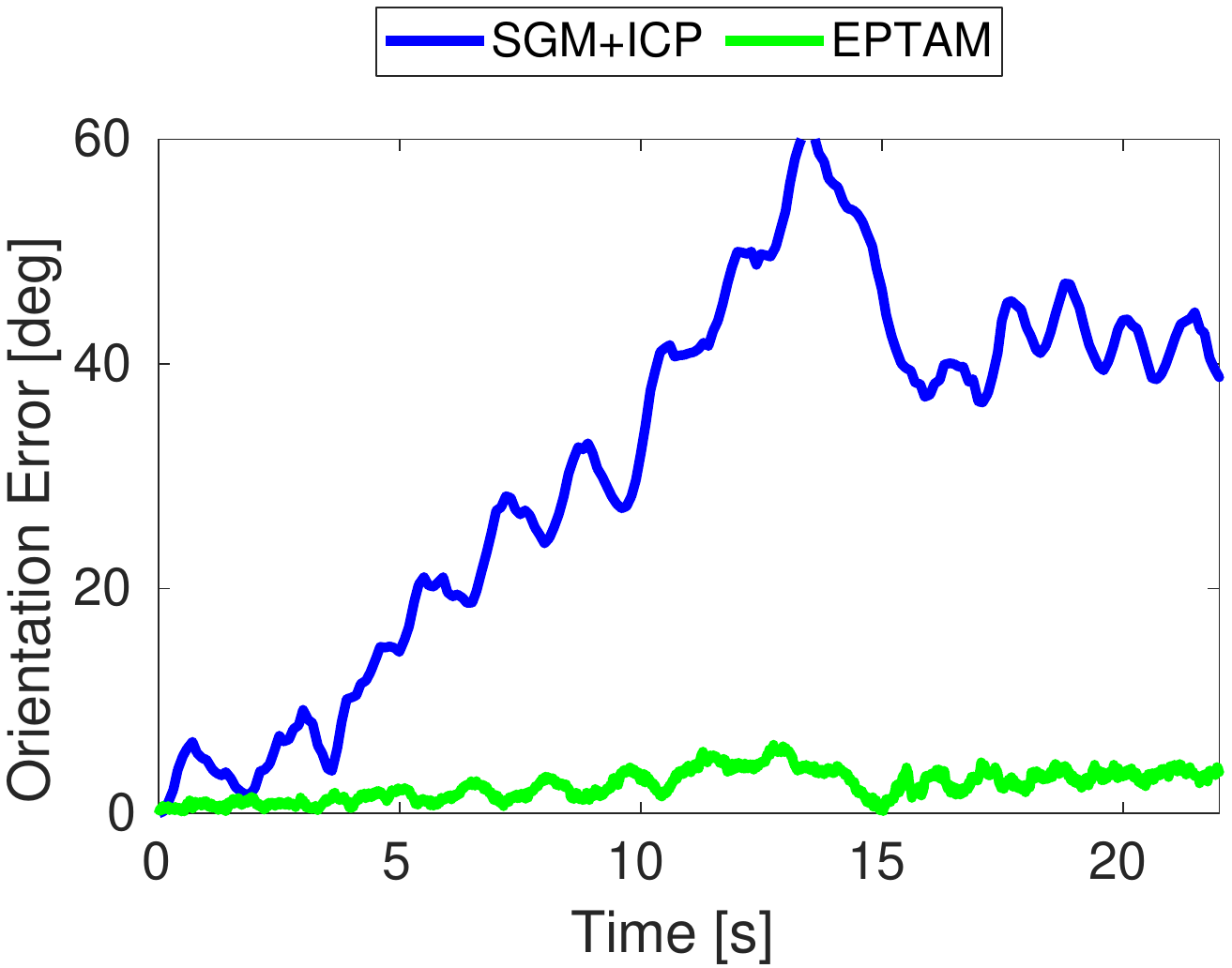}
  \\
   \includegraphics[trim={0 0 0 1.18cm},clip,width=\linewidth]{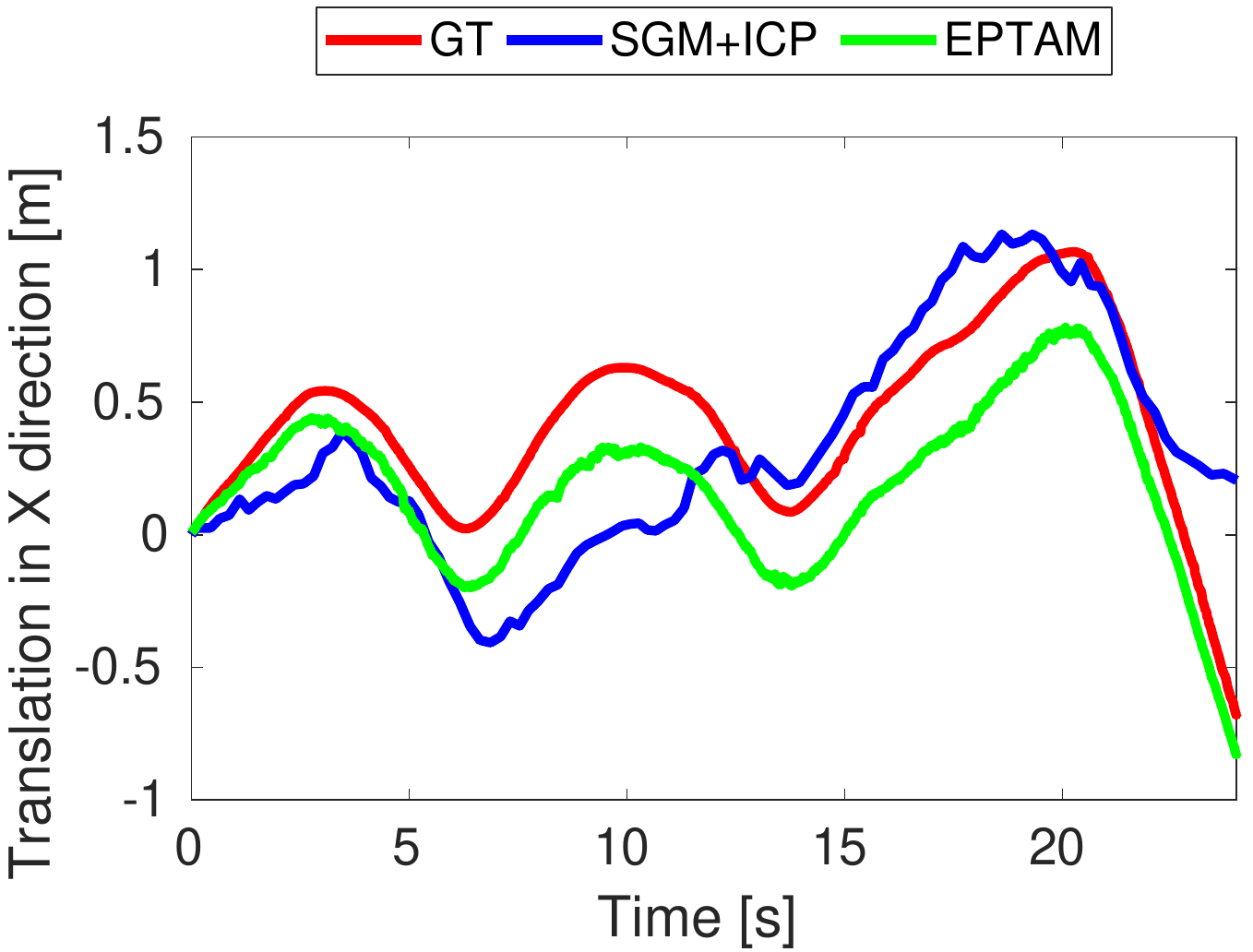}
  &\includegraphics[trim={0 0 0 1.18cm},clip,width=\linewidth]{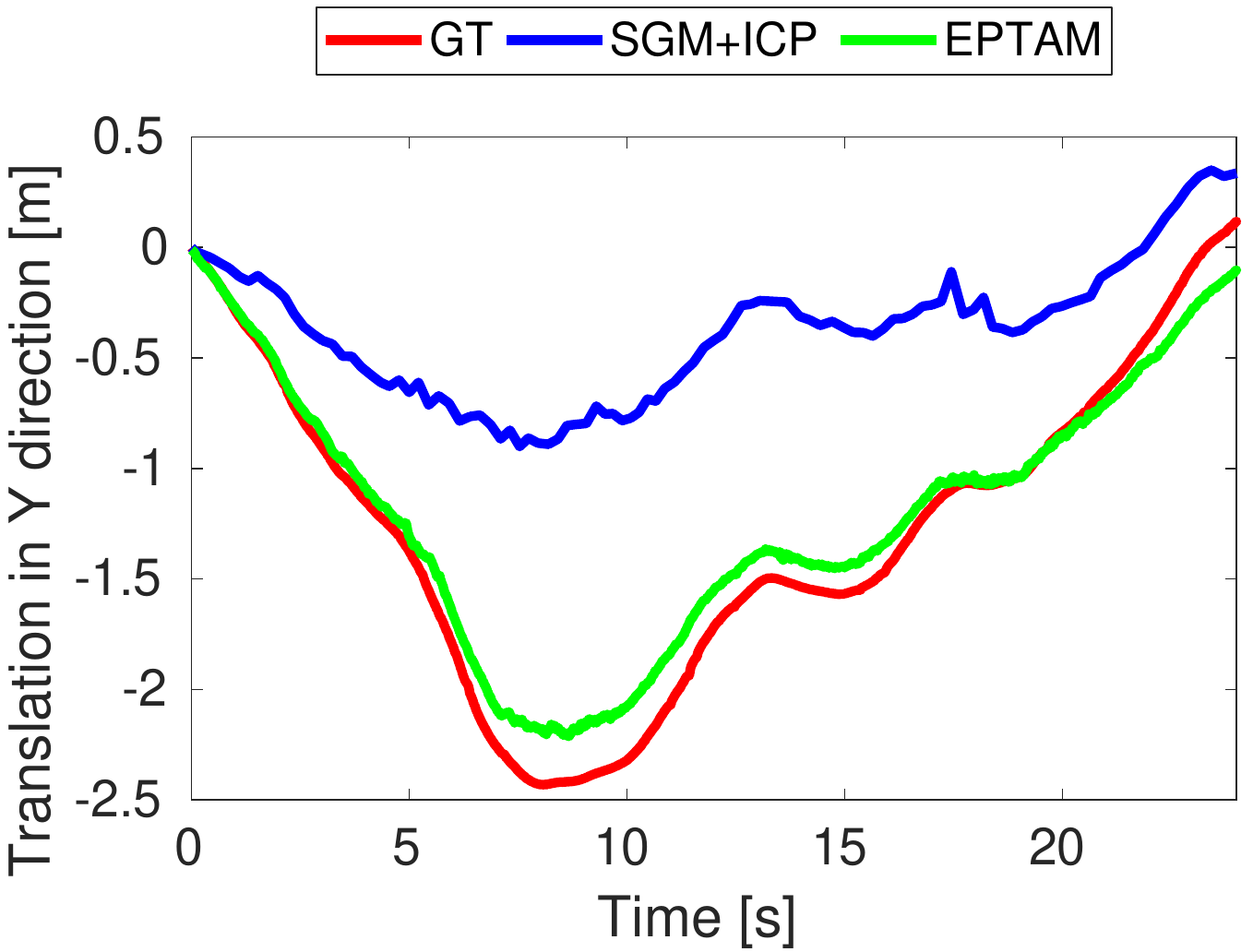}
  &\includegraphics[trim={0 0 0 1.18cm},clip,width=\linewidth]{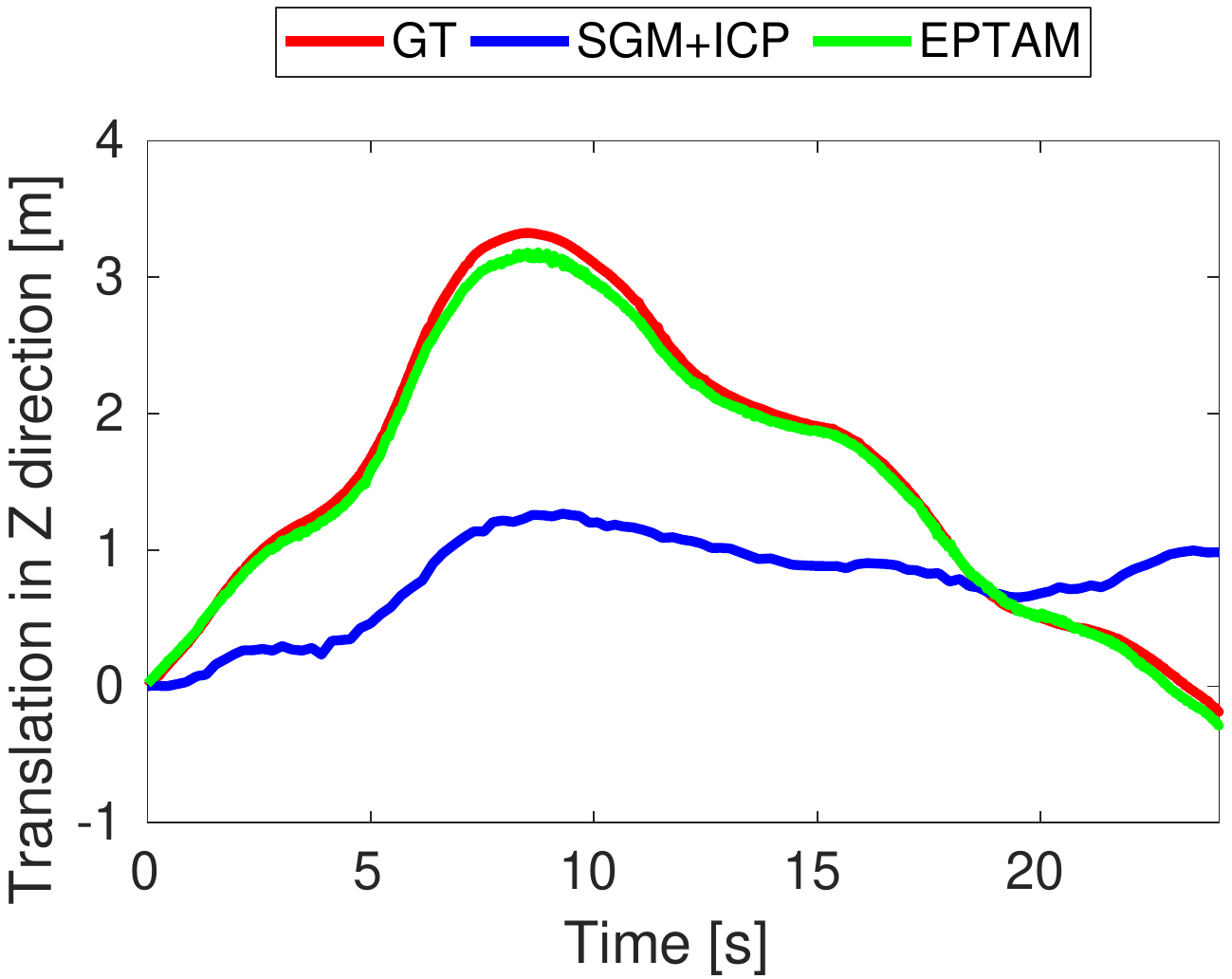}
  &\includegraphics[trim={0 0 0 1.18cm},clip,width=\linewidth]{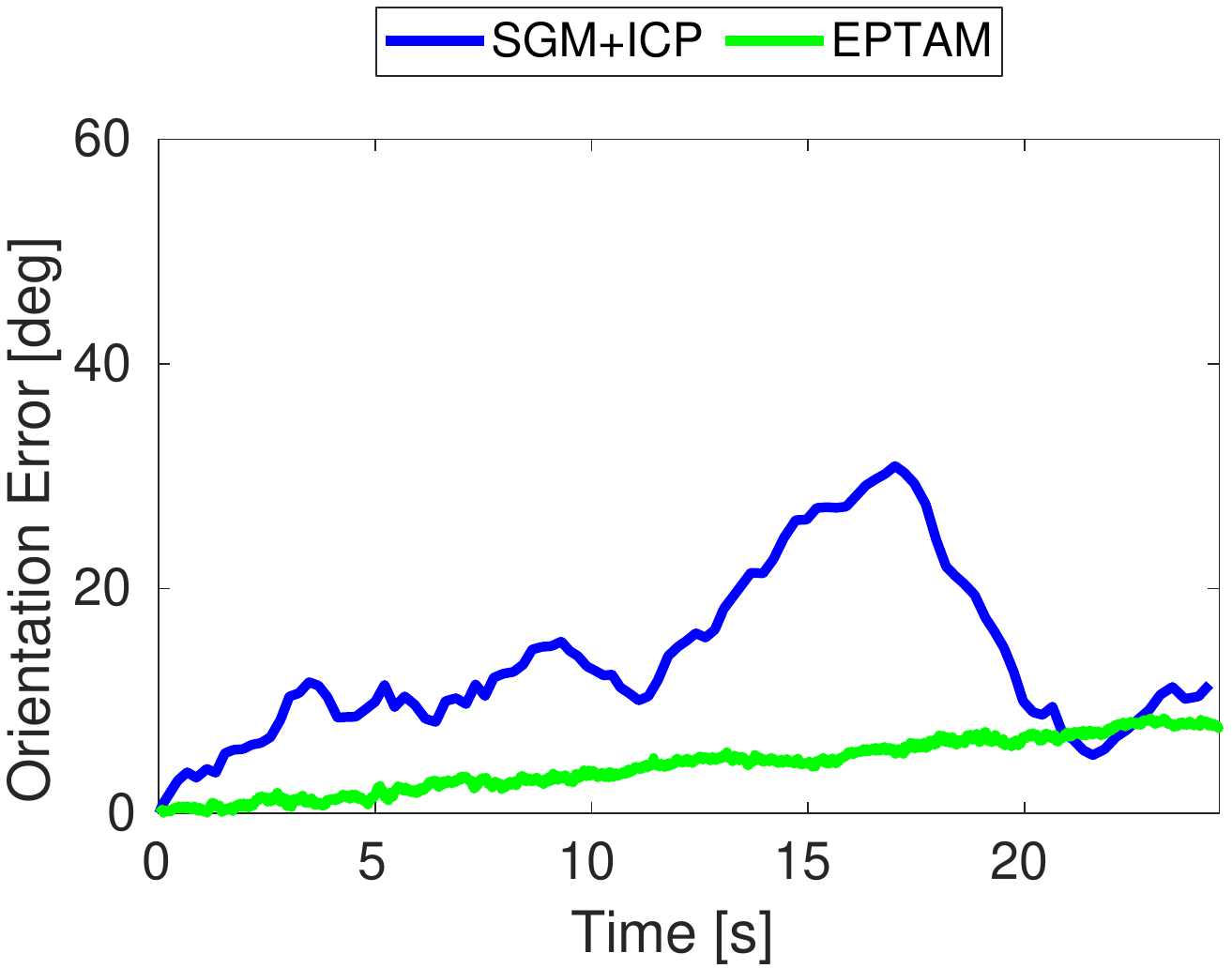}
  \\
   \includegraphics[trim={0 0 0 1.18cm},clip,width=\linewidth]{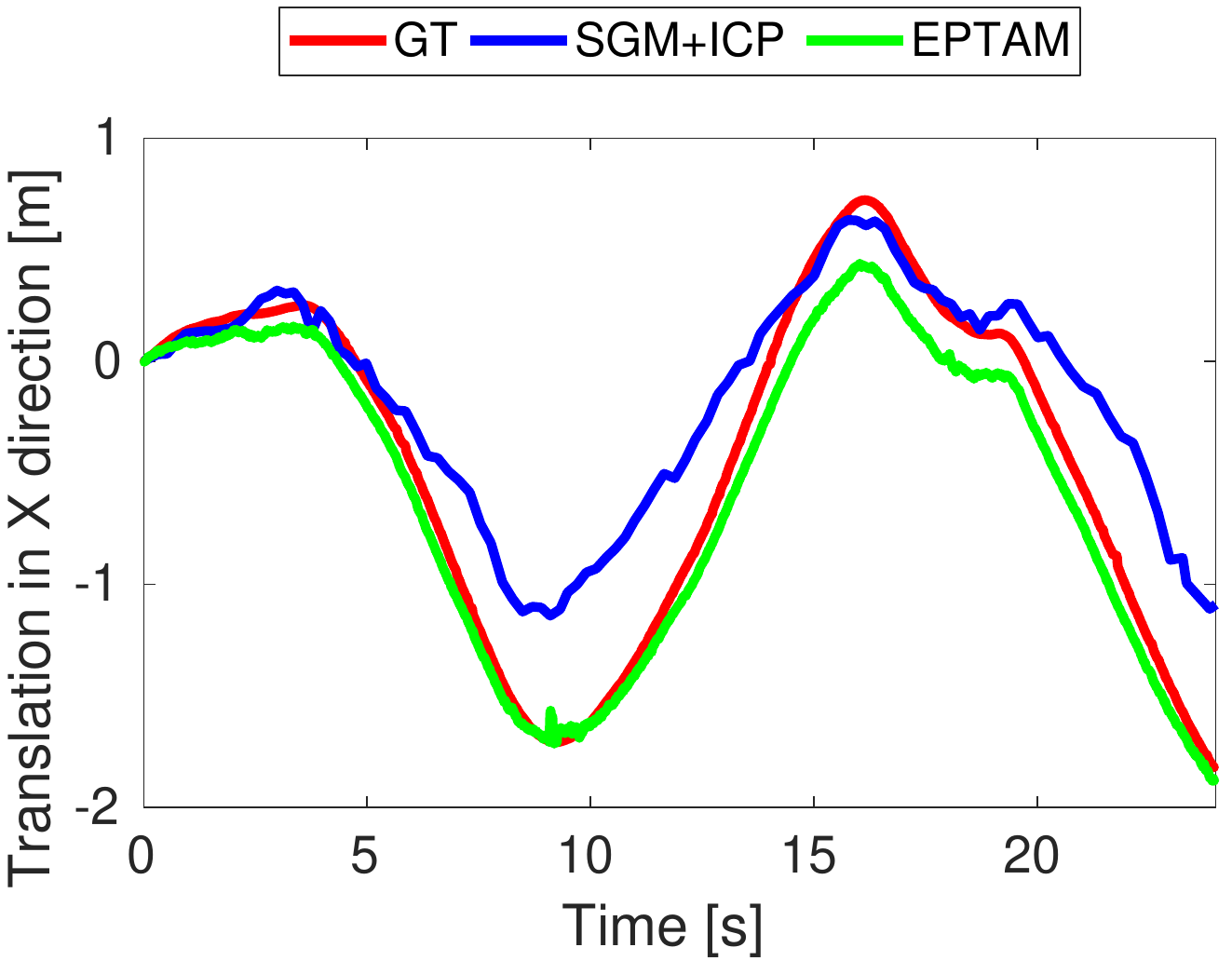}
  &\includegraphics[trim={0 0 0 1.18cm},clip,width=\linewidth]{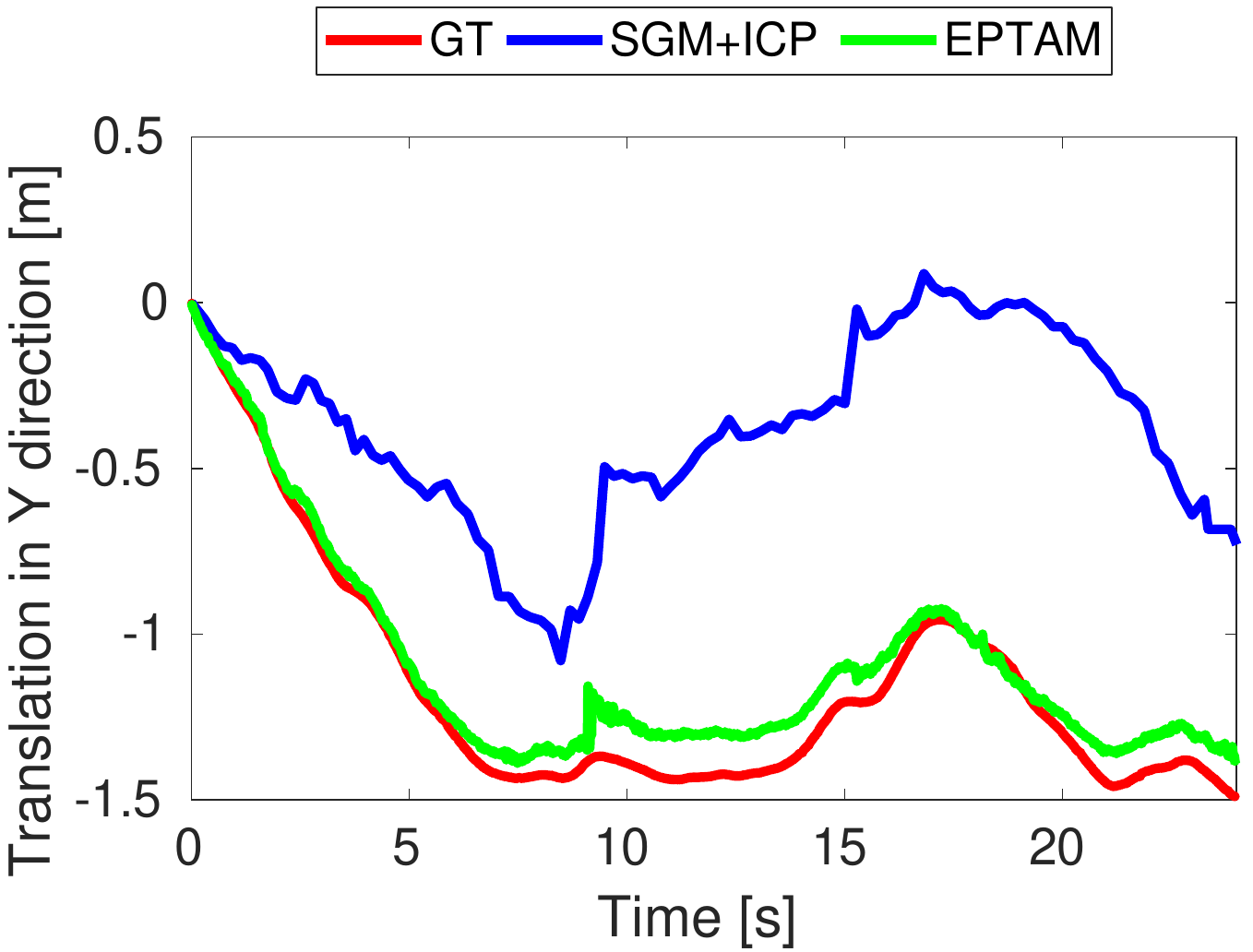}
  &\includegraphics[trim={0 0 0 1.18cm},clip,width=\linewidth]{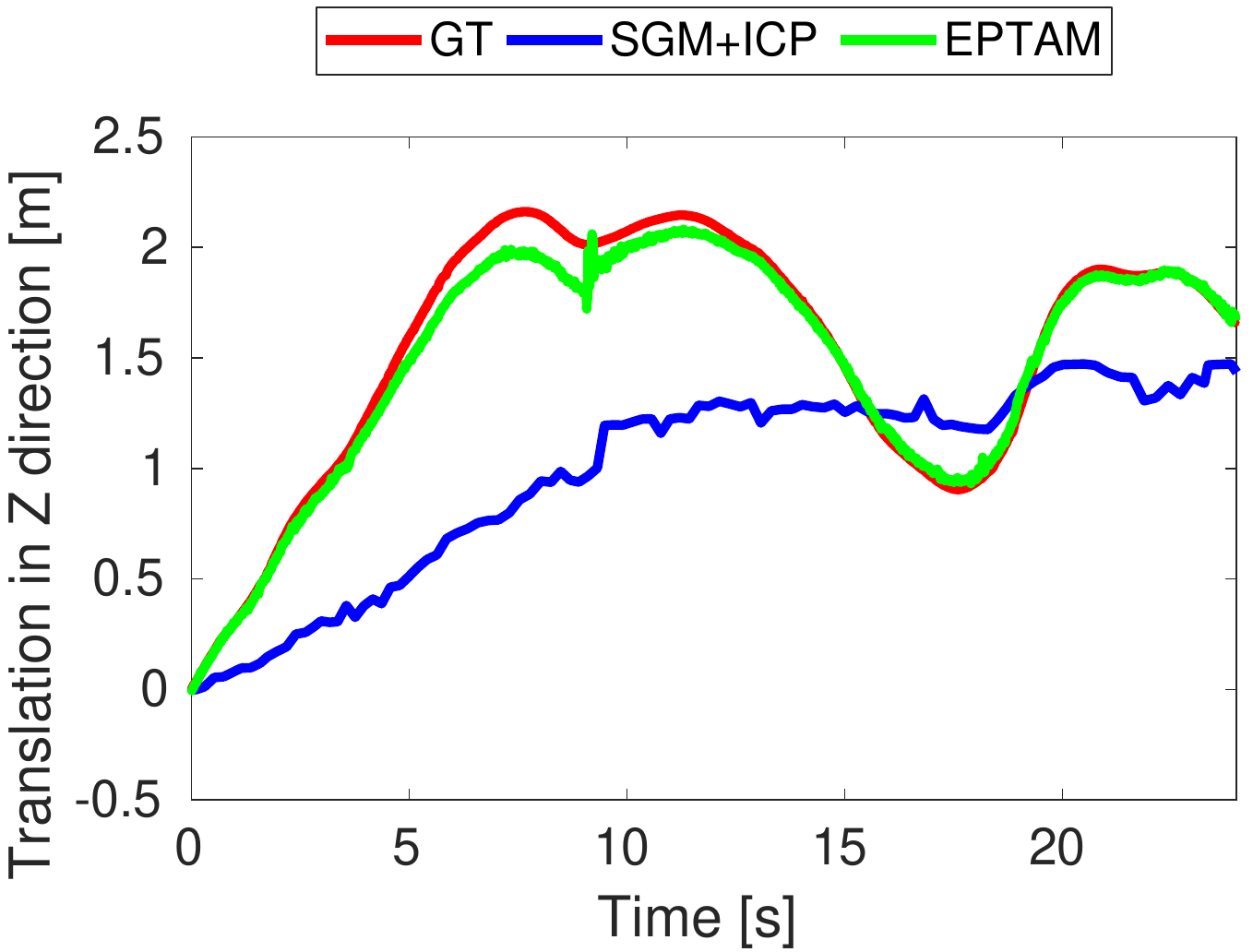}
  &\includegraphics[trim={0 0 0 1.18cm},clip,width=\linewidth]{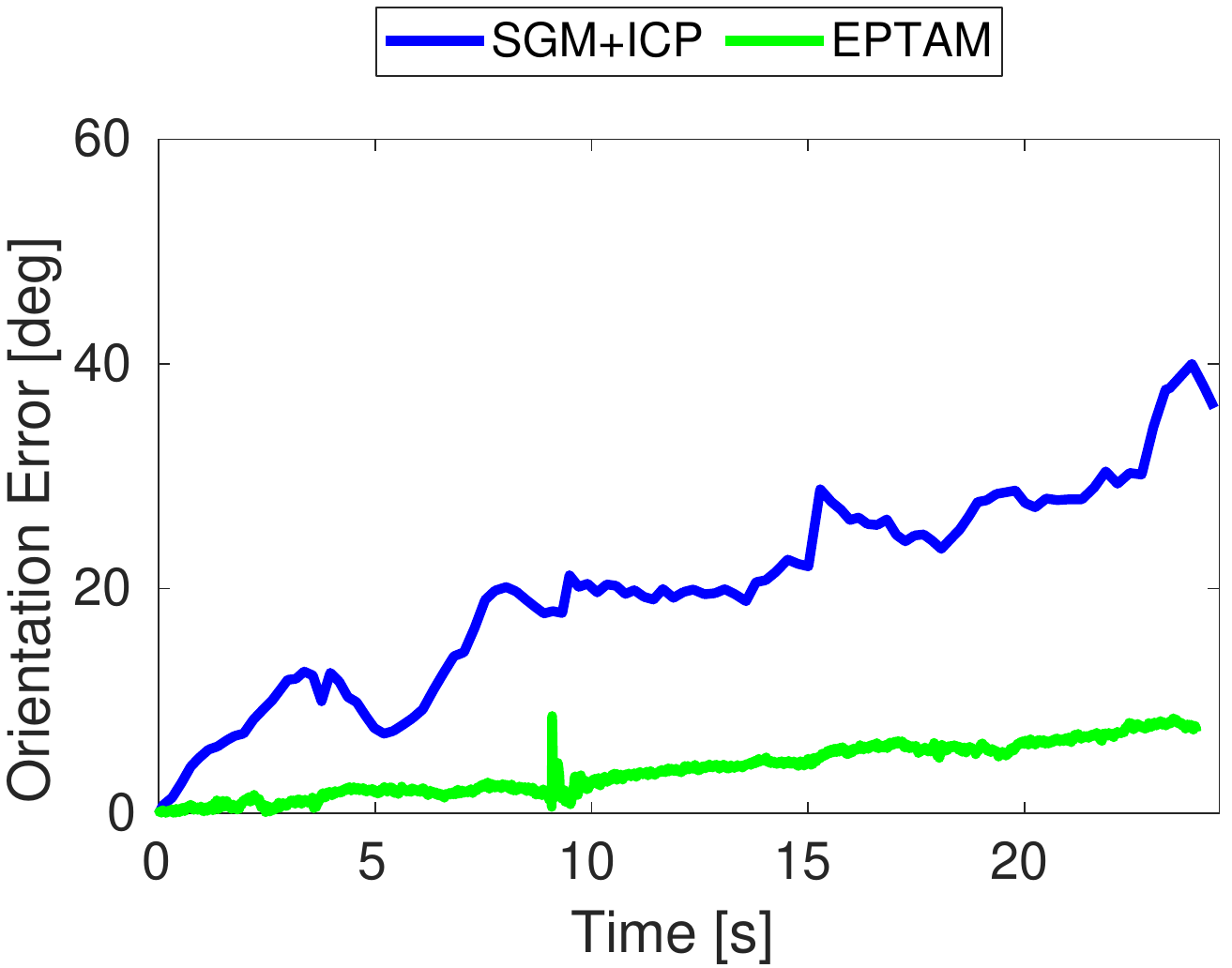}
  \\
  \end{tabular}
  \includegraphics[trim={0 9.5cm 0 0},clip,width=0.9\columnwidth]{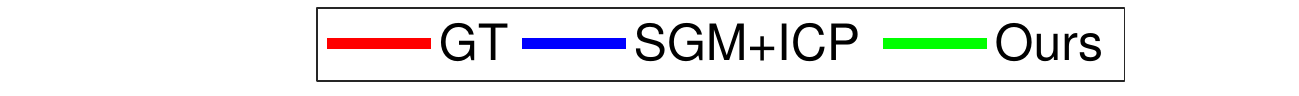}
  }
  \caption{\emph{Tracking - DoF plots}. 
  Comparison of two tracking methods against the ground truth camera trajectory provided by the motion capture system.
  Columns 1 to 3 show the translational degrees of freedom (in meters). 
  The last column shows rotational error in terms of the geodesic distance in \emph{SO(3)} (the angle of the relative rotation between the ground truth rotation and the estimated one).
  Each row corresponds to a different sequence: \rpgbin{}, \rpgbox{}, \rpgdesk{}, \rpgmonitor{}, \upennflyOne{} and \upennflyThree{}, respectively.
  The ground truth is depicted with red color (\textcolor{red}{\textbf{---}}), the ``SGM+ICP'' method with blue (\textcolor{blue}{\textbf{---}}) and our method with green~(\textcolor{green}{\textbf{---}}).
  In the error plots the ground truth corresponds to the reference, i.e., zero.
   The \emph{rpg} sequences~\cite{Zhou18eccv} are captured with a hand-held stereo rig moving under a locally loopy behavior (top four rows).
   In contrast, the \emph{upenn\_flying} sequences~\cite{Zhu18ral} are acquired using a stereo rig mounted on a drone which switches between hovering and moving dominantly in a translating manner (bottom two rows). 
  }
  \label{fig:tracking:dof-plots}
\end{figure*}
We also evaluate the proposed system on the \emph{hkust\_lab} sequence collected using our stereo event-camera rig.
The scene represents a cluttered environment which consists of various machine facilities.
The stereo rig was hand-held and moved from left to right under a locally loopy behavior.
The 3D point cloud together with the trajectory of the sensor are displayed in Fig.~\ref{fig:system-hkust}.
Additionally, the estimated inverse depth maps at selected views are visualized.
The live demonstration can be found in the supplemental video.

\subsection{Experiments in Low Light and HDR Environments}
\label{sec:experiments:HDR}
In addition to the evaluation under normal illumination conditions, we test the VO system in difficult conditions for frame-based cameras.
To this end, we run the algorithm on two sequences collected in a dark room. 
One of them is lit with a lamp to increase the range of scene brightness variations, creating high dynamic range conditions.
Results are shown in Fig.~\ref{fig:HDR-experiment}.
Under such conditions, the frame-based sensor of the DAVIS (with \SI{55}{\decibel} dynamic range) can barely see anything in the dark regions using its built-in auto-exposure, which would lead to failure of VO pipelines working on this visual modality.
By contrast, our event-based method is able to work robustly in these challenging illumination conditions due to the natural HDR properties of event cameras (\SI{120}{\decibel} range).
\global\long\def\figHeightHDR{2.48cm}
\global\long\def\figWidth{0.18\linewidth}
\begin{figure*}[h!]
	\centering
    {\small
    \setlength{\tabcolsep}{2pt}
	\begin{tabular}{
	>{\centering\arraybackslash}m{0.5cm} 
	>{\centering\arraybackslash}m{\figWidth} 
	>{\centering\arraybackslash}m{\figWidth}
	>{\centering\arraybackslash}m{\figWidth} 
	>{\centering\arraybackslash}m{\figWidth} 
	>{\centering\arraybackslash}m{\figWidth}}
		& DAVIS frame & Time surface (left) & Inverse depth map & Reprojected map & 3D reconstruction
		\\\addlinespace[1ex]

        \rotatebox{90}{\makecell{Without lamp}}
        &\includegraphics[height=\figHeightHDR]{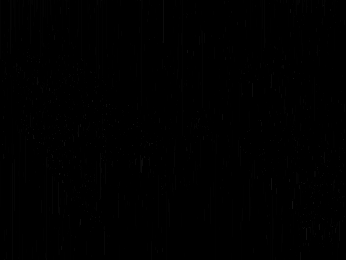}
        &\includegraphics[height=\figHeightHDR]{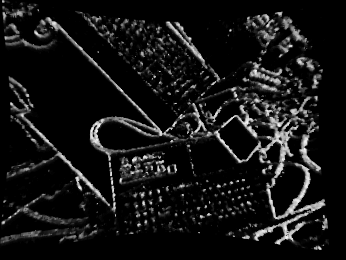}
        &\includegraphics[height=\figHeightHDR]{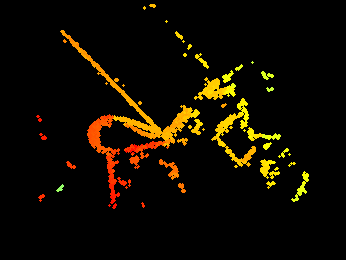}
        &\includegraphics[height=\figHeightHDR]{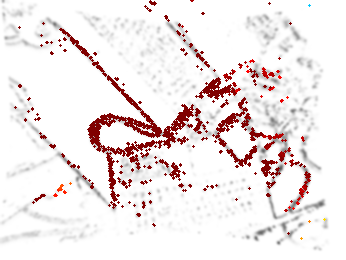}
        &\includegraphics[height=\figHeightHDR]{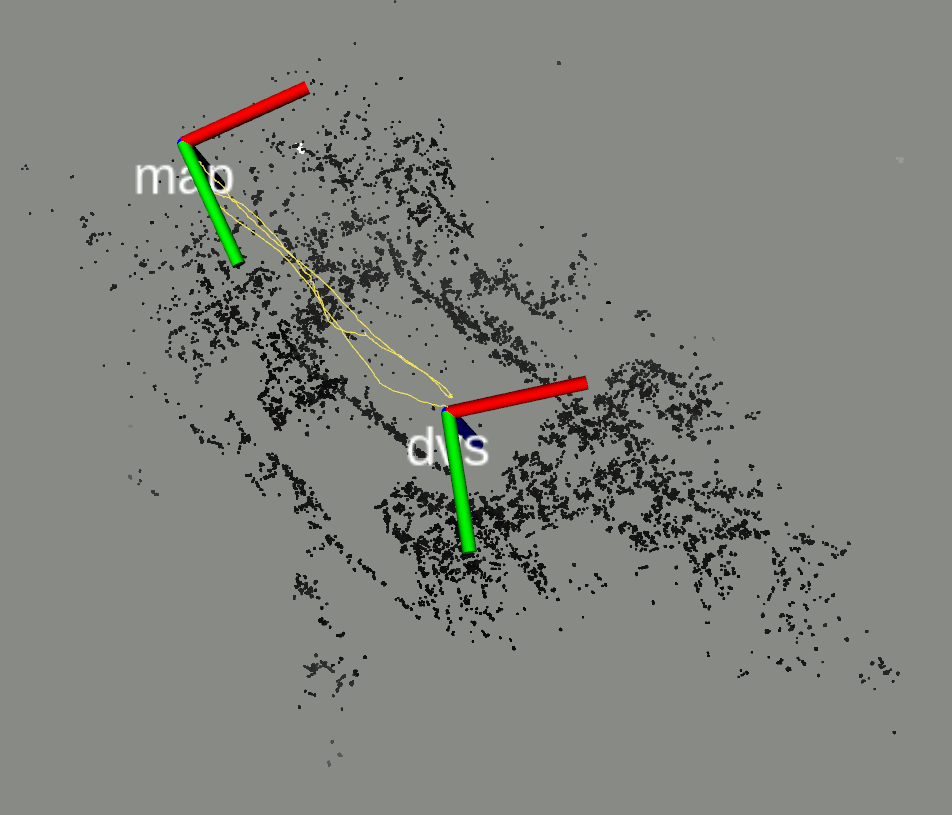}
        \\
  
        \rotatebox{90}{\makecell{With lamp}}
        &\includegraphics[height=\figHeightHDR]{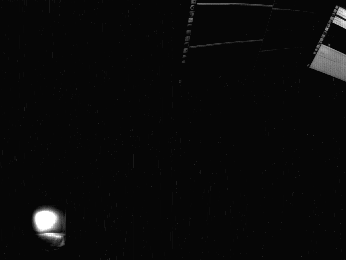}
        &\includegraphics[height=\figHeightHDR]{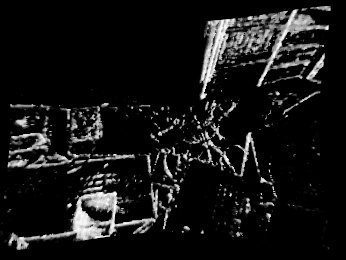}
        &\includegraphics[height=\figHeightHDR]{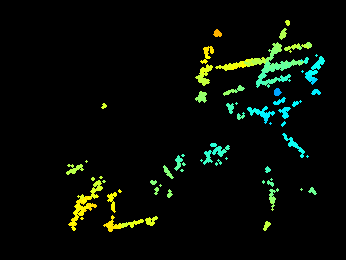}
        &\includegraphics[height=\figHeightHDR]{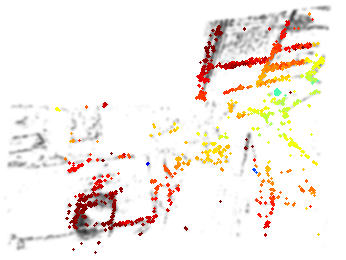}
        &\includegraphics[height=\figHeightHDR]{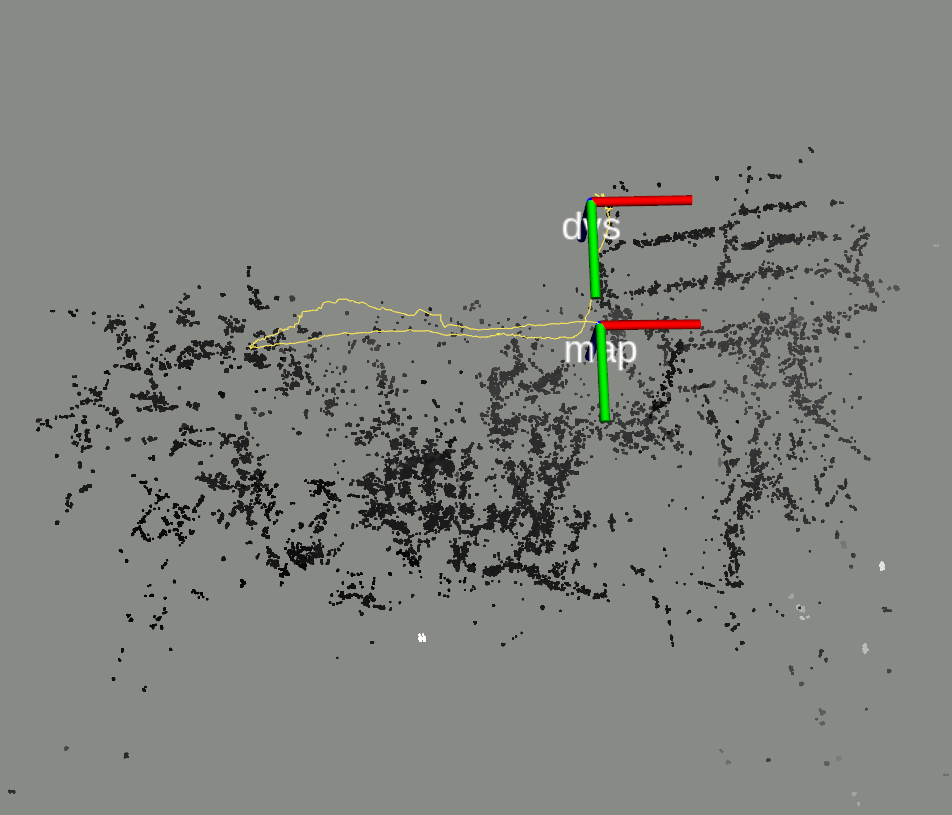}
        \\
	\end{tabular}
	}
   \caption{\emph{Low light and HDR scenes.}
   Top row: results in a dark room; 
   Bottom row: results in a dark room with a directional lamp.
   From left to right: grayscale frames (for visualization purpose only), time surfaces, estimated depth maps, reprojected maps on time surface negatives (tracking), and 3D reconstruction with overlaid camera trajectory estimates, respectively.
   }
  \label{fig:HDR-experiment}	
\end{figure*}

\subsection{Computational Performance}
\label{sec:experiments:timing}
\begin{table}[t]
\centering
\caption{Computational performance}
\begin{adjustbox}{max width=\linewidth}
\setlength{\tabcolsep}{3pt}
\begin{tabular}{@{}lll@{}}
\toprule
Node (\#Threads) & Function & Time (\si{\milli\second}) \\
\midrule
\textbf{Time surfaces} (1) & Exponential decay  &    $5-10$ \\
\textbf{Initialize depth} (1) & SGM \& masking &    $12-17$ \\
\textbf{Mapping} (4)        & Event matching & $6$ ($\sim$ 1000 events) \\
                            & Depth optimization & $15$ ($\sim$ 500 events) \\
                            & Depth fusion   & $20$ ($\sim$ 60000 fusions) \\
\textbf{Tracking} (2)       & Non-linear solver & $10$ (300 points $\times$ 5 iterations) \\
\bottomrule
\end{tabular}
\end{adjustbox}
\label{tab:computational-cost}
\end{table}
The proposed stereo visual odometry system is implemented in C++ on ROS and runs in real-time on a laptop with an Intel Core i7-8750H CPU.
Its computational performance is summarized in Table~\ref{tab:computational-cost}.
To accelerate processing, some nodes (mapping and tracking) are implemented with hyper-threading technology.
The number of threads used by each node is indicated in parentheses next to the name of the node.

The creation of the time-surface maps takes about \SIrange{5}{10}{\milli\second}, depending on the sensor resolution.
The initialization node, active only while bootstrapping, takes \SIrange{12}{17}{\milli\second} (up to sensor resolution) to produce the first local map (depth map given by the SGM method and masked with an event map). %

The mapping node uses 4 threads and takes about \SI{41}{\milli\second}, spent in three major functions.
($i$) The matching function takes $\approx$ \SI{6}{\milli\second} to search for 1000 corresponding patches across a pair of time surfaces.
The matching success rate is $\approx$ \SIrange{40}{50}{\percent}, depending on how well the spatio-temporal consistency holds in the data.
($ii$) The depth refinement function returns 500 inverse depth estimates in \SI{15}{\milli\second}.
($iii$) The fusion function (propagation and update steps) does 60000 operations in \SI{20}{\milli\second}.
Thus, the mapping node runs at \SI{20}{\hertz} typically.

\begin{figure}[t]
  \centering
  \subfigure[t][\small{500 events.}]{
  \includegraphics[width=0.31\columnwidth]{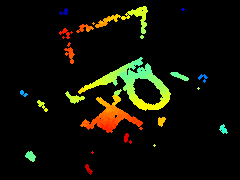}
  \label{fig: reconstruction density with 500 events}}\!\!\subfigure[t][\small{1000 events.}]{
  \includegraphics[width=0.31\columnwidth]{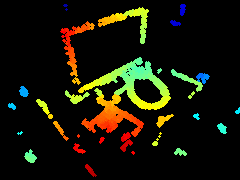}
  \label{fig: reconstruction density with 1000 events}}\!\!\subfigure[t][\small{2000 events.}]{
  \includegraphics[width=0.31\columnwidth]{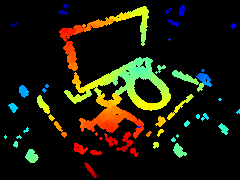}
  \label{fig: reconstruction density with 2000 events}}
  \caption{Influence of the number of events used for (inverse) depth estimation on the density of the fused depth map.}
  \label{fig:evaluation:reconstruction-density}
\end{figure}

Regarding the choice for the number of events being processed in the inverse depth estimation (\ie, 1000 as mentioned above), we justify it by showing its influence on the reconstruction density of the estimated depth maps.
Fig.~\ref{fig:evaluation:reconstruction-density} shows mapping results using 500, 1000 and 2000 events for inverse depth estimation.
We randomly pick these events out of the latest 10000 events.
For a fair comparison, the number of fusion steps remains constant. 
As it is observed, the more events are used the more dense the inverse depth map becomes. 
The map obtained using 500 events is the sparsest.
We notice that using 1000 or 2000 events produces nearly the same reconstruction density. 
However the latter (2000 events) is computationally more expensive (computation time is approximately proportional to the number of events); hence, for real-time performance opt for 1000 events.

The tracking node uses 2 threads and takes $\approx$\SI{10}{\milli\second} to solve the pose estimation problem using an IRLS solver 
(a batch of 300 points are randomly sampled in each iteration and at most five iterations are performed).
Hence, it can run up to \SI{100}{\hertz}.

\subsection{Discussion: Missing Edges in Reconstructions}
\label{sec:experiments:missing-edges}
\begin{figure}[t]
  \centering
  \subfigure[t][\small{Time surface}]{
  \includegraphics[width=0.22\textwidth]{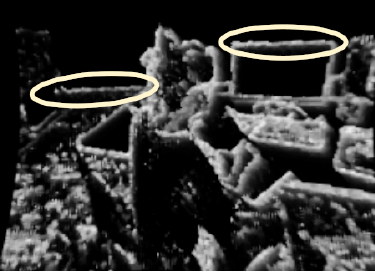}
  \label{fig:mapping:missing-edges:time-surface}}\,
  \subfigure[t][\small{(Inverse) depth uncertainty}]{
  \includegraphics[width=0.22\textwidth]{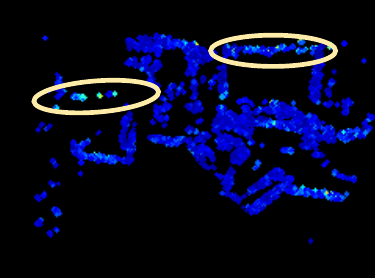}
  \label{fig:mapping:missing-edges:uncertainty-map}}\,
  \subfigure[t][\small{Depth map before pruning estimates with low uncertainty.}]{
  \includegraphics[width=0.22\textwidth]{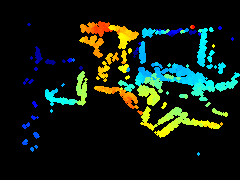} \label{fig:mapping:missing-edges:before-thresholding}}\,
  \subfigure[t][\small{Depth map after pruning estimates with low uncertainty.}]{
  \includegraphics[width=0.22\textwidth]{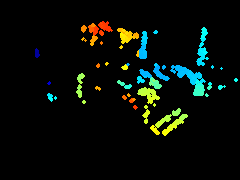}
  \label{fig:mapping:missing-edges:after-thresholding}}
  \caption{Depth uncertainty allows to filter unreliable estimates.}
  \label{fig:mapping:missing-edges}
\end{figure}
Here we note an effect that appears in some reconstructions, even when computed using ground truth poses (Section~\ref{sec:experiments:mapping:other-stereo}).
We observe that edges that are parallel to the baseline of the stereo rig,
such as the upper edge of the monitor in \rpgreader{} and the hoops on the barrel in \upennflyThree{} (Fig.~\ref{fig:mapping:depthmaps-grid}), are difficult to recover regardless of the motion.
All stereo methods suffer from this: 
although \GTS{}, SGM and CopNet can return depth estimates for those parallel structures, they are typically unreliable; 
our method is able to reason about uncertainty and therefore rejects such estimates.
In this respect, Fig.~\ref{fig:mapping:missing-edges} shows two horizontal patterns (highlighted with yellow ellipses in Fig.~\ref{fig:mapping:missing-edges:time-surface}) and their corresponding uncertainties (Fig.~\ref{fig:mapping:missing-edges:uncertainty-map}), which  are larger than those of other edges.
By thresholding on the depth uncertainty map (Fig.~\ref{fig:mapping:missing-edges:before-thresholding}), we obtain a more reliable albeit sparser depth map (Fig.~\ref{fig:mapping:missing-edges:after-thresholding}). %
Improving the completeness of reconstructions suffering from the above effect is left as future work.

\subsection{Dependency of Spatio-Temporal Consistency on Motion}
\label{sec:experiments:STC Degeneracy Analysis}
\begin{figure}[t]
  \centering
  \subfigure[t][\small{(Inverse) depth map under pure translation along $Y$ axis.}]{
  \includegraphics[width=0.22\textwidth]{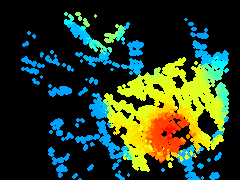}
  \label{fig:mapping:degeneracy:invDepth map translationY}}\,
  \subfigure[t][\small{(Inverse) depth map under pure rotation around $Z$ axis.}]{
  \includegraphics[width=0.22\textwidth]{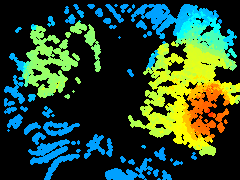}
  \label{fig:mapping:degeneracy:invDepth map rotationZ}}\\
  \subfigure[t][\small{Distribution of residuals for pure translation along $Y$ axis.}]{
  \includegraphics[width=0.22\textwidth]{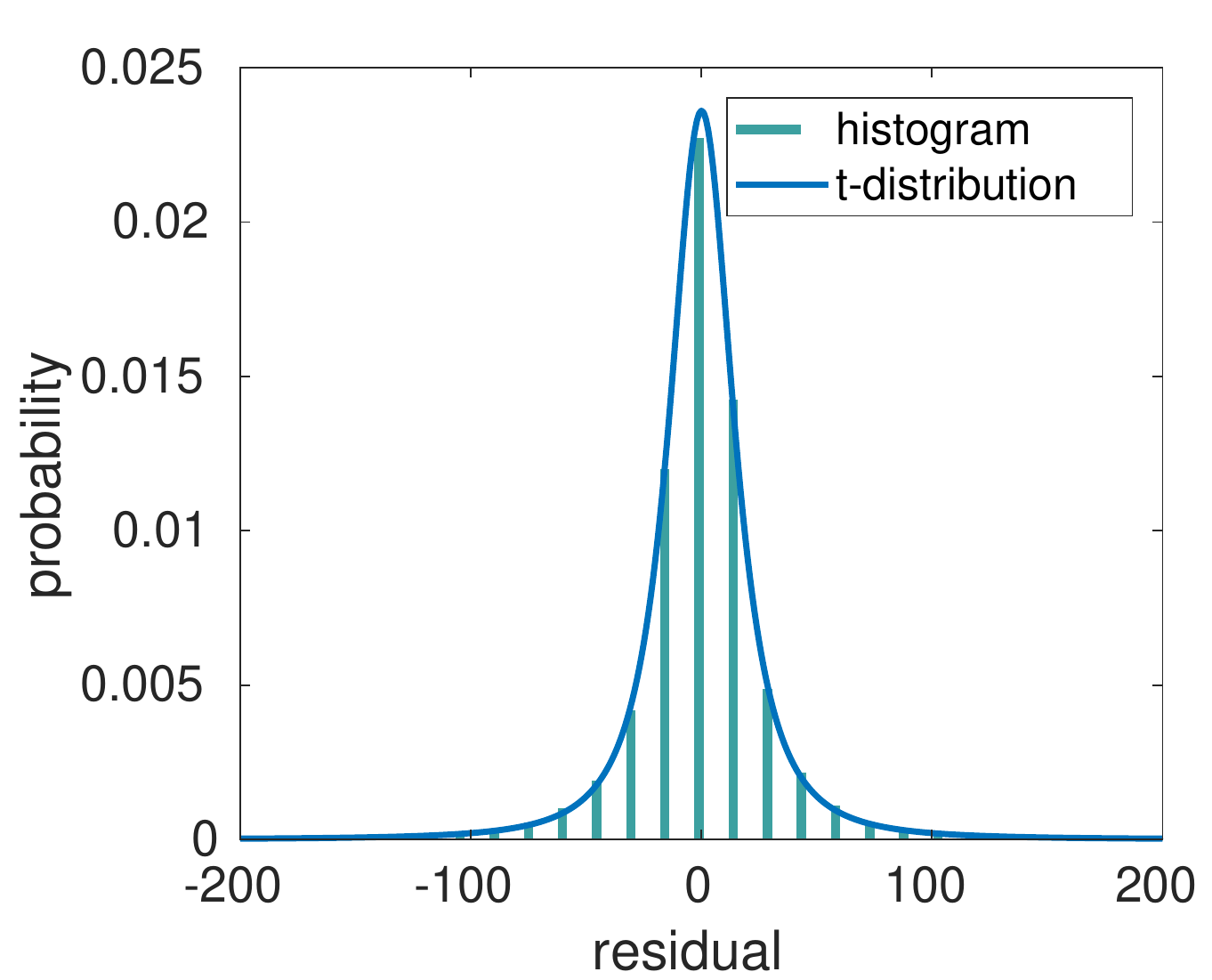}
  \label{fig:mapping:degeneracy:translationY}}\,
  \subfigure[t][\small{Distribution of residuals for pure rotation around $Z$ axis.}]{
  \includegraphics[width=0.22\textwidth]{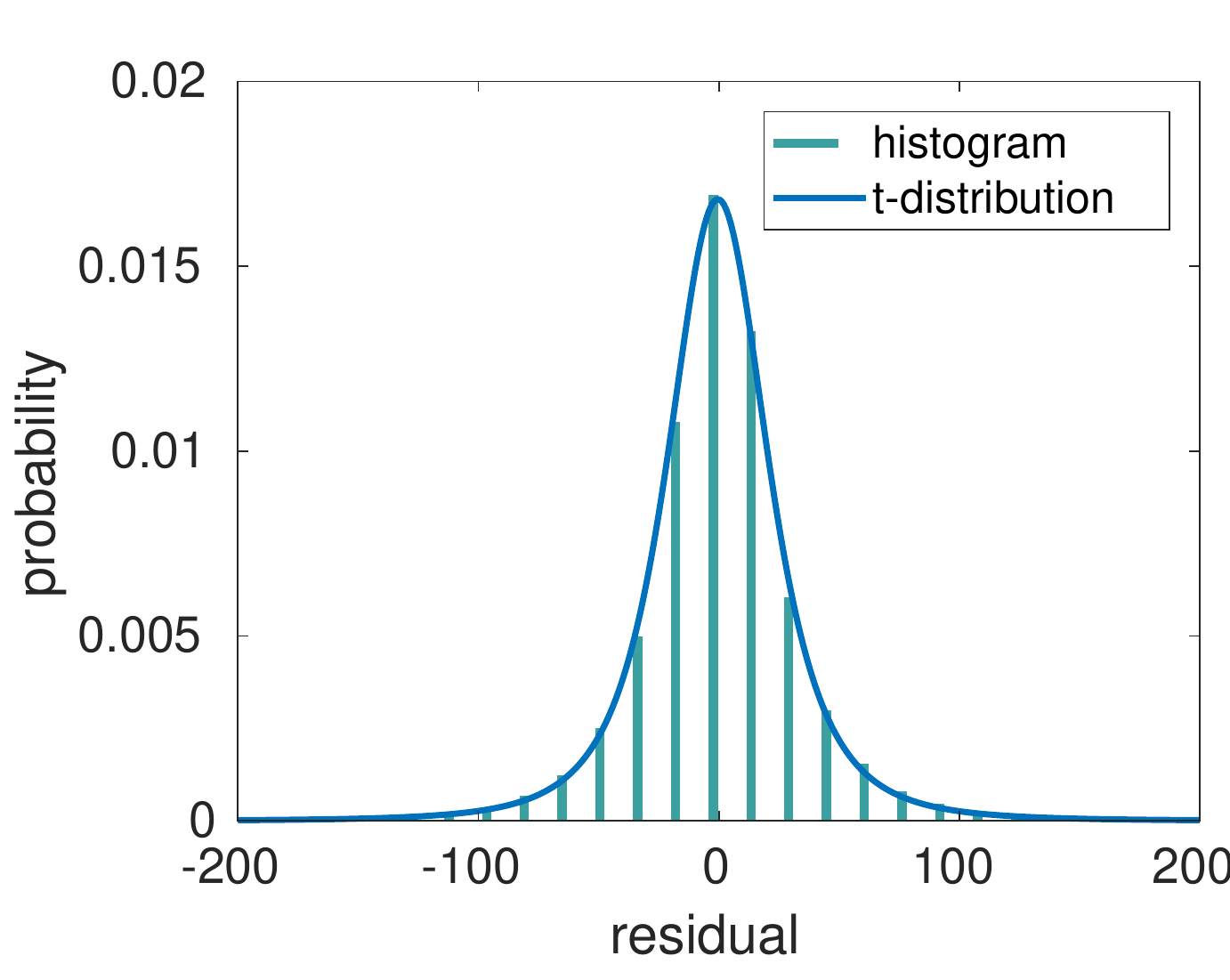}
  \label{fig:mapping:degeneracy:rotationZ}}
  \caption{Analysis of spatio-temporal consistency. 
  (a)-(b): Inverse depth estimates under two different types of motion.
  (c)-(d): Corresponding histograms of temporal residuals.
  Scene: \emph{toy\_room} in~\cite{Bryner19icra}.
  The corresponding videos can be found at \url{https://youtu.be/QY82AcX1LDo} (translation along $Y$ axis);
  and \url{https://youtu.be/RkxBn304gJI} (rotation around $Z$ axis).
  }
  \label{fig:mapping:degeneracy}
\end{figure}
Time surfaces are motion dependent, and consequently, even in the noise-free case the proposed spatio-temporal consistency criterion may not hold perfectly when the stereo rig undergoes some specific motions.
One extreme case could be a pure rotation of the left camera around its optical axis; thus the right camera would rotate and translate.
Intuitively, the additional translation component of the right camera would produce spatio-temporal inconsistency between the left-right time surfaces such that the mapping module would suffer.
To analyze the sensitivity of the mapping module with respect to the spatio-temporal consistency, we carried out the following experiment (Fig.~\ref{fig:mapping:degeneracy}).
We used an event camera simulator~\cite{Rebecq18corl} to generate sequences with perfect control over the motion.
Specifically, we generated sequences with pure rotation of the left camera around the optical axis ($Z$ axis), 
and compared the mapping results against those of pure translational motion along the $X$ or $Y$ axis of the camera.
The fused depth maps were slightly worse in the former case (partly because there are fewer events triggered around the centre of the image plane), 
but they were still accurate in most pixels (see Fig.~\ref{fig:mapping:degeneracy:invDepth map translationY} and \ref{fig:mapping:degeneracy:invDepth map rotationZ}).
Additionally, we analyzed the temporal inconsistency through the histogram of temporal residuals (like in Fig.~\ref{fig:temporal-residual-distribution}).
The histogram of residuals for the rotation around the $Z$ axis (Fig.~\ref{fig:mapping:degeneracy:rotationZ}) is broader than the one for translation around the $X$/$Y$ axes (Fig.~\ref{fig:mapping:degeneracy:translationY}). 
Numerically, the scale values of the $t$-distributions are $s_\text{trans\_Y} = 14.995$ and $s_\text{rot\_Z} = 21.838$. 
Compared to those in Fig.~\ref{fig:temporal-residual-distribution}, the residuals in Fig.~\ref{fig:mapping:degeneracy:rotationZ} are similar to those of the \upennflyOne\, sequence.

We conclude that, in spite of time surfaces being motion dependent,
we did not observe a significant temporal inconsistency that would break down the system in a priori difficult motions for stereo.
Actually the proposed method performed well in practice, as shown in all previous experiments with real data.
We leave a more theoretical and detailed analysis of such motions for future research since we consider this work %
to address the most general motion case.

\ifclearsectionlook\cleardoublepage\fi \section{Conclusion}
\label{sec:conclusion}

This paper has presented a complete event-based stereo visual odometry system for a pair of calibrated and synchronized event cameras in stereo configuration.
To the best of our knowledge, this is the first published work that tackles this problem.
The proposed mapping method is based on the optimization of an objective function designed to measure spatio-temporal consistency across stereo event streams.
To improve density and accuracy of the recovered 3D structure, a fusion strategy based on the learned probabilistic characteristics of the estimated inverse depth has been carried out.
The tracking method is based on $\ThreeDtoTwoD$ registration that leverages the inherent distance field nature of a compact and efficient event representation (time surfaces).
Extensive experimental evaluation, on publicly available datasets and our own, has demonstrated the versatility of our system.
Its performance is comparable with mature, state-of-the-art VO methods for frame-based cameras in normal conditions.
We have also demonstrated the potential advantages that event cameras bring to stereo SLAM in difficult illumination conditions.
The system is computationally efficient and runs in real-time on a standard CPU.
The software, design of the stereo rig and datasets used for evaluation have been open sourced.
Future work may include fusing the proposed method with inertial observations (i.e., Event-based Stereo Visual-Inertial Odometry) 
and investigating novel methods for finding correspondences in time on each event camera (i.e., ``temporal'' event-based stereo). 
These are closely related topics to the problem here addressed of Event-based Stereo VO.

\section*{Acknowledgment}
The authors would like to thank Ms. Siqi Liu for the help in data collection,
and Dr. Alex Zhu for providing the CopNet baseline~\cite{Zhou18eccv,Piatkowska17cvprw} and assistance in using the dataset~\cite{Zhu18ral}.
We also thank the editors and anonymous reviewers of IEEE TRO for their suggestions, which led us to improve the paper. %

\bibliographystyle{IEEEtran}

\begin{thebibliography}{10}
\providecommand{\url}[1]{#1}
\csname url@samestyle\endcsname
\providecommand{\newblock}{\relax}
\providecommand{\bibinfo}[2]{#2}
\providecommand{\BIBentrySTDinterwordspacing}{\spaceskip=0pt\relax}
\providecommand{\BIBentryALTinterwordstretchfactor}{4}
\providecommand{\BIBentryALTinterwordspacing}{\spaceskip=\fontdimen2\font plus
\BIBentryALTinterwordstretchfactor\fontdimen3\font minus
  \fontdimen4\font\relax}
\providecommand{\BIBforeignlanguage}[2]{{%
\expandafter\ifx\csname l@#1\endcsname\relax
\typeout{** WARNING: IEEEtran.bst: No hyphenation pattern has been}%
\typeout{** loaded for the language `#1'. Using the pattern for}%
\typeout{** the default language instead.}%
\else
\language=\csname l@#1\endcsname
\fi
#2}}
\providecommand{\BIBdecl}{\relax}
\BIBdecl

\bibitem{Lichtsteiner08ssc}
P.~Lichtsteiner, C.~Posch, and T.~Delbruck, ``{A 128$\times$128 120 dB 15
  $\mu$s latency asynchronous temporal contrast vision sensor},'' \emph{{IEEE}
  J. Solid-State Circuits}, vol.~43, no.~2, pp. 566--576, 2008.

\bibitem{Gallego20pami}
G.~Gallego, T.~Delbruck, G.~Orchard, C.~Bartolozzi, B.~Taba, A.~Censi,
  S.~Leutenegger, A.~Davison, J.~Conradt, K.~Daniilidis, and D.~Scaramuzza,
  ``Event-based vision: A survey,'' \emph{{IEEE} Trans. Pattern Anal. Mach.
  Intell.}, 2020.

\bibitem{Lagorce15tnnls}
X.~Lagorce, C.~Meyer, S.-H. Ieng, D.~Filliat, and R.~Benosman, ``Asynchronous
  event-based multikernel algorithm for high-speed visual features tracking,''
  \emph{{IEEE} Trans. Neural Netw. Learn. Syst.}, vol.~26, no.~8, pp.
  1710--1720, Aug. 2015.

\bibitem{Zhu17icra}
A.~Z. Zhu, N.~Atanasov, and K.~Daniilidis, ``Event-based feature tracking with
  probabilistic data association,'' in \emph{{IEEE} Int. Conf. Robot. Autom.
  (ICRA)}, 2017, pp. 4465--4470.

\bibitem{Gehrig19ijcv}
D.~Gehrig, H.~Rebecq, G.~Gallego, and D.~Scaramuzza, ``{EKLT}: Asynchronous
  photometric feature tracking using events and frames,'' \emph{Int. J. Comput.
  Vis.}, vol. 128, pp. 601--618, 2020.

\bibitem{Mueggler15rss}
E.~Mueggler, G.~Gallego, and D.~Scaramuzza, ``Continuous-time trajectory
  estimation for event-based vision sensors,'' in \emph{Robotics: Science and
  Systems (RSS)}, 2015.

\bibitem{Gallego17pami}
G.~Gallego, J.~E.~A. Lund, E.~Mueggler, H.~Rebecq, T.~Delbruck, and
  D.~Scaramuzza, ``Event-based, 6-{DOF} camera tracking from photometric depth
  maps,'' \emph{{IEEE} Trans. Pattern Anal. Mach. Intell.}, vol.~40, no.~10,
  pp. 2402--2412, Oct. 2018.

\bibitem{Gallego17ral}
G.~Gallego and D.~Scaramuzza, ``Accurate angular velocity estimation with an
  event camera,'' \emph{{IEEE} Robot. Autom. Lett.}, vol.~2, no.~2, pp.
  632--639, 2017.

\bibitem{Bryner19icra}
S.~Bryner, G.~Gallego, H.~Rebecq, and D.~Scaramuzza, ``Event-based, direct
  camera tracking from a photometric {3D} map using nonlinear optimization,''
  in \emph{{IEEE} Int. Conf. Robot. Autom. (ICRA)}, 2019, pp. 325--331.

\bibitem{Conradt09iscas}
J.~Conradt, M.~Cook, R.~Berner, P.~Lichtsteiner, R.~J. Douglas, and
  T.~Delbruck, ``A pencil balancing robot using a pair of {AER} dynamic vision
  sensors,'' in \emph{{IEEE} Int. Symp. Circuits Syst. (ISCAS)}, 2009, pp.
  781--784.

\bibitem{Delbruck13fns}
T.~Delbruck and M.~Lang, ``Robotic goalie with 3ms reaction time at 4\% {CPU}
  load using event-based dynamic vision sensor,'' \emph{Front. Neurosci.},
  vol.~7, p. 223, 2013.

\bibitem{Falanga20scirob}
D.~Falanga, K.~Kleber, and D.~Scaramuzza, ``Dynamic obstacle avoidance for
  quadrotors with event cameras,'' \emph{Science Robotics}, vol.~5, no.~40, p.
  eaaz9712, Mar. 2020.

\bibitem{Kim16eccv}
H.~Kim, S.~Leutenegger, and A.~J. Davison, ``Real-time {3D} reconstruction and
  6-{DoF} tracking with an event camera,'' in \emph{Eur. Conf. Comput. Vis.
  (ECCV)}, 2016, pp. 349--364.

\bibitem{Rebecq17ral}
H.~Rebecq, T.~Horstsch{\"a}fer, G.~Gallego, and D.~Scaramuzza, ``{EVO}: A
  geometric approach to event-based 6-{DOF} parallel tracking and mapping in
  real-time,'' \emph{{IEEE} Robot. Autom. Lett.}, vol.~2, no.~2, pp. 593--600,
  2017.

\bibitem{Rosinol18ral}
A.~{Rosinol Vidal}, H.~Rebecq, T.~Horstschaefer, and D.~Scaramuzza, ``Ultimate
  {SLAM}? combining events, images, and {IMU} for robust visual {SLAM} in {HDR}
  and high speed scenarios,'' \emph{{IEEE} Robot. Autom. Lett.}, vol.~3, no.~2,
  pp. 994--1001, Apr. 2018.

\bibitem{Mueggler18tro}
E.~Mueggler, G.~Gallego, H.~Rebecq, and D.~Scaramuzza, ``Continuous-time
  visual-inertial odometry for event cameras,'' \emph{{IEEE} Trans. Robot.},
  vol.~34, no.~6, pp. 1425--1440, Dec. 2018.

\bibitem{Weikersdorfer14icra}
D.~Weikersdorfer, D.~B. Adrian, D.~Cremers, and J.~Conradt, ``Event-based {3D}
  {SLAM} with a depth-augmented dynamic vision sensor,'' in \emph{{IEEE} Int.
  Conf. Robot. Autom. (ICRA)}, 2014, pp. 359--364.

\bibitem{Censi14icra}
A.~Censi and D.~Scaramuzza, ``Low-latency event-based visual odometry,'' in
  \emph{{IEEE} Int. Conf. Robot. Autom. (ICRA)}, 2014, pp. 703--710.

\bibitem{Kueng16iros}
B.~Kueng, E.~Mueggler, G.~Gallego, and D.~Scaramuzza, ``Low-latency visual
  odometry using event-based feature tracks,'' in \emph{IEEE/RSJ Int. Conf.
  Intell. Robot. Syst. (IROS)}, 2016, pp. 16--23.

\bibitem{Klein07ismar}
G.~Klein and D.~Murray, ``Parallel tracking and mapping for small {AR}
  workspaces,'' in \emph{{IEEE} ACM Int. Sym. Mixed and Augmented Reality
  (ISMAR)}, Nara, Japan, Nov. 2007, pp. 225--234.

\bibitem{Zhou18eccv}
Y.~Zhou, G.~Gallego, H.~Rebecq, L.~Kneip, H.~Li, and D.~Scaramuzza,
  ``Semi-dense {3D} reconstruction with a stereo event camera,'' in \emph{Eur.
  Conf. Comput. Vis. (ECCV)}, 2018, pp. 242--258.

\bibitem{Kogler11isvc}
J.~Kogler, M.~Humenberger, and C.~Sulzbachner, ``Event-based stereo matching
  approaches for frameless address event stereo data,'' in \emph{Int. Symp.
  Adv. Vis. Comput. (ISVC)}, 2011, pp. 674--685.

\bibitem{Rogister12tnnls}
P.~Rogister, R.~Benosman, S.-H. Ieng, P.~Lichtsteiner, and T.~Delbruck,
  ``Asynchronous event-based binocular stereo matching,'' \emph{{IEEE} Trans.
  Neural Netw. Learn. Syst.}, vol.~23, no.~2, pp. 347--353, 2012.

\bibitem{CamunasMesa14fns}
L.~A. Camunas-Mesa, T.~Serrano-Gotarredona, S.~H. Ieng, R.~B. Benosman, and
  B.~Linares-Barranco, ``On the use of orientation filters for {3D}
  reconstruction in event-driven stereo vision,'' \emph{Front. Neurosci.},
  vol.~8, p.~48, 2014.

\bibitem{Hartley03book}
R.~Hartley and A.~Zisserman, \emph{Multiple View Geometry in Computer
  Vision}.\hskip 1em plus 0.5em minus 0.4em\relax Cambridge University Press,
  2003, 2nd Edition.

\bibitem{Ieng18fnins}
S.-H. Ieng, J.~Carneiro, M.~Osswald, and R.~Benosman, ``Neuromorphic
  event-based generalized time-based stereovision,'' \emph{Front. Neurosci.},
  vol.~12, p. 442, 2018.

\bibitem{Posch11ssc}
C.~Posch, D.~Matolin, and R.~Wohlgenannt, ``A {QVGA} 143 {dB} dynamic range
  frame-free {PWM} image sensor with lossless pixel-level video compression and
  time-domain {CDS},'' \emph{{IEEE} J. Solid-State Circuits}, vol.~46, no.~1,
  pp. 259--275, Jan. 2011.

\bibitem{Piatkowska14msci}
E.~Piatkowska, A.~N. Belbachir, and M.~Gelautz, ``Cooperative and asynchronous
  stereo vision for dynamic vision sensors,'' \emph{Meas. Sci. Technol.},
  vol.~25, no.~5, p. 055108, Apr. 2014.

\bibitem{Firouzi16npl}
M.~Firouzi and J.~Conradt, ``Asynchronous event-based cooperative stereo
  matching using neuromorphic silicon retinas,'' \emph{Neural Proc. Lett.},
  vol.~43, no.~2, pp. 311--326, 2016.

\bibitem{Osswald17srep}
M.~Osswald, S.-H. Ieng, R.~Benosman, and G.~Indiveri, ``A spiking neural
  network model of {3D} perception for event-based neuromorphic stereo vision
  systems,'' \emph{Sci. Rep.}, vol.~7, no.~1, Jan. 2017.

\bibitem{Marr76Science}
D.~Marr and T.~Poggio, ``Cooperative computation of stereo disparity,''
  \emph{Science}, vol. 194, no. 4262, pp. 283--287, 1976.

\bibitem{Steffen19fnbot}
L.~Steffen, D.~Reichard, J.~Weinland, J.~Kaiser, A.~R{\"{o}}nnau, and
  R.~Dillmann, ``Neuromorphic stereo vision: {A} survey of bio-inspired sensors
  and algorithms,'' \emph{Front. Neurorobot.}, vol.~13, p.~28, 2019.

\bibitem{Rebecq18ijcv}
H.~Rebecq, G.~Gallego, E.~Mueggler, and D.~Scaramuzza, ``{EMVS}: Event-based
  multi-view stereo---{3D} reconstruction with an event camera in real-time,''
  \emph{Int. J. Comput. Vis.}, vol. 126, no.~12, pp. 1394--1414, Dec. 2018.

\bibitem{Gallego18cvpr}
G.~Gallego, H.~Rebecq, and D.~Scaramuzza, ``A unifying contrast maximization
  framework for event cameras, with applications to motion, depth, and optical
  flow estimation,'' in \emph{{IEEE} Conf. Comput. Vis. Pattern Recog. (CVPR)},
  2018, pp. 3867--3876.

\bibitem{Cook11ijcnn}
M.~Cook, L.~Gugelmann, F.~Jug, C.~Krautz, and A.~Steger, ``Interacting maps for
  fast visual interpretation,'' in \emph{Int. Joint Conf. Neural Netw.
  (IJCNN)}, 2011, pp. 770--776.

\bibitem{Kim14bmvc}
H.~Kim, A.~Handa, R.~Benosman, S.-H. Ieng, and A.~J. Davison, ``Simultaneous
  mosaicing and tracking with an event camera,'' in \emph{British Mach. Vis.
  Conf. (BMVC)}, 2014.

\bibitem{Reinbacher17iccp}
C.~Reinbacher, G.~Munda, and T.~Pock, ``Real-time panoramic tracking for event
  cameras,'' in \emph{{IEEE} Int. Conf. Comput. Photography (ICCP)}, 2017, pp.
  1--9.

\bibitem{Weikersdorfer12robio}
D.~Weikersdorfer and J.~Conradt, ``Event-based particle filtering for robot
  self-localization,'' in \emph{{IEEE} Int. Conf. Robot. Biomimetics (ROBIO)},
  2012, pp. 866--870.

\bibitem{Weikersdorfer13icvs}
D.~Weikersdorfer, R.~Hoffmann, and J.~Conradt, ``Simultaneous localization and
  mapping for event-based vision systems,'' in \emph{Int. Conf. Comput. Vis.
  Syst. (ICVS)}, 2013, pp. 133--142.

\bibitem{Mueggler14iros}
E.~Mueggler, B.~Huber, and D.~Scaramuzza, ``Event-based, 6-{DOF} pose tracking
  for high-speed maneuvers,'' in \emph{IEEE/RSJ Int. Conf. Intell. Robot. Syst.
  (IROS)}, 2014, pp. 2761--2768.

\bibitem{Gallego19cvpr}
G.~Gallego, M.~Gehrig, and D.~Scaramuzza, ``Focus is all you need: Loss
  functions for event-based vision,'' in \emph{{IEEE} Conf. Comput. Vis.
  Pattern Recog. (CVPR)}, 2019, pp. 12\,272--12\,281.

\bibitem{Migliore20icraw}
D.~{Migliore (Prophesee)}, ``Sensing the world with event-based cameras,''
  \url{https://robotics.sydney.edu.au/icra-workshop/}, Jun. 2020.

\bibitem{Lagorce17pami}
X.~Lagorce, G.~Orchard, F.~Gallupi, B.~E. Shi, and R.~Benosman, ``{HOTS}: A
  hierarchy of event-based time-surfaces for pattern recognition,''
  \emph{{IEEE} Trans. Pattern Anal. Mach. Intell.}, vol.~39, no.~7, pp.
  1346--1359, Jul. 2017.

\bibitem{Quigley09icraoss}
M.~Quigley, K.~Conley, B.~Gerkey, J.~Faust, T.~Foote, J.~Leibs, R.~Wheeler, and
  A.~Y. Ng, ``{ROS}: an open-source {R}obot {O}perating {S}ystem,'' in
  \emph{{ICRA} Workshop Open Source Softw.}, vol.~3, no.~2, 2009, p.~5.

\bibitem{Hirschmuller08pami}
H.~Hirschmuller, ``Stereo processing by semiglobal matching and mutual
  information,'' \emph{{IEEE} Trans. Pattern Anal. Mach. Intell.}, vol.~30,
  no.~2, pp. 328--341, Feb. 2008.

\bibitem{Liu10nb}
S.-C. Liu and T.~Delbruck, ``Neuromorphic sensory systems,'' \emph{Current
  Opinion in Neurobiology}, vol.~20, no.~3, pp. 288--295, 2010.

\bibitem{Delbruck08issle}
T.~Delbruck, ``Frame-free dynamic digital vision,'' in \emph{Proc. Int. Symp.
  Secure-Life Electron.}, 2008, pp. 21--26.

\bibitem{Gehrig19iccv}
D.~Gehrig, A.~Loquercio, K.~G. Derpanis, and D.~Scaramuzza, ``End-to-end
  learning of representations for asynchronous event-based data,'' in
  \emph{Int. Conf. Comput. Vis. (ICCV)}, 2019.

\bibitem{Benosman14tnnls}
R.~Benosman, C.~Clercq, X.~Lagorce, S.-H. Ieng, and C.~Bartolozzi,
  ``Event-based visual flow,'' \emph{{IEEE} Trans. Neural Netw. Learn. Syst.},
  vol.~25, no.~2, pp. 407--417, 2014.

\bibitem{Zhu18rss}
A.~Z. Zhu, L.~Yuan, K.~Chaney, and K.~Daniilidis, ``{EV-FlowNet}:
  Self-supervised optical flow estimation for event-based cameras,'' in
  \emph{Robotics: Science and Systems (RSS)}, 2018.

\bibitem{Engel14eccv}
J.~Engel, J.~Sch\"ops, and D.~Cremers, ``{LSD-SLAM}: Large-scale direct
  monocular {SLAM},'' in \emph{Eur. Conf. Comput. Vis. (ECCV)}, 2014, pp.
  834--849.

\bibitem{zhou2018canny}
Y.~Zhou, H.~Li, and L.~Kneip, ``Canny-{VO}: Visual odometry with {RGB-D}
  cameras based on geometric 3-{D}--2-{D} edge alignment,'' \emph{{IEEE} Trans.
  Robot.}, vol.~35, no.~1, pp. 184--199, 2018.

\bibitem{benosman11tnn}
R.~Benosman, S.-H. Ieng, P.~Rogister, and C.~Posch, ``Asynchronous event-based
  {H}ebbian epipolar geometry,'' \emph{{IEEE} Trans. Neural Netw.}, vol.~22,
  no.~11, pp. 1723--1734, 2011.

\bibitem{Newcombe11iccv}
R.~A. Newcombe, S.~J. Lovegrove, and A.~J. Davison, ``{DTAM}: Dense tracking
  and mapping in real-time,'' in \emph{Int. Conf. Comput. Vis. (ICCV)}, 2011,
  pp. 2320--2327.

\bibitem{Zhu18ral}
A.~Z. Zhu, D.~Thakur, T.~Ozaslan, B.~Pfrommer, V.~Kumar, and K.~Daniilidis,
  ``The multivehicle stereo event camera dataset: An event camera dataset for
  {3D} perception,'' \emph{{IEEE} Robot. Autom. Lett.}, vol.~3, no.~3, pp.
  2032--2039, Jul. 2018.

\bibitem{kotz2004multivariate}
S.~Kotz and S.~Nadarajah, \emph{Multivariate t-distributions and their
  applications}.\hskip 1em plus 0.5em minus 0.4em\relax Cambridge University
  Press, 2004.

\bibitem{Kerl13icra}
C.~Kerl, J.~Sturm, and D.~Cremers, ``Robust odometry estimation for rgb-d
  cameras,'' in \emph{{IEEE} Int. Conf. Robot. Autom. (ICRA)}, 2013.

\bibitem{Mueggler17ijrr}
E.~Mueggler, H.~Rebecq, G.~Gallego, T.~Delbruck, and D.~Scaramuzza, ``The
  event-camera dataset and simulator: Event-based data for pose estimation,
  visual odometry, and {SLAM},'' \emph{Int. J. Robot. Research}, vol.~36,
  no.~2, pp. 142--149, 2017.

\bibitem{roth2017robust}
M.~Roth, T.~Ardeshiri, E.~{\"O}zkan, and F.~Gustafsson, ``Robust bayesian
  filtering and smoothing using student's t distribution,'' \emph{arXiv
  preprint arXiv:1703.02428}, 2017.

\bibitem{Baker04ijcv}
S.~Baker and I.~Matthews, ``Lucas-kanade 20 years on: A unifying framework,''
  \emph{Int. J. Comput. Vis.}, vol.~56, no.~3, pp. 221--255, 2004.

\bibitem{cayleyparameter}
A.~Cayley, ``About the algebraic structure of the orthogonal group and the
  other classical groups in a field of characteristic zero or a prime
  characteristic,'' in \emph{Reine Angewandte Mathematik}, 1846.

\bibitem{Piatkowska17cvprw}
E.~Piatkowska, J.~Kogler, N.~Belbachir, and M.~Gelautz, ``Improved cooperative
  stereo matching for dynamic vision sensors with ground truth evaluation,'' in
  \emph{{IEEE} Conf. Comput. Vis. Pattern Recog. Workshops (CVPRW)}, 2017.

\bibitem{Zhu18eccv}
A.~Z. Zhu, Y.~Chen, and K.~Daniilidis, ``Realtime time synchronized event-based
  stereo,'' in \emph{Eur. Conf. Comput. Vis. (ECCV)}, 2018, pp. 438--452.

\bibitem{Brandli13fns}
C.~Brandli, T.~Mantel, M.~Hutter, M.~H{\"o}pflinger, R.~Berner, R.~Siegwart,
  and T.~Delbruck, ``Adaptive pulsed laser line extraction for terrain
  reconstruction using a dynamic vision sensor,'' \emph{Front. Neurosci.},
  vol.~7, p. 275, 2014.

\bibitem{Sturm12iros}
J.~Sturm, N.~Engelhard, F.~Endres, W.~Burgard, and D.~Cremers, ``A benchmark
  for the evaluation of {RGB-D} {SLAM} systems,'' in \emph{IEEE/RSJ Int. Conf.
  Intell. Robot. Syst. (IROS)}, Oct. 2012.

\bibitem{MurArtal17tro}
R.~Mur-Artal and J.~D. Tard{\'o}s, ``{ORB-SLAM2}: An open-source {SLAM} system
  for monocular, stereo, and {RGB-D} cameras,'' \emph{{IEEE} Trans. Robot.},
  vol.~33, no.~5, pp. 1255--1262, Oct. 2017.

\bibitem{Besl92pami}
P.~J. Besl and N.~D. McKay, ``A method for registration of {3-D} shapes,''
  \emph{{IEEE} Trans. Pattern Anal. Mach. Intell.}, vol.~14, no.~2, pp.
  239--256, 1992.

\bibitem{Son17isscc}
B.~Son, Y.~Suh, S.~Kim, H.~Jung, J.-S. Kim, C.~Shin, K.~Park, K.~Lee, J.~Park,
  J.~Woo, Y.~Roh, H.~Lee, Y.~Wang, I.~Ovsiannikov, and H.~Ryu, ``A 640x480
  dynamic vision sensor with a 9$\mu$m pixel and {300Meps} address-event
  representation,'' in \emph{{IEEE} Intl. Solid-State Circuits Conf. (ISSCC)},
  2017.

\bibitem{Liu18bmvc}
M.~Liu and T.~Delbruck, ``Adaptive time-slice block-matching optical flow
  algorithm for dynamic vision sensors,'' in \emph{British Mach. Vis. Conf.
  (BMVC)}, 2018.

\bibitem{Rebecq18corl}
H.~Rebecq, D.~Gehrig, and D.~Scaramuzza, ``{ESIM}: an open event camera
  simulator,'' in \emph{Conf. on Robotics Learning (CoRL)}, 2018.

\end{thebibliography}

\vskip -2\baselineskip plus -1fil

\begin{IEEEbiography}[{\includegraphics[width=1in,height=1.25in,clip,keepaspectratio]{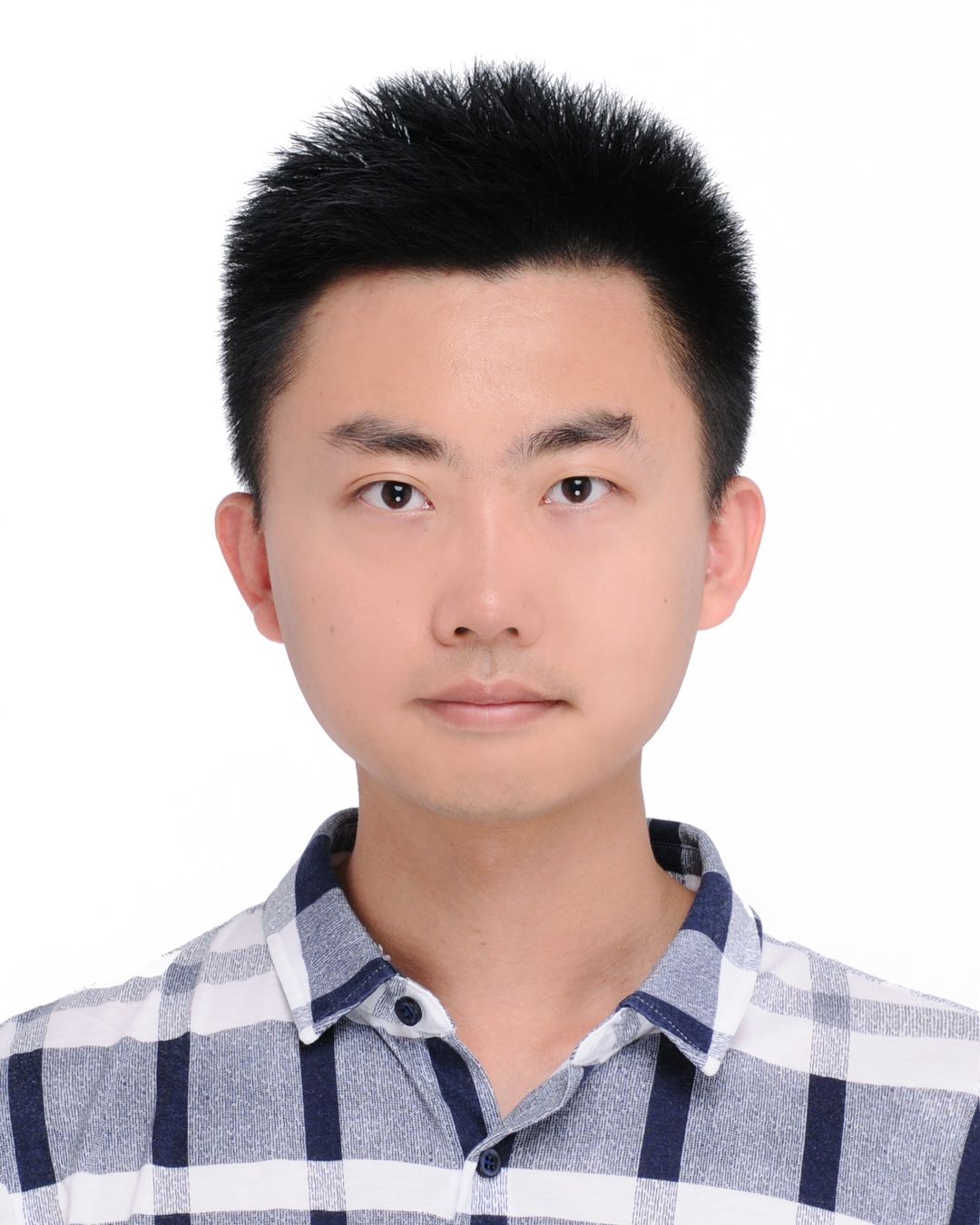}}]{Yi Zhou} received the B.Sc. degree in aircraft manufacturing and engineering from Beijing University of Aeronautics and Astronautics, Beijing, China in 2012, and the Ph.D. degree with the Research School of Engineering, Australian National University, Canberra, ACT, Australia in 2018.
Since 2019 he is a postdoctoral researcher at the Hong Kong University of Science and Technology, Hong Kong.
His research interests include visual odometry / simultaneous localization and mapping, geometry problems in computer vision, and dynamic vision sensors.
Dr. Zhou was awarded the NCCR Fellowship Award for the research on event based vision in 2017 by the Swiss National Science Foundation through the National Center of Competence in Research Robotics.
\end{IEEEbiography}

\begin{IEEEbiography}[{\includegraphics[width=1in,height=1.25in,clip,keepaspectratio]{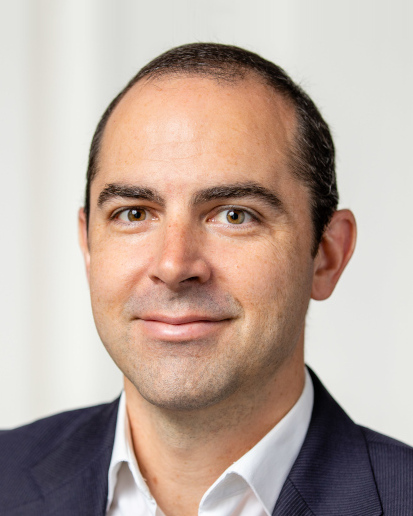}}]{Guillermo Gallego} (SM'19) is Associate Professor at the Technische Universit\"at Berlin, in the Dept. of Electrical Engineering and Computer Science, and at the Einstein Center Digital Future, both in Berlin, Germany.
He received the PhD degree in Electrical and Computer Engineering from the Georgia Institute of Technology, USA, in 2011, supported by a Fulbright Scholarship.
From 2011 to 2014 he was a Marie Curie researcher with Universidad Politecnica de Madrid, Spain, and from 2014 to 2019 he was a postdoctoral researcher at the Robotics and Perception Group, 
University of Zurich, Switzerland.
His research interests include robotics, computer vision, signal processing, optimization and geometry. 
\end{IEEEbiography}

\begin{IEEEbiography}[{\includegraphics[width=1in,height=1.25in,clip,keepaspectratio]{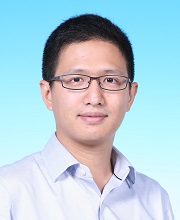}}]{Shaojie Shen} (Member, IEEE) received the B.Eng. degree in electronic engineering from the Hong Kong University of Science and Technology, Hong Kong, in 2009, the M.S. degree in robotics and the Ph.D. degree in electrical and systems engineering, both from the University of Pennsylvania, Philadelphia, PA, USA, in 2011 and 2014, respectively.
He joined the Department of Electronic and Computer Engineering, Hong Kong University of Science and Technology in September 2014 as an Assistant Professor, and was promoted to associate professor in 2020.
His research interests are in the areas of robotics and unmanned aerial vehicles, with focus on state estimation, sensor fusion, computer vision, localization and mapping, and autonomous navigation in complex environments.
\end{IEEEbiography}

\end{document}